\documentclass[%
   , hidempi
]{mpi2015-cscpreprint}

\usepackage[american]{babel}

\usepackage{graphicx}

\usepackage{amssymb}
\usepackage{amsthm}

\numberwithin{equation}{section}
\numberwithin{figure}{section}
\numberwithin{table}{section}

\usepackage{subcaption} 
\usepackage{enumitem}

\usepackage{multirow}

\usepackage{tikz}
\usetikzlibrary{calc}

\usepackage{xcolor}

\usepackage{algorithm}
\usepackage{algorithmic}

\usepackage{import}

\newcommand*\xbar[1]{%
	\hbox{%
		\vbox{%
			\hrule height 0.5pt 
			\kern0.5ex%
			\hbox{%
				\kern-0.35em%
				\ensuremath{#1}%
				\kern-0em%
			}%
		}%
	}%
}
\newcommand{\bx}{\ensuremath{\boldsymbol{{x}}}}
\newcommand{\bbx}{\ensuremath{\xbar{\boldsymbol{{\ x}}}}}

\DeclareMathOperator*{\argmax}{argmax}

\newtheorem{remark}{Remark}[section]

\usepackage{combelow}

\usepackage{placeins}

\begin{document}
  

\title{Active-Learning-Driven Surrogate Modeling for Efficient Simulation of Parametric Nonlinear Systems}
  
\author[$\ast$]{Harshit Kapadia} 
\affil[$\ast$]{Max Planck Institute for Dynamics of Complex Technical Systems, 39106 Magdeburg, Germany.\authorcr
  Corresponding author, \email{kapadia@mpi-magdeburg.mpg.de}, \orcid{0000-0003-3214-0713}}
  
\author[$\dagger$]{Lihong Feng}
\affil[$\dagger$]{Max Planck Institute for Dynamics of Complex Technical Systems, 39106 Magdeburg, Germany.\authorcr
  \email{feng@mpi-magdeburg.mpg.de}, \orcid{0000-0002-1885-3269}}

\author[$\ddagger$]{Peter Benner}
\affil[$\ddagger$]{Max Planck Institute for Dynamics of Complex Technical Systems, 39106 Magdeburg, Germany.\authorcr
	\email{benner@mpi-magdeburg.mpg.de}, \orcid{0000-0003-3362-4103}}
  
\shorttitle{Active-Learning-Driven POD-KSNN Surrogate Model} 

\shortauthor{H. Kapadia, L. Feng, P. Benner}
\shortdate{}


\abstract{%
	When repeated evaluations for varying parameter configurations of a high-fidelity physical
	model are required, surrogate modeling techniques based on model order reduction are desired. In absence of the governing equations describing the dynamics, we need to construct the parametric reduced-order surrogate model in a non-intrusive fashion. In this setting, the usual residual-based error estimate for optimal parameter sampling associated with the reduced basis method is not directly
	available. Our work provides a non-intrusive optimality criterion to efficiently populate the
	parameter snapshots, thereby, enabling us to effectively construct a parametric surrogate model. We consider separate parameter-specific proper orthogonal decomposition (POD) subspaces and propose an active-learning-driven surrogate model using kernel-based shallow neural networks, abbreviated as ActLearn-POD-KSNN surrogate model. To demonstrate the validity of our proposed ideas, we present numerical experiments using two	physical models, namely Burgers' equation and shallow water equations. Both the models have mixed---convective and diffusive---effects within their respective parameter domains, with each of them dominating in certain regions. The proposed ActLearn-POD-KSNN surrogate model efficiently predicts the solution at new parameter locations, even for a setting with multiple interacting shock profiles.
}

\novelty{
	\begin{itemize}
		\item ActLearn-POD-KSNN: A novel surrogate modeling framework for parametric systems that is driven by actively learning the high-fidelity solution snapshots.
		\item A new non-intrusive error estimator based on the surrogate solution approximated in parameter-specific POD subspace which enables active learning.
		\item The active learning framework identifies areas in the parameter space with high variation in solution features and generates new snapshots in those regions, enhancing surrogate solution accuracy and refining the learning process iteratively.
		\item The parameter-specific adaptive POD subspaces makes our approach efficient for problems with mixed---convective and diffusive---phenomena, even in settings with multiple interacting shock profiles under convection domination.
		\item The offline training and the online querying is fast due to the shallow neural network architecture used in the construction of ActLearn-POD-KSNN.
	\end{itemize}
}

\maketitle

  
\section{Introduction}%
\label{sec:intro}

In scenarios where computing the full-order model (FOM) becomes computationally expensive, reduced-order modeling techniques provide beneficial alternatives. In recent years, there has been significant interest in developing non-intrusive model order reduction (MOR) approaches as they do not require access to first principle models. As a result, non-intrusive MOR is flexible for constructing reduced-order models (ROMs) for systems that are simulated using a black-box software or systems with limited access to the governing equations. Many of the non-intrusive MOR methods are based on machine learning: some use shallow neural networks, such as radial basis functions (RBFs), while many others use deep learning (DL) networks. 

Many existing DL-MOR methods and RBF-MOR methods learn the ROM by assuming that the solution manifold is well approximated by a linear subspace. Then a uniform reduced basis is computed from the proper orthogonal decomposition (POD) of a snapshot matrix including trajectories of the solutions at different parameter samples~\cite{morKanE18, morGuoH18, morHesU18, morMohG18, morWanVKetal18, morGuoH19, morWanHR19, morBhaHKetal21, morRenMR20, morKosMJ20, morCheHJetal18, morDutFPetal21, morKasGH20, morBerHKetal20, morCheWHetal20}. 
In contrast, the RBF-MOR method in~\cite{morXiaFN17} allows the solution manifold to be nonlinear with respect to the parameter. The snapshots at a new parameter sample are learned via RBF interpolation, which can be interpreted as a neural network with one hidden layer. The reduced basis for the solution space at the new parameter is then available via singular value decomposition (SVD) of the snapshot matrix corresponding to the new parameter. 

Non-intrusively learning a ROM by assuming a nonlinear solution manifold is also proposed in~\cite{morFreDM21, morGonB18, morOttR19} based on deep learning. 
The method in~\cite{morFreDM21} uses a deep feed-forward neural network to compute the reduced state,  then uses a decoder to recover the full state that approximates the original solution. The method in~\cite{morGonB18} instead uses the encoder function to nonlinearly transform the initial state to the reduced state, then uses a recurrent neural network, namely a long short term memory (LSTM) to predict the reduced space at any desired future time. Finally, a decoder function is applied to recover the approximate full solution. Applicability of the method to parametric dynamical problems whose initial condition remains unchanged is unclear. A similar method is proposed in~\cite{morOttR19} where the Koopman matrix is used instead of LSTM in the reduced space for time evolution. 
Intrusive DL-MOR methods and relevant error estimation are proposed in~\cite{morHarM, morLeeC20, morFreC19, morKanE17} that require the discretized governing equations of the PDEs to be known. 

Compared with the high computational cost (repeated optimization, many epochs) of DL-MOR that needs large amount of training data---solution snapshots---the RBF-MOR approach is computationally cheaper due to no optimization, basically employing only a single epoch. Moreover, since the key step in RBF-MOR is interpolation, the RBF-ROM reproduces the snapshot data, while the training data are not guaranteed to be reproduced by the DL-ROMs~\cite{morKosMJ20}.


\subsection{Active-Learning-Driven Surrogate Modeling} 
\label{subsec:main-method}

\begin{figure}[!htb]
	\centering
	\includegraphics[width=0.6\textwidth]{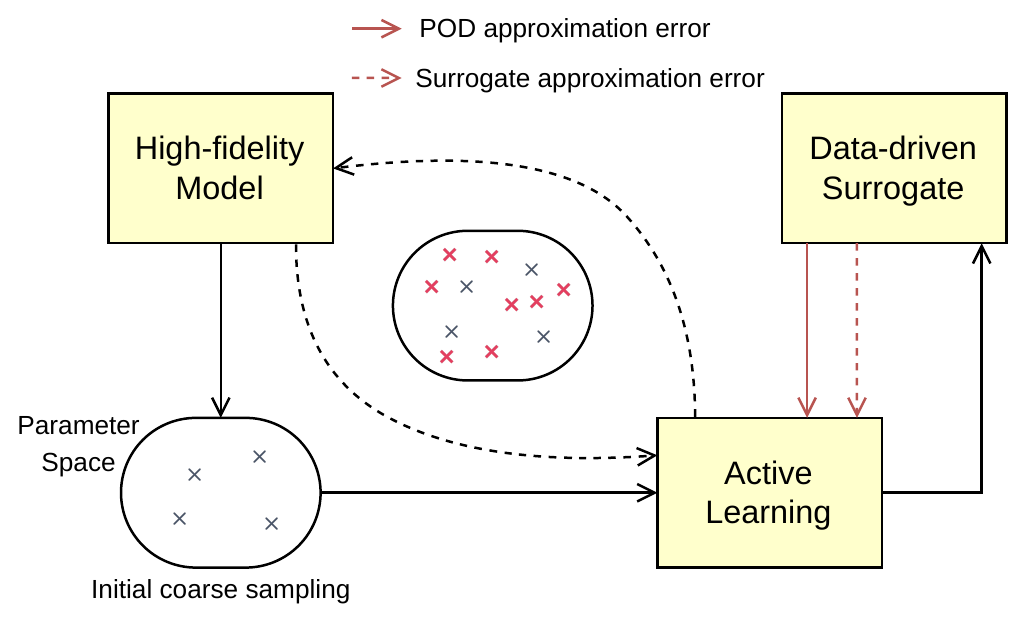}
	\caption{Overview of the active-learning-driven surrogate modeling paradigm.}
	\label{fig:overview}
\end{figure}

When data-driven surrogate models are employed, a substantial amount of training data is typically required to obtain a reasonable approximation of the underlying physics. Generating such a vast amount of training data is computationally expensive since it is obtained by repeated evaluations of a high-fidelity model. Alternately, if the solution data is collected from experimental measurements, conducting repeated experiments for a vast pool of parametric configurations could become practically infeasible. To alleviate this situation, we propose a surrogate modeling framework for parametric nonlinear dynamical systems that actively generates solution snapshots at new parameter locations by querying the high-fidelity model only when necessary. This enables us to iteratively arrive at a set of optimal training data corresponding to important parameter values. By doing this, we improve the surrogate model in an efficient fashion---by relaxing the vast data requirement to some extent and also providing an accuracy estimation of the constructed surrogate model.

\Cref{fig:overview} provides an overview of our proposed active-learning-driven surrogate modeling \linebreak paradigm. An initial coarse sampling of the parameter space is first considered, and the corresponding high-fidelity solution snapshots are generated and stored for a particular discrete time trajectory. With this initial set of snapshots, a data-driven surrogate model is trained. By designing an appropriate optimality criterion that helps us to pick new important parameter locations, we can actively improve the accuracy of the surrogate model. Such an optimality criterion can be designed by using the error caused in the surrogate approximation. However, in this work, we employ a new strategy to design the optimality criterion---constructing an error estimator from the parameter-specific POD-based solution approximations. We utilize a shallow neural network architecture equipped with an RBF kernel as the nonlinear activation to construct the error-estimate-based optimality criterion, as well as to construct the actively learned reduced-order surrogate model. The shallow architecture renders a fast offline training phase, as well as a fast online evaluation phase.

The proposed ActLearn-POD-KSNN surrogate iteratively detects locations in the parameter domain where the variation between solution features is high, and queries the FOM solver in those regions to generate new training snapshots. During this iterative procedure, a POD subspace for the new parameter sample is created and appropriately enriched in an adaptive fashion by using the error estimator. Such a POD subspace enrichment results in a varying number of POD bases between each of the parameter-specific subspaces, corresponding to different levels of energy (information) retention in each subspace. This enables us to choose an appropriate energy criterion for creating POD subspaces at newly queried parameter samples in the online phase such that the subspaces are expressive enough to provide a solution approximation up to a desired accuracy.  For a setting that requires a multi-query parametric generation of solution, the proposed active learning framework becomes useful to build a surrogate model in an efficient fashion---by limiting the generation of the expensive FOM snapshots to an optimal set of parameter samples which still provide a sufficient exploration of the parameter space.


\subsection{Relation to Previous Work}%
\label{subsec:rel-prv-wrk}

Estimating the error of the reduced approximations is crucial to assess their quality. For the reduced basis method \cite{morQuaMN16, morHesRS16-NEW}, an a posteriori error estimator is constructed by using the governing equations which then drives the greedy algorithm for constructing the reduced-order model. To reduce the offline time of the reduced basis method when a large training set of parameter samples are required, authors in \cite{morCheFB22} propose a RBF-interpolation-based surrogate for the error estimator. This reduces the numerous ROM evaluations that are required for the error estimator construction, while enabling sufficient exploration of the parameter space for the reduced basis method. However, efficient error estimation for non-intrusive MOR is still rarely discussed in the literature.
In~\cite{morXia19}, a machine learning technique is applied to learn the error of the RBF-ROM in~\cite{morXiaFN17}. The error is the error of the approximate solution computed from the ROM and is a long vector of the FOM dimension. The error estimator is obtained via two ROMs: the ROM of the FOM and the ROM of the error, so that machine learning via Gaussian processes (GP) is done on the ROM of the error. However, the learning process needs to be implemented for each element of the error vector, i.e., one GP error model is learned for each element of the error vector.  

In this work, we propose a non-intrusive error estimator, built using a KSNN which is equipped with RBF kernels, to assess the quality of a data-driven surrogate model that emulates the physics  of a nonlinear parametric dynamical system. The error estimator is constructed by learning the norm of the POD-approximate state-vector error using interpolation in the parameter-time space. No extra ROM for the error vector needs to be constructed as in~\cite{morXia19}. Furthermore, a single error model is learned rather than quite a few GP models for all the elements of the error vector in~\cite{morXia19}. The proposed error estimator is computationally much cheaper. The training data for the KSNN-based interpolation are the snapshots at certain samples of the parameter and time instances that can then be updated adaptively. We employ the RBF-MOR from~\cite{morXiaFN17} to show the robustness of the proposed error estimator. Beyond the RBF-MOR method in~\cite{morXiaFN17}, we propose a greedy procedure in order to actively learn the POD-KSNN surrogate by adaptively and iteratively updating the snapshot data. This process iteratively improves the accuracy of the POD-KSNN surrogate model and updates the proposed error estimator at the same time. We further propose to use an energy criterion to identify different POD bases corresponding to different parameters. An adaptive technique for enriching the identified POD basis is proposed. This further significantly improves the accuracy of the RBF-MOR method from~\cite{morXiaFN17}, especially for convection-dominated problems. 

Active learning is also proposed in~\cite{morCheHJetal18, morDutFPetal21, morKasGH20, morZhuHBetal22}.  
The method in~\cite{morCheHJetal18} proposes a greedy non-intrusive method that selects the parameters iteratively according to a proposed indicator. However, the indicator has nothing to do with the the error of the approximate solution. The method in~\cite{morDutFPetal21} proposes RBF interpolation for predicting the reduced state vector in the future time instances. Greedy algorithms are proposed to adaptively select the snapshots of the reduced state vector that are called the projected snapshots.  Moreover, the projected snapshots are greedily selected according to a residual and a power function, rather than by error estimation of the approximate solution. The projected snapshots are selected from precomputed solution snapshots at a given set of time instances that need a lot of offline computations. The method applies only to non-parametric time-dependent cases. 

In~\cite{morKasGH20}, Gaussian process regression (GPR) is proposed to learn the reduced state vector as a function of parameters. Active learning using deviation of GP as an indicator to iteratively enrich the training data (snapshots) that are then used for retraining GPR. Again, the deviation of GP cannot tell the error of the approximate solution computed from the proposed method there. Steady-state problems are only addressed in~\cite{morKasGH20}, and extension of the method to time-dependent problems is not straightforward. Similarly, in the most recent work \cite{morZhuHBetal22}, an error estimator based GPR is proposed to perform active learning by using single-time step snapshots of the parametric system states. Their method works for time-dependent problems, but its performance for models with mixed---convective and diffusive---effects is unclear. 

The non-intrusive error estimator in our method is built from the error arising in a parameter-specific POD-approximation of the solution states. To the best of our knowledge, this is in contrast to all the previously proposed frameworks for active learning. As a result, our error estimator based non-intrusive optimality criteria allows us to actively learn important solution snapshots at new parameter locations, completely in the offline phase, without the need to repeatedly evaluate and retrain the entire surrogate model or non-intrusive ROM. This further reduces the computational burden. During the online phase, we do not need to evaluate the high-fidelity model in real-time, but can simply query the actively learned reduced-order surrogate model and obtain efficient approximation of the physics.


\subsection{Organization}%
\label{subsec:orga}

The remaining article is organized as follows. In \Cref{sec:ni-rom}, the general setting for the parametric nonlinear dynamical system is introduced. This is followed by introducing the kernel-based shallow neural network (KSNN) which uses radial basis functions (RBFs) as the kernel functions. Finally, we formulate the POD-based data-driven surrogate model using KSNNs. Next, in \Cref{sec:active-learn-rom}, we propose a non-intrusive optimality criterion based on an error estimator which can be used for actively learning any POD-based surrogate model. \Cref{sec:method-summary} summarizes the novel ActLearn-POD-KSNN surrogate model by detailing its complete algorithm. Then, we provide detailed numerical experiments for models with mixed---convective and diffusive---physical phenomena in \Cref{sec:num-exp}. At the end, we draw some conclusions in \Cref{sec:conclude}.


\section{Data-Driven Surrogate Model for Parametric Systems}
\label{sec:ni-rom}

We can represent a full-order nonlinear dynamical system arising from the spatial discretization of a parametric partial differential equation as
\begin{equation} \label{eqn:parametric-ode}
	\frac{d \mathbf{u}}{d t}=\mathbf{f}(\mathbf{u}, t ; \boldsymbol{\mu}), \quad \mathbf{u}(0)=\mathbf{u}_{0}(\boldsymbol{\mu}), \quad t \in[0, T],
\end{equation}
where $T \in \mathbb{R}^{+}$ denotes the final time; $\mathbf{u} \equiv \mathbf{u}(t, \boldsymbol{\mu})$ with $\mathbf{u}:[0, T] \times \mathcal{D} \rightarrow \mathbb{R}^{N}$ denotes the solution; $\mathbf{u}_{0}: \mathcal{D} \rightarrow \mathbb{R}^{N}$ denotes the parameterized initial condition; $\boldsymbol{\mu} \in \mathcal{D} \subseteq \mathbb{R}^{N_{\mu}}$ denotes the parameters; and $\mathbf{f}: \mathbb{R}^{N} \times[0, T] \times \mathcal{D} \rightarrow \mathbb{R}^{N}$ denotes a nonlinear function. In this section, we provide a formulation of a reduced-order surrogate model which can be constructed directly from the high-fidelity solution snapshots of \cref{eqn:parametric-ode}. The surrogate model is built by employing a series of neural networks with a shallow neural network architecture. The shallowness enables a fast offline training procedure as well as a rapid online querying of the surrogate model at new out-of-training parameter locations.


\subsection{Kernel-Based Shallow Neural Network} 

\begin{figure}[t] 
	\centering
		\import{figures/}{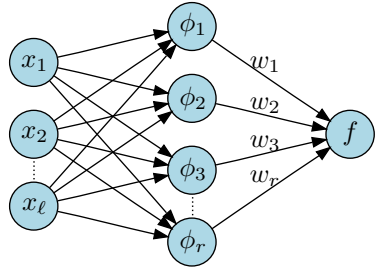}
	\caption{Illustration of a radial kernel-based shallow neural network with a scalar output.}
	\label{fig:rbf-net}
\end{figure}

We formulate the interpolation technique that will be used in this work as a radial kernel-based shallow neural network (KSNN). The network is as shown in \Cref{fig:rbf-net} with an input, a hidden, and an output layer. The input layer includes the data points $\bx_j$ where $j=1,\ldots,\ell$. The activation functions in the hidden layer are $\phi_i:=\phi(\| \bbx - \bbx_{i} \|)$. The output $f$ of the network is used to learn (approximate) a scalar-valued function $\widehat{f}(\bbx)$. Mathematically, this can be expressed as follows,
\begin{equation} \label{eqn:KSNN-map}
	\widehat{f} (\bbx) \approx f(\bbx) = \sum_{i=1}^{r} w_{i} \phi_{i}(\| \bbx - \bbx_{i} \|).
\end{equation}
where $\{w_{i}\}_{i=1}^{r}$ are the network weights; $\{\bbx_{i}\}_{i=1}^{r}$ are the centers; and $\phi$ is a kernel function depending on the radial distance of input $\bbx$ from a specified center $\bbx_{i}$. As an example, \Cref{tab:rbf-kernels} lists a few different radial basis kernels with the shape factor $\epsilon$, and the radial distance $d$. Note that we indicate the vectors of input and centers with a bar at the top to highlight that they can attain any generic input and center values.

\begin{table}[!htb]
	\begin{center}
		\begin{tabular}{lc}
			Name & Function\\
			\hline\noalign{\medskip}
			Gaussian &  $e^{-(d/\epsilon)^2}$ \\
			Multi-quadric & $\sqrt{(d/\epsilon)^2 + 1}$ \\
			Inverse multi-quadric & $1 / (\sqrt{(d/\epsilon)^2 + 1})$ \\
			Linear spline & $d$ \\
			Cubic spline & $d^3$ \\
			Quintic spline & $d^5$ \\
			Thin-plate spline & $d^2 \log(d)$ \\
			\noalign{\medskip}\hline\noalign{\smallskip}
		\end{tabular}
	\caption{List of radial basis kernels.}
	\label{tab:rbf-kernels}
	\end{center}
\end{table}

The radial kernel or basis functions operate on multivariate input data, which in turn reduces to a scalar function of the Euclidean norm of $(\bbx - \bbx_{i})$. For exact interpolation, we take $\bbx_i = \bx_i$, $r=l$, and enforce $\widehat{f}(\bx_j) = f(\bx_j)$ with $j=1,\ldots,\ell$ in \cref{eqn:KSNN-map}. This reduces the training step to a linear system solve for the weights $\{w_{i}\}_{i=1}^{r}$. The coefficient matrix of the linear system is a distance matrix $D \in \mathbb R^{\ell \times \ell}$, where the entries of $D$ are the kernel values evaluated at all the data points ($D_{j,i} = \phi_i(\bx_j)$ with $i = j = 1, \dots, \ell$). After training, we can evaluate the scalar value $\widehat{f}$ at any new input $\bbx$.

\begin{figure}[!b]
	\centering
		\import{figures/}{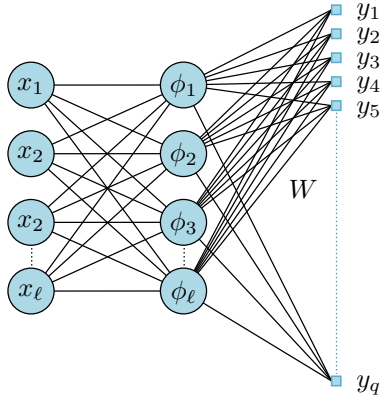}
	\caption{Illustration of a radial kernel-based shallow neural network with a vectorial output. For interpolation, the width of the hidden layer is the same as the number of inputs $\ell$.} 
	\label{fig:rbf-net-vector}
\end{figure}

To learn a vector-valued function $\boldsymbol{\widehat{y}}(\bbx) \in \mathbb{R}^{q}$, the output layer needs to have width $q$. More precisely, the output $\boldsymbol{y}(\bbx)$ of the KSNN can be written as follows,
\begin{align} 
	\hspace{-3.5pt} \boldsymbol{\widehat{y}} (\bbx) \approx \boldsymbol{y}(\bbx) &=  [ \ y_1(\bbx) , \ y_2(\bbx) , \ \dots , \ y_q(\bbx) \ ]^T \label{eqn:KSNN-vector-map-1}
	\\
	&= \Bigg[ \ \sum_{i=1}^{r} w_{i}^{(1)} \phi_{i}(\| \bbx - \bbx_{i} \|) , \ \sum_{i=1}^{r} w_{i}^{(2)} \phi_{i}(\| \bbx - \bbx_{i} \|) , \ \dots , \ \sum_{i=1}^{r} w_{i}^{(q)} \phi_{i}(\| \bbx - \bbx_{i} \|) \ \Bigg]^T \label{eqn:KSNN-vector-map-2}
\end{align}
where, for exact interpolation, we again take $\bbx_i = \bx_i$, $r=l$, and enforce $\widehat{y}_k(\bx_j) = y_k(\bx_j)$ with $j=1,\ldots,\ell$ and $k=1,\ldots,q$. An illustration of such a network is provided in \Cref{fig:rbf-net-vector}. 

All the weights $\{w_{i}^{(k)}\}_{i=1}^{l}$ corresponding to components $k=1,\ldots,q$ of $\boldsymbol{y}(\bbx)$ can be collected in a weight matrix $W$ such that its entries are defined as below, 
\begin{equation}
	W_{i,k} := w_{i}^{(k)}.
\end{equation}
To obtain the weights in each column of $W$, we need to solve a linear system with the coefficient matrix $D \in \mathbb R^{\ell \times \ell}$ during the training process. We avoid directly solving $q$ linear systems and first perform a pivoted LU decomposition of the distance matrix $D$,
\begin{equation}
	D = P L U
\end{equation}
where $P$ is a permutation matrix, $L$ is a lower triangular matrix with unit diagonal elements, and $U$ is an upper triangular matrix. When such a decomposition is available, each column of $W$ can be obtained by simply performing forward and back substitutions. As a result, instead of performing $q$ computations in $\mathcal{O}(l^3)$, we just perform one LU factorization in $\mathcal{O}(l^3)$ followed by $q$ operations in $\mathcal{O}(l^2)$. The benefit of such an adjacent training approach for vectorial interpolation becomes more prominent as the number of data points $\{\bbx_{j}\}_{j=1}^{l}$ increases, as well as when the size of the vectors to be interpolated becomes significantly large, i.e., $q \gg l$. We employ the multi-quadric kernel function (which is not positive definite) to conduct all our numerical experiments, so LU decomposition is used. But when a symmetric positive definite kernel is considered, we can use the Cholesky decomposition instead of LU decomposition and obtain a further reduction in computational complexity from $\frac{2}{3} \ell^3$ to $\frac{1}{3} \ell^3$ during the first step of matrix factorization while training any KSNN with a vectorial output.

\begin{remark}[\textbf{Preserving positivity when interpolating error values}] 
	In our work, we repeatedly build or retrain KSNNs to construct interpolants for the norm of the relative error caused in the POD-approximate solution, which will be discussed at length in \Cref{sec:active-learn-rom}. While interpolating these small error values, it could happen that the result is a negative value close to zero, which would be nonphysical. This phenomenon is dependent on the distribution of the training data, as well as on the shape factor's ($\epsilon$) value. To ensure the positivity of the error values, we modify the training data by taking the logarithm of all the error values used for training. After querying the network, we need to take the exponential of the result, to obtain the correct (positive) interpolated error value.
\end{remark}


\subsection{POD-KSNN Surrogate Model}%
\label{subsec:pod-rbf-rom}

Consider that we have the solution snapshots along discrete time trajectories $\{ t_0, t_1, \dots , t_{N_t} \}$ with $t_0 = 0$ and $t_{N_t} = T$. The snapshots can be collected in matrices $U(\boldsymbol{\mu}_i)$ sized $N \times (N_t+1)$ corresponding to each parameter sample $\boldsymbol{\mu}_i$ for $i \in \{1, \dots, m\}$,
\begin{equation}
	\label{eqn:sol-snap}
	U(\boldsymbol{\mu}_i) = [ \ \mathbf{u}(t_0,\boldsymbol{\mu}_i) \ | \ \mathbf{u}(t_1,\boldsymbol{\mu}_i) \ | \ \dots \ | \ \mathbf{u}(t_{N_t},\boldsymbol{\mu}_i) \ ].
\end{equation}

We follow a two-step interpolation approach~\cite{morXiaFN17} to construct the non-intrusive reduced-order surrogate model. In the first step, by building $(N_t + 1)$ KSNNs (refer to \cref{eqn:KSNN-vector-map-2}), we interpolate the snapshot data $U(\boldsymbol{\mu}_i)$ in the parameter space corresponding to all time-instances $t_j$ with $j \in \{0, \dots, N_t\}$. The KSNNs' construction and training is done in the offline phase, whereas in the online phase, they are queried at a new parameter instance $\boldsymbol{\mu}^*$. The result is an estimation of the snapshot matrix corresponding to any new $\boldsymbol{\mu}^*$,
\begin{equation} \label{eqn:query-KSNN-surrogates}
	U^I(\boldsymbol{\mu}^*) = [ \ \mathcal{I}^\mu_{t_0}(\boldsymbol{\mu}^*) \ | \ \mathcal{I}^\mu_{t_1}(\boldsymbol{\mu}^*) \ | \ \dots \ | \ \mathcal{I}^\mu_{t_{N_t}}(\boldsymbol{\mu}^*) \ ].
\end{equation}
In this case, $\boldsymbol{y}$ in \cref{eqn:KSNN-vector-map-1} corresponds to each $\mathcal{I}^\mu_{t_j}$ in \cref{eqn:query-KSNN-surrogates} (an $N$-dimensional vector), $\bbx_i$ in \cref{eqn:KSNN-vector-map-2} are $\boldsymbol{\mu}_i$, and $\bbx$ is $\boldsymbol{\mu}^*$ when the KSNN is queried. Each column of $U^I$ is obtained by performing a vectorial interpolation. So, $\{\mathcal{I}^\mu_{t_j}\}_{j=0}^{N_t}$ denote $(N_t + 1)$ KSNNs interpolating the snapshots in the parameter space. Due to this reason, we interchangeably refer to the KSNNs as interpolants. It is crucial to note that once the KSNNs $\{\mathcal{I}^\mu_{t_j}\}_{j=0}^{N_t}$ are constructed, we do not need to store any snapshot matrices $U(\boldsymbol{\mu}_i)$ that were used to train the KSNNs.

\begin{figure}[b] %
	\centering
	\includegraphics[width=1\textwidth]{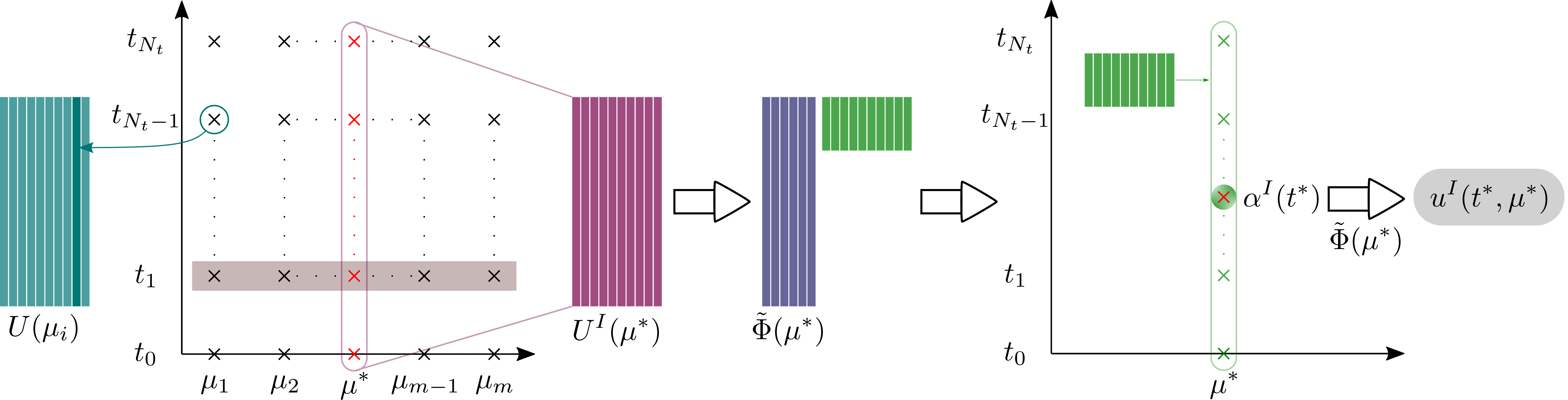} 
	\caption{Framework for POD-KSNN data-driven surrogate modeling.} 
	\label{fig:NIROM}
\end{figure}

The optimal linear subspace spanned by the approximate snapshot data for $\boldsymbol{\mu}^*$ can be computed via proper orthogonal decomposition (POD). In practice, we can compute the POD bases $\tilde{\Phi}(\boldsymbol{\mu}^*):=(\tilde{\phi}_1(\boldsymbol{\mu}^*), \tilde{\phi}_2(\boldsymbol{\mu}^*), \ldots, \tilde{\phi}_s(\boldsymbol{\mu}^*))$ either using the singular value decomposition or (for large-scale problems) using the method of snapshots, when $N$ is very large and $N_t\ll N$ \cite{Sir1987}. The interpolated snapshots at $\boldsymbol{\mu}^*$ can then be written as a linear combination of the bases $\tilde{\Phi}(\boldsymbol{\mu}^*)$. By collecting all the coefficients for such a linear combination in a matrix $A(\boldsymbol{\mu}^*)$, the interpolated snapshots $U^I(\boldsymbol{\mu}^*)$ can be represented as,
\begin{align}
	U^I(\boldsymbol{\mu}^*) &= \tilde{\Phi}(\boldsymbol{\mu}^*) A(\boldsymbol{\mu}^*), \label{eqn:KSNN-sol-projection}\\
	A(\boldsymbol{\mu}^*) &= [ \ \alpha^0(\boldsymbol{\mu}^*) \ | \ \alpha^1(\boldsymbol{\mu}^*) \ | \ \dots \ | \ \alpha^{N_t}(\boldsymbol{\mu}^*) \ ], \label{eqn:KSNN-sol-reduced}
\end{align}
where $\alpha^j \in \mathbb{R}^{s}$ is the vector of coefficients at time $t_j$. For the complete POD space, $s$ equals to the number of nonzero singular values of $U^I(\boldsymbol{\mu}^*)$. However, in practice, usually the POD space is truncated. In that scenario, $s$ corresponds to the number of retained singular values.

To obtain the approximation of the solution corresponding to $\boldsymbol{\mu}^*$ at a new time instance $t^* \in [0, T]$, a second interpolation step is carried out in the time domain by building and training an additional KSNN (refer to \cref{eqn:KSNN-vector-map-2}). We construct an interpolant $\mathcal{I}^t_{\mu^*}$ for the reduced coordinates $\alpha^j(\boldsymbol{\mu}^*)$ in the time domain,
\begin{equation} \label{eqn:reduced-coord-new-time}
	\alpha^I(t^*) := \mathcal{I}^t_{\mu^*}(t^*).
\end{equation}
In this case, $\boldsymbol{y}$ in \cref{eqn:KSNN-vector-map-1} corresponds to $\alpha^I$ in \cref{eqn:reduced-coord-new-time} (an $s$-dimensional vector), $\bbx_i$ in \cref{eqn:KSNN-vector-map-2} is $t_j$, and $\bbx$ is $t^*$ when the interpolant is queried. The surrogate approximation of the solution is obtained as follows,
\begin{equation} \label{eqn:sol-new-time}
	\mathbf{u}_s(t^*, \boldsymbol{\mu}^*) = \tilde{\Phi}(\boldsymbol{\mu}^*) \ \alpha^I(t^*).
\end{equation}

The complete POD-KSNN surrogate model is summarized in \Cref{fig:NIROM}. We adhere to a two-step interpolation approach to divide the function complexity between space and time domains, thereby allowing us to work with multiple reduced-sized KSNNs. The network size (layer width) is directly related to the number of centers or data points under consideration. More data points lead to a wider KSNN, whose training will require a larger linear system solve for the weights. As the number of centers increases, the memory required for the linear system solve goes up considerably. This is due to the quadratic dependence of the required memory on the number of centers. The training could become infeasible in such a scenario. However, the two-step interpolation strategy enables us to isolate the centers between space and time, thereby relaxing the total permissible center count.

\begin{remark}[\textbf{Extension to system of parametric PDEs}]
	One approach is to prepare different snapshot matrices for all the solution components that are present in the system of equations. Then prepare separate KSNNs for all components to interpolate between the snapshots in the parameter domain. Later, in the second step, we create POD subspaces, individually, for each of the components. An additional KSNN for each component interpolates the reduced coordinates in the time domain. We employ this methodology in our numerical experimants with shallow water equations.
	
	An alternate approach is to merge the solutions for all the components and form one big snapshot matrix corresponding to each parameter instance. This allows us to proceed in the same fashion as detailed above in this section. Undertaking this approach of first merging the component snapshots and then applying POD leads to a single ROM, but with larger reduced size.
\end{remark}


\section{Active Learning for POD-Based Data-Driven Surrogates} 
\label{sec:active-learn-rom}

We are concerned with dynamical systems which are parametric in nature. For this setting, one typically requires a substantial amount of training data at several parameter samples to create a good reduced-order surrogate model. Our aim is to be efficient and choose a set of optimal training samples corresponding to different parameters, from a vast pool of parameters. However, there is no trivial notion of optimality. We address this by proposing a non-intrusive error estimator as an optimality criterion. This is further used to actively create the training or snapshot data and leverage the most out of the POD-KSNN surrogate. 


\subsection{Non-Intrusive Optimality Criterion}%
\label{subsec:error-estimate}

To assess the quality of the POD-based data-driven surrogate solution in \Cref{sec:ni-rom}, we require a way to estimate the error in its approximate solution, in comparison with the full-order (or high-fidelity) solution. There are two types of errors induced while constructing and deploying the POD-based surrogate: the error caused due to restricting the solution corresponding to each parameter sample in an (active) linear subspace obtained via POD, and the amalgamation of errors arising from the chosen interpolation or regression technique. The total error $\mathcal{E} \in \mathbb{R}^{N}$ in the spatial domain for some parameter $\boldsymbol{\mu}$ corresponding to a time instance $t$ can be represented as
\begin{equation} \label{eqn:total-error}
	\mathcal{E}(t,\boldsymbol{\mu}) = \mathcal{E}_{POD}(t,\boldsymbol{\mu}) + \mathcal{E}_{I}(t,\boldsymbol{\mu}),
\end{equation}
where $\mathcal{E}_{POD} \in \mathbb{R}^{N}$ represents the POD projection error, and $\mathcal{E}_{I} \in \mathbb{R}^{N}$ represents the total interpolation error when the surrogate model is evaluated. 

To understand the additive decomposition of the total error $\mathcal{E}(t,\boldsymbol{\mu})$, let us look more concretely at the error in the surrogate solution $\mathbf{u}_s(t,\boldsymbol{\mu})$, 
\begin{equation}
	\mathcal{E}(t, \boldsymbol{\mu}) := \mathbf{u}(t,\boldsymbol{\mu}) - \mathbf{u}_s(t,\boldsymbol{\mu}).
\end{equation}
This can be written in the following fashion for the POD-KSNN surrogate model:
\begin{equation} \label{eqn:total-error-expanded}
	\mathcal{E} = (\mathbf{u} - \Phi \Phi^\top \mathbf{u})
	+ 
	(\Phi \Phi^\top \mathbf{u} - \Phi \Phi^\top \mathbf{u}^I)
	+
	(\Phi \Phi^\top \mathbf{u}^I - \tilde{\Phi} \tilde{\Phi}^\top \mathbf{u}^I)
	+
	(\tilde{\Phi} \tilde{\Phi}^\top \mathbf{u}^I - \mathbf{u}_s).
\end{equation}
Here, $\Phi(\boldsymbol{\mu})=(\phi_1(\boldsymbol{\mu}), \phi_2(\boldsymbol{\mu}), \ldots, \phi_s(\boldsymbol{\mu}))$ are the bases obtained by performing POD of the solution $\mathbf{u}$; $\mathbf{u}^I(t,\boldsymbol{\mu}) = \mathcal{I}^\mu_{t}(\boldsymbol{\mu})$ is an approximation of $\mathbf{u}$ obtained via interpolating the solutions in the parameter domain using KSNNs, as shown in \cref{eqn:query-KSNN-surrogates}; $\tilde{\Phi}(\boldsymbol{\mu})=(\tilde{\phi}_1(\boldsymbol{\mu}), \tilde{\phi}_2(\boldsymbol{\mu}), \ldots, \tilde{\phi}_s(\boldsymbol{\mu}))$ are the bases obtained by performing POD of the approximate solution $\mathbf{u}^I$. 

By following \cref{eqn:sol-new-time}, we can write the surrogate solution as $\mathbf{u}_s(t,\boldsymbol{\mu}) = \tilde{\Phi}(\boldsymbol{\mu}) \alpha^I(t)$. Here, an approximation of the reduced solution coordinates, $\alpha^I(t)$, is obtained by interpolating $\{\alpha^j(\boldsymbol{\mu})\}_{j=0}^{N_t}$ from \cref{eqn:KSNN-sol-reduced} in the time domain using a KSNN, i.e., $\alpha^I(t) = \mathcal{I}^t_{\mu}(t)$. This allows us to write \cref{eqn:total-error-expanded} in the following way:
\begin{equation} \label{eqn:total-error-expanded-2}
	\mathcal{E} = (\mathbf{u} - \Phi \Phi^\top \mathbf{u})
	+ 
	(\Phi \Phi^\top \mathbf{u} - \Phi \Phi^\top \mathbf{u}^I)
	+
	(\Phi \Phi^\top \mathbf{u}^I - \tilde{\Phi} \tilde{\Phi}^\top \mathbf{u}^I)
	+
	(\tilde{\Phi} \tilde{\Phi}^\top \mathbf{u}^I - \tilde{\Phi} \alpha^I).
\end{equation}

The first term in \cref{eqn:total-error-expanded-2} arises due to the retention of only the leading $s$ POD modes. This results in a parameter-specific linear subspace that captures most of the solution, but not in its entirety. We refer to this omitted contribution as the POD approximation error and define it as, 
\begin{equation} \label{eqn:error-POD-contrib}
	\mathcal{E}_{POD} := (\mathbf{u} - \Phi \Phi^\top \mathbf{u}).
\end{equation}
The second and third terms in \cref{eqn:total-error-expanded-2} arise from the solution approximation $\boldsymbol{u}^I$, due to interpolation in the parameter domain. More precisely, the second term accounts for the solution error resulting from the reduced representation of the approximate solution obtained by projection onto the true solution bases $\Phi$. Whereas, the third term accounts for the solution error caused due to projection of the approximate solution $\mathbf{u}^I$ onto the bases $\tilde{\Phi}$ instead of $\Phi$, which is obtained from a POD of $\mathbf{u}^I$. Finally, the fourth term in \cref{eqn:total-error-expanded-2} provides the solution error caused because of an approximation of the reduced coordinates $\alpha^I$ via interpolation in the time domain. As a result, we can define the total interpolation error in the following way,
\begin{equation} \label{eqn:error-interp-contrib}
	\mathcal{E}_{I} := (\Phi \Phi^\top \mathbf{u} - \Phi \Phi^\top \mathbf{u}^I)
	+
	(\Phi \Phi^\top \mathbf{u}^I - \tilde{\Phi} \tilde{\Phi}^\top \mathbf{u}^I)
	+
	(\tilde{\Phi} \tilde{\Phi}^\top \mathbf{u}^I - \mathbf{u}_s).
\end{equation}

From \cref{eqn:total-error-expanded-2,eqn:error-POD-contrib,eqn:error-interp-contrib} it is clear that we can decompose the total error $\mathcal{E}$ in an additive fashion as represented in \cref{eqn:total-error}. Now, consider the norm of the total error in the spatial domain,
\begin{equation}
	\| \mathcal{E}(t,\boldsymbol{\mu}) \| = \| \mathcal{E}_{POD}(t,\boldsymbol{\mu}) + \mathcal{E}_{I}(t,\boldsymbol{\mu}) \|.
\end{equation}
Upon application of the triangle inequality, we obtain
\begin{align}
	\| \mathcal{E}(t,\boldsymbol{\mu}) \| &\le \| \mathcal{E}_{POD}(t,\boldsymbol{\mu}) \| + \| \mathcal{E}_{I}(t,\boldsymbol{\mu}) \|,
	\\
	\epsilon(t,\boldsymbol{\mu}) &\le \epsilon_{POD}(t,\boldsymbol{\mu}) + \epsilon_{I}(t,\boldsymbol{\mu}),
\end{align}
where $\epsilon$, $\epsilon_{POD}$, and $\epsilon_{I}$ represents the spatial norms of the total, POD projection, and interpolation errors respectively. More precisely, they are defined as $\epsilon := \| \mathcal{E} \|$, $\epsilon_{POD} := \| \mathcal{E}_{POD} \|$, and $\epsilon_{I} := \| \mathcal{E}_{I} \|$.

During the construction of the POD-KSNN surrogate model, the KSNN interpolation procedure reproduces the training data exactly which results in $\epsilon_{I} = 0$ for all the training parameter samples. This makes $\epsilon_{POD}$ the bound for the total error $\epsilon$. As a result, we can construct an estimator for the error $\epsilon$ caused in the surrogate solution by interpolating between the errors $\epsilon_{POD}$ corresponding to all the training parameter samples. Afterward, this error estimator can be queried at new parameter samples, providing an approximation for $\epsilon$ at any out-of-training parameters. Through our numerical experiments in \Cref{sec:num-exp}, we see that employing such an error estimate is a reasonable strategy to drive the active learning procedure. The key benefit is that we can carry out active learning entirely in the offline phase without the need to repeatedly evaluate the POD-KSNN surrogate model at the training parameter samples.

\begin{figure}[t] 
	\centering
	\includegraphics[width=0.9\textwidth]{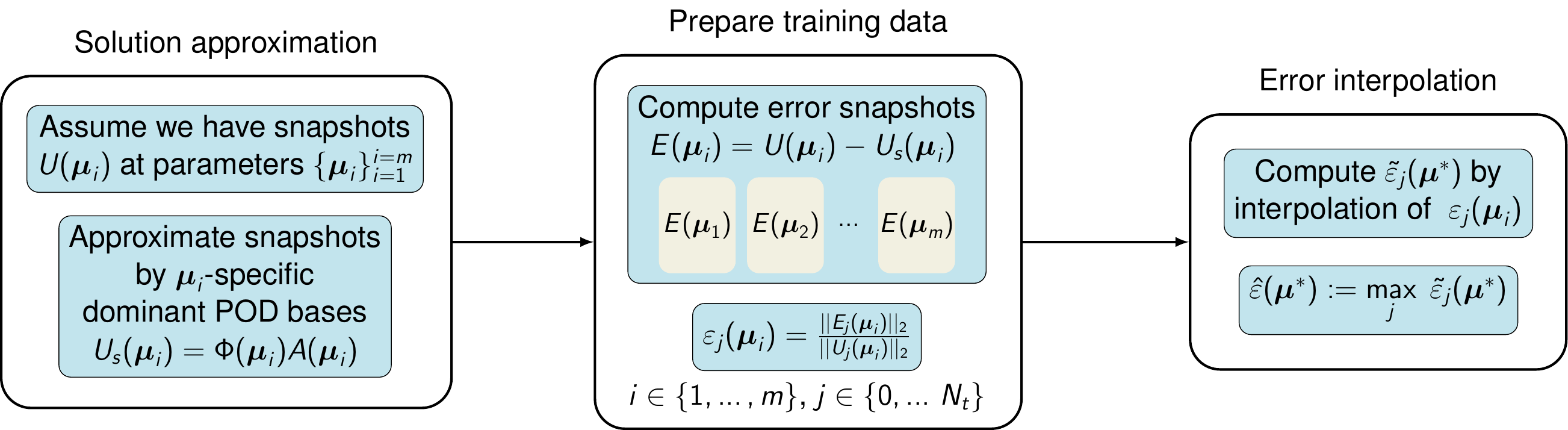}
	\caption{Procedure to non-intrusively estimate the error at new parameter locations.}
	\label{fig:flowchart-error-estimate}
\end{figure}

Let us now dive into the details of the error estimator construction, which we will use as the optimality criterion. Consider a successful parameter sampling, followed by snapshot data collection of all the chosen parameter points corresponding to the same time horizon by simulating the high-fidelity model. For instance, following the setup for \cref{eqn:sol-snap}, we have the snapshot matrices $U(\boldsymbol{\mu}_i)$. Let us consider a POD approximation of the solution snapshots,
\begin{equation} \label{eqn:pod-approx}
	U_{POD}(\boldsymbol{\mu}_i) = \Phi(\boldsymbol{\mu}_i) A(\boldsymbol{\mu}_i)
\end{equation}
with $\Phi(\boldsymbol{\mu}_i) \in \mathbb{R}^{N_x \times r_i}$ and $A(\boldsymbol{\mu}_i) \in \mathbb{R}^{r_i \times (N_t+1)}$. Here, $r_i \le \min \{N_x, (N_t+1)\}$ denotes the level of POD space truncation for each parameter $\boldsymbol{\mu}_i$.

The error of the POD approximate solution at the parameters $\{\boldsymbol{\mu}_1, \boldsymbol{\mu}_2, \dots, \boldsymbol{\mu}_m\}$ is given by
\begin{equation} \label{eqn:error-snap}
	E(\boldsymbol{\mu}_i) = U(\boldsymbol{\mu}_i) - U_{POD}(\boldsymbol{\mu}_i).
\end{equation}
Note that $E(\boldsymbol{\mu}_i)$ is also the POD-KSNN surrogate error because $U_{surrogate} = U_{POD}$ = $U_s$ for all $\boldsymbol{\mu}_i$ (since they are the training parameter samples). We avoid interpolating $E(\boldsymbol{\mu}_i)$ entry-wise at a new parameter location as the computational cost for doing that is equivalent to interpolating the high-dimensional snapshots. Our motivation is to alleviate this computational burden, but at the same time, have an informative indication for the error. This is accomplished by taking the relative norm of the solution error $E_j(\boldsymbol{\mu}_i) \in \mathbb{R}^{N_x}$ at each time instance $t_j$. In our experiments, we have tested using $l^2$, $l^1$, and $l^{\infty}$ norms. We noticed similar qualitative results for all the norms. Depending on the problem setting, one can be preferred over the other, for instance, physical systems prone to advective effects might benefit from $l^1$ norm usage. As a generic choice, we use the $l^2$ norm for our discussion because the considered test problems in \Cref{sec:num-exp} have mixed diffusive and convective effects. We denote the relative norm by,
\begin{equation} \label{eqn:rel-error}
	\varepsilon_j(\boldsymbol{\mu}_i) = \frac{||E_j(\boldsymbol{\mu}_i)||_2}{||U_j(\boldsymbol{\mu}_i)||_2},
\end{equation}
where $U_j(\boldsymbol{\mu}_i)$ denotes the $j^{th}$ column (corresponding to $t_j$) of the snapshot matrix for $\boldsymbol{\mu}_i$.

The relative error values $\varepsilon_j(\boldsymbol{\mu}_i)$ are used to train $(N_t + 1)$ KSNNs and obtain their weight values $w_i^{(j)}$. The interpolated relative $l^2$ error at time $t_j$ for any new parameter value $\boldsymbol{\mu}^*$ becomes
\begin{equation} \label{eqn:error-est-time-each-entry}
	\tilde{\varepsilon}_j(\boldsymbol{\mu}^*) = \sum_{i=1}^{m} w_i^{(j)} \phi_i^{(j)} (||\boldsymbol{\mu}^*-\boldsymbol{\mu}_i||).
\end{equation}
This can be written compactly as a vector with entries corresponding to each time instance,
\begin{equation} \label{eqn:error-est-time}
	\boldsymbol{\tilde{\varepsilon}}(\boldsymbol{\mu}^*) = [ \ \tilde{\varepsilon}_0(\boldsymbol{\mu}^*) \ | \ \tilde{\varepsilon}_1(\boldsymbol{\mu}^*) \ | \ \dots \ | \ \tilde{\varepsilon}_{N_t}(\boldsymbol{\mu}^*) \ ].
\end{equation}
The final error estimate is taken to be the maximum interpolated relative error in time, given by
\begin{equation} \label{eqn:error-est}
	\hat{\varepsilon}(\boldsymbol{\mu}^*) := ||\boldsymbol{\tilde{\varepsilon}}(\boldsymbol{\mu}^*)||_\infty = \max (|\tilde{\varepsilon}_0(\boldsymbol{\mu}^*)|, |\tilde{\varepsilon}_1(\boldsymbol{\mu}^*)|, \dots, |\tilde{\varepsilon}_{N_t}(\boldsymbol{\mu}^*)|).
\end{equation}

The entire procedure to compute the non-intrusive error estimator is summarized in \Cref{fig:flowchart-error-estimate}.


\subsection{Active Learning Framework}%
\label{subsec:active-learn}

The intention of the active learning procedure is to enrich the snapshot data in a fashion that is most beneficial for the reduced-order surrogate model. In other words, each enrichment of the training snapshots lead to an optimal or near-optimal improvement of the approximate dynamics. The motivation is similar to the greedy procedure used for the reduced basis method \cite{morHesRS16-NEW}. However, in our setting we do not have access to the first principle models, so we cannot leverage the equations to decide the choice of new parameter samples for efficient training of the non-intrusive ROM. Instead, we utilize the non-intrusive error estimator as the optimality criterion to enable active learning.

We initialize the parameter set $P$ with a coarse sampling of the parameter space $\mathcal{D}$. Additionally, a second set $P^*$ is prepared which holds all the candidate parameter values that could be included in set $P$ as the active learning progresses. Consider the coarsest initial sampling set $P$, i.e., $P = \{ \boldsymbol{\mu}_i | i \in I \}$ with index set $I = \{1,2\}$. And the candidate set $P^*$ is composed of a fine sampling in $(\boldsymbol{\mu}_1, \boldsymbol{\mu}_2)$, given by
\begin{equation}
	P^* = \{ \boldsymbol{\bar{\mu}}_1, \boldsymbol{\bar{\mu}}_2, \dots, \boldsymbol{\bar{\mu}}_q \} \ \text{with} \ \boldsymbol{\bar{\mu}}_j \neq \boldsymbol{\mu}_i; \ i=1,2; \ j=1,\ldots, q.
\end{equation}

The high-fidelity solution is computed for all the parameters in set $P$. This is followed by performing a POD approximation \cref{eqn:pod-approx} for the snapshot matrix at each $\boldsymbol{\mu}_i$ in $P$. The level of POD space truncation $r_i$ for each $\boldsymbol{\mu}_i \in P$ is obtained by maintaining a constant energy criterion $\eta(\boldsymbol{\mu}_i)$ defined at $\boldsymbol{\mu}_i$ as
\begin{equation} \label{eqn:energy-criterion}
	\eta(\boldsymbol{\mu}_i) = 1 - \frac{\sum_{k=1}^{k=r_i} (\sigma^{(i)}_k)^2}{\sum_{k=1}^{k=s_i} (\sigma^{(i)}_k)^2}.
\end{equation}
Here, $\sigma_k^{(i)}$ with $k=1,\ldots, s_i$, are the nonzero singular values obtained from SVD of the snapshot matrix $U(\boldsymbol{\mu}_i)$.

Using the high-fidelity and POD approximate solutions, error snapshots for the parameters in set $P$ are computed,
\begin{equation} \label{eqn:initial-error-snap}
	E^{(1)}(P) := \{E(\boldsymbol{\mu}_1), E(\boldsymbol{\mu}_2)\}
\end{equation}
where $E(\boldsymbol{\mu}_i)$ is given by \cref{eqn:error-snap} and the superscript in $E^{(1)}$ denotes the first iteration of the active learning loop, i.e. $iter =1$. An error estimator is constructed for the parameters in the candidate set $P^*$,
\begin{equation} \label{eqn:initial-error-estimator}
	\hat{\varepsilon}^{(1)}(P^*) := \{ \hat{\varepsilon}(\boldsymbol{\bar{\mu}}_1), \hat{\varepsilon}(\boldsymbol{\bar{\mu}}_2), \dots, \hat{\varepsilon}(\boldsymbol{\bar{\mu}}_q) \},
\end{equation}
where $\hat{\varepsilon}(\cdot)$ is computed by \cref{eqn:error-est}.

The parameter sample corresponding to the maximal value of the error estimate is chosen ($iter = 1$),
\begin{equation} \label{eqn:para-select-greedy}
	\boldsymbol{\bar{\mu}}^{(iter)} = \argmax_{\boldsymbol{\bar{\mu}}' \in P^*} \hat{\varepsilon}(\boldsymbol{\bar{\mu}}').
\end{equation}
If $\hat{\varepsilon}(\boldsymbol{\bar{\mu}}^{(iter)}) > tol$, the set $P$ is extended by including the chosen parameter $\boldsymbol{\bar{\mu}}^{(iter)}$. The candidate set is also updated, $P^* = P^* \ \backslash \ \boldsymbol{\bar{\mu}}^{(iter)}$. Here, $tol$ is a predefined tolerance level that we intend to achieve. In other words, we terminate the active learning process as soon as the error estimator reaches a value which is less than the target tolerance.

Next, we compute the high-fidelity snapshots $U(\boldsymbol{\bar{\mu}}^{(iter)})$, and the POD approximation error at $\boldsymbol{\bar{\mu}}^{(iter)}$,
\begin{equation}
	E(\boldsymbol{\bar{\mu}}^{(iter)}) = U(\boldsymbol{\bar{\mu}}^{(iter)}) - U_{POD}(\boldsymbol{\bar{\mu}}^{(iter)}),
\end{equation}
where $U_{POD}$ is constructed by adhering to the energy criterion $\eta$ in \cref{eqn:energy-criterion}. The subspace of the POD bases for $\boldsymbol{\bar{\mu}}^{(iter)}$ is further enriched if the maximal relative-error-norm at any time instance in the extended set $P$ corresponds to $\boldsymbol{\bar{\mu}}^{(iter)}$. The POD basis is incremented until the following holds true:
\begin{equation} \label{eqn:pod-adapt}
	\hat{\varepsilon}(\boldsymbol{\bar{\mu}}^{(iter)}) < \max_{\boldsymbol{\mu}' \in P, \ j \in \{0, \dots, N_t\}} \varepsilon_j(\boldsymbol{\mu}').
\end{equation}
This step essentially makes sure that the POD subspace of the newly selected parameter $\boldsymbol{\bar{\mu}}^{(iter)}$ is at least as expressive as all the other parameter-specific POD subspaces corresponding to samples in $P$. Moreover, due to this procedure, we can obtain an update of the energy criterion $\eta(\boldsymbol{\bar{\mu}}^{(iter)})$, which is later used during the online phase to construct an accurate POD subspace for any newly queried parameter $\boldsymbol{\mu^*}$. A more detailed discussion about obtaining and using such an updated energy criterion is provided in \Cref{sec:method-summary}.

\begin{figure}[t] 
	\centering
	\includegraphics[width=1\textwidth]{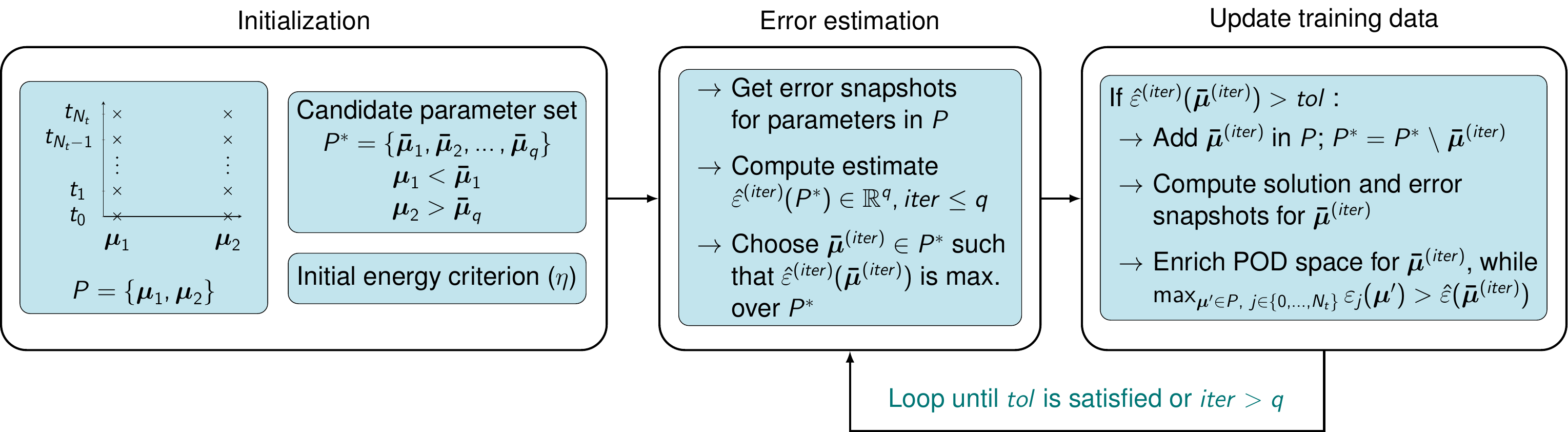}
	\caption{Procedure to actively learn the POD-KSNN surrogate model by greedy refinement of the high-fidelity solution snapshots along the parameter domain.}
	\label{fig:flowchart-active-learning}
\end{figure}

With the necessary POD space enrichment for $\boldsymbol{\bar{\mu}}^{(iter)}$ and the updated parameter set $P$, the error snapshots $E^{(iter)}(P)$ are collected to start the next iteration. This is used as the training data for the interpolation procedure to construct the new error estimator $\hat{\varepsilon}^{(iter)}(P^*)$. The next parameter $\boldsymbol{\bar{\mu}}^{(iter)}$ is then greedily picked using \cref{eqn:para-select-greedy}, followed by an adaptive refinement of the POD space using \cref{eqn:pod-adapt}. In this fashion, we iteratively expand the solution snapshots by greedy selection of new parameters along with their POD space adaptation, until the tolerance $tol$ is satisfied. The complete active learning procedure is outlined in \Cref{fig:flowchart-active-learning}.

\begin{remark}[\textbf{Choice of candidate parameter set $P^*$}]
	The preparation of $P^*$ is not trivial, in fact, a decent sampling procedure needs to be maintained while choosing the candidate samples for $P^*$. As the parameter values range over several orders of magnitude for the physical models presented in our numerical experiments, we consider a uniform sampling of the logarithm of the entire parameter space. This ensures a reasonable selection of parameters through all of the parameter space. 
\end{remark} 

\begin{remark}[\textbf{Extension to system of parametric PDEs}]
	We prepare different snapshot matrices corresponding to all the solution components in the system of equations, i.e., $\{U^{(k)}\}_{k=1}^{q}$ for $q$ components. This is followed by forming the error matrices $\{E^{(k)}\}_{k=1}^{q}$, and then proceeding to apply the aforementioned active learning procedure, but now obtaining component-wise error estimators $\{(\hat{\varepsilon}^{(iter)})^{(k)}\}_{k=1}^{q}$ by following \cref{eqn:initial-error-estimator} for each component. The new parameter sample is picked based on the maximal value of the average between all the component-wise error estimates. So, \cref{eqn:para-select-greedy} takes the following form:
	\begin{equation} \label{eqn:para-select-greedy-SYSTEM}
		\boldsymbol{\bar{\mu}}^{(iter)} = \argmax_{\boldsymbol{\bar{\mu}}' \in P^*} \Bigg( \frac{1}{q} \sum_{k=1}^{q} (\hat{\varepsilon}^{(iter)})^{(k)}(\boldsymbol{\bar{\mu}}') \Bigg).
	\end{equation}
	Similar to the scalar equation setting, now, if $\frac{1}{q} \sum_{k=1}^{q} (\hat{\varepsilon}^{(iter)})^{(k)}(\boldsymbol{\bar{\mu}}') > tol$, the set $P$ is extended by adding $\boldsymbol{\bar{\mu}}^{(iter)}$ to it. The candidate set is also updated, $P^* = P^* \ \backslash \ \boldsymbol{\bar{\mu}}^{(iter)}$. The POD bases are incremented until the component-averaged error estimate value at the newly selected parameter $\boldsymbol{\bar{\mu}}^{(iter)}$ is less than the maximal component-averaged relative-error-norm values among all the parameter samples already present in $P$. So, \cref{eqn:pod-adapt} takes the following form:
	\begin{equation} \label{eqn:pod-adapt-SYSTEM}
		\frac{1}{q} \sum_{k=1}^{q} (\hat{\varepsilon}^{(iter)})^{(k)}(\boldsymbol{\bar{\mu}}^{(iter)}) < \max_{\boldsymbol{\mu}' \in P, \ j \in \{0, \dots, N_t\}} \frac{1}{q} \sum_{k=1}^{q} (\varepsilon_j)^{(k)}(\boldsymbol{\mu}').
	\end{equation} 
\end{remark}

\section{ActLearn-POD-KSNN Surrogate Model} 
\label{sec:method-summary}

We summarize the complete methodology to construct and deploy the ActLearn-POD-KSNN reduced-order surrogate model in this section. \Cref{alg:summary-offline} details the complete offline phase. To iteratively construct the non-intrusive error estimator, a new KSNN (refer to \cref{eqn:KSNN-map}) is automatically built, trained, and queried at steps $5$ and $18$ of the algorithm. Solution snapshots corresponding to new parameter values are actively selected, and the parameter set $P$ is updated during the offline phase. Since the POD subspace is refined adaptively during the active learning process, the relative energy corresponding to the retained POD modes could be different for the selected parameters in the final pool of $P$. We denote it by $\tilde{\eta}(\boldsymbol{\mu}_i)$ for each $\boldsymbol{\mu}_i \in P$. The minimum energy criteria would be
\begin{equation} \label{eqn:new-energy-criterion}
	\hat{\eta} = \min_{\boldsymbol{\mu}' \in P} \tilde{\eta}(\boldsymbol{\mu}').
\end{equation}
We adhere to this updated energy criterion, $\hat{\eta}$, to construct the POD subspace for any new parameter during the online phase.

\begin{algorithm}
	\caption{\texttt{Offline phase: Constructing the ActLearn-POD-KSNN surrogate}} 
	\label{alg:summary-offline}  
	\begin{algorithmic}[1]
		\REQUIRE{Initial parameter set $P$, snapshots $U(\boldsymbol{\mu}_i)$ for $\boldsymbol{\mu}_i \in P$, candidate parameter set $P^*$, initial energy criterion $\eta$, tolerance value ($tol$) to terminate the active learning loop, $iter=1$.}
		\ENSURE{Updated parameter set $P$, energy criterion $\hat{\eta}$, KSNN surrogates $\{\mathcal{I}^\mu_{t_j}\}_{j=0}^{N_t}$.}
		\STATE Based on $\eta$, calculate the POD truncation level $r_i$ for all $\boldsymbol{\mu}_i \in P$ such that \cref{eqn:energy-criterion} holds.
		\STATE Compute nonlinear reduced bases, i.e., the truncated parameter-specific POD subspaces $\Phi(\boldsymbol{\mu}_i)$. \hspace{-1em} 
		\STATE Construct the POD approximate solutions $U_{POD}(\boldsymbol{\mu}_i)$.
		\STATE Obtain the error snapshots $E(\boldsymbol{\mu}_i)$ following \cref{eqn:initial-error-snap,eqn:error-snap}.
		\STATE Compute the error estimate $\hat{\varepsilon}(\boldsymbol{\bar{\mu}}_j)$ using \cref{eqn:initial-error-estimator,eqn:error-est}, where $\boldsymbol{\bar{\mu}}_j \in P^*$. 
		\STATE Pick a parameter $\boldsymbol{\bar{\mu}}^{(iter)}$ from $P^*$ following \cref{eqn:para-select-greedy}. Store $\mathcal{E}^{(iter)} = \hat{\varepsilon}(\boldsymbol{\bar{\mu}}^{(iter)})$.
		\WHILE {$\mathcal{E}^{(iter)} > tol$}  
		\STATE Extend $P$ by including $\boldsymbol{\bar{\mu}}^{(iter)}$. Update the candidate set, $P^* = P^* \ \backslash \ \boldsymbol{\bar{\mu}}^{(iter)}$.
		\STATE Get the solution snapshots $U(\boldsymbol{\bar{\mu}}^{(iter)})$ from a high-fidelity model or experiments.
		\STATE Based on $\eta$, calculate the POD truncation level for $\boldsymbol{\bar{\mu}}^{(iter)}$. 
		\STATE Compute the truncated POD subspace $\Phi(\boldsymbol{\bar{\mu}}^{(iter)})$. Construct $U_{POD}(\boldsymbol{\bar{\mu}}^{(iter)})$.
		\STATE Compute the error snapshots $E(\boldsymbol{\bar{\mu}}^{(iter)})$. 
		\WHILE {$\max_{\boldsymbol{\mu}' \in P, \ j \in \{0, \dots, N_t\}} \varepsilon_j(\boldsymbol{\mu}') > \mathcal{E}^{(iter)}$}
		\STATE Enhance $\Phi(\boldsymbol{\bar{\mu}}^{(iter)})$ by incrementing the truncation level for $\boldsymbol{\bar{\mu}}^{(iter)}$, i.e., adding a new basis. \hspace{-1em} 
		\STATE Recompute $U(\boldsymbol{\bar{\mu}}^{(iter)})$ and $E(\boldsymbol{\bar{\mu}}^{(iter)})$.
		\ENDWHILE 
		\STATE $iter = iter+1$
		\STATE Build the error estimate $\hat{\varepsilon}(\boldsymbol{\mu}_j)$ using \cref{eqn:error-est} for each $\boldsymbol{\mu}_j \in P^*$.
		\STATE Following \cref{eqn:para-select-greedy}, pick a parameter $\boldsymbol{\bar{\mu}}^{(iter)}$ from $P^*$. Store $\mathcal{E}^{(iter)} = \hat{\varepsilon}(\boldsymbol{\bar{\mu}}^{(iter)})$.
		\ENDWHILE
		\STATE Compute $\hat{\eta}$ following \cref{eqn:new-energy-criterion}.
		\STATE Using $U(\boldsymbol{\mu}_i)$ ($\boldsymbol{\mu}_i \in P$) as training data, construct $\{\mathcal{I}^\mu_{t_j}\}_{j=0}^{N_t}$ by building and training $(N_t+1)$ KSNNs \cref{eqn:KSNN-vector-map-2}. As per \cref{eqn:query-KSNN-surrogates}, they will be queried in the online phase at new parameter location $\boldsymbol{\mu}^*$.
	\end{algorithmic}
\end{algorithm}

Building upon the successful active learning procedure carried out during the offline phase, in the online phase, we acquire the surrogate solution at a new time $t^*$ and parameter $\boldsymbol{\mu}^*$ value as described in \Cref{alg:summary-online}. We do not query the full-order model in the online phase. The bases $\tilde{\Phi}(\boldsymbol{\mu}^*)$ is obtained from the snapshot approximation $U^I(\boldsymbol{\mu}^*)$ by performing its SVD to extract the leading $r^*$ POD basis. Consequently, we have to do computations with the cost in $\mathcal{O}(N N_t^2)$, where $N_t < N$. The truncation level $r^*$ is obtained by maintaining the energy criterion $\hat{\eta}$. If $r^*$ can be learned without relying on the energy criterion, the cost of evaluating the bases $\Phi(\boldsymbol{\mu}^*)$ can be reduced to $\mathcal{O}(N N_t \log(r^*))$ with negligible loss of accuracy by leveraging randomized SVD \cite{HalMT11}. This will be included in our subsequent work. Although the online computational cost depends on the dimension of the full-order model, $N$, the online phase is fast, as evidenced by the numerical results presented in \Cref{sec:num-exp}.

\begin{algorithm}
	\caption{\texttt{Online phase: Querying the ActLearn-POD-KSNN surrogate}} 
	\label{alg:summary-online}
	\begin{algorithmic}[1]
		\REQUIRE{New parameter $\boldsymbol{\mu}^*$, new time $t^*$, energy criterion $\hat{\eta}$, KSNN surrogates $\{\mathcal{I}^\mu_{t_j}\}_{j=0}^{N_t}$.}
		\ENSURE{Surrogate solution at ($t^*$, $\boldsymbol{\mu}^*$).}
		\STATE Evaluate $U^I(\boldsymbol{\mu}^*)$, the KSNN approximate snapshots  for $\boldsymbol{\mu}^*$, on the training time grid via \cref{eqn:KSNN-vector-map-2,eqn:query-KSNN-surrogates}. \hspace{-1em}
		\STATE Compute POD bases $\tilde{\Phi}(\boldsymbol{\mu}^*)$ by deciding the truncation level from the energy criterion $\hat{\eta}$.
		\STATE Compute reduced coordinates $A(\boldsymbol{\mu}^*)$ with \cref{eqn:KSNN-sol-projection}, after projecting $U^I(\boldsymbol{\mu}^*)$ onto $\tilde{\Phi}(\boldsymbol{\mu}^*)$ via \cref{eqn:KSNN-sol-reduced}. \hspace{-2em}
		\STATE Build and train a KSNN \cref{eqn:KSNN-vector-map-2}, $\mathcal{I}^t_{\mu^*}$, by using $A(\boldsymbol{\mu}^*)$ as the training data.
		\STATE Obtain a vector of approximate reduced coordinates, $\alpha^I(t^*)$,  at $t^*$ by evaluating $\mathcal{I}^t_{\mu^*}$ via \cref{eqn:KSNN-vector-map-2,eqn:reduced-coord-new-time}. \hspace{-1em}
		\STATE Compute the surrogate solution $u_s(t^*, \boldsymbol{\mu}^*)$ using \cref{eqn:sol-new-time}.
	\end{algorithmic}
\end{algorithm}


\section{Numerical Results}%
\label{sec:num-exp}

We validate the proposed active learning framework with POD-KSNN reduced-order surrogate models by performing numerical experiments on two test cases. The first test case is the Burgers' equation, which is parametrized by the viscosity. It is known to develop an advecting shock in finite time, even when starting with smooth solutions, given a low enough viscosity value. Additionally, we also parametrize the initial condition with viscosity. The second test case is the shallow water equation, which is parametrized by the viscosity and the mean-free path. It is used to model the flow under a pressure surface in a fluid. Its solution comprises two waves moving with opposing characteristic speeds. Due to a periodic boundary condition, the traveling waves repeatedly interact with each other over time. In the remainder of this section, we provide details about both the problem setups and our results for them.


\subsection{Burgers’ Equation}%
\label{subsec:burgers-exp}

 We consider the following viscous Burgers' equation in a 1D spatial domain with Dirichlet boundary conditions:   
\begin{equation} \label{eqn:burgers-eqn}
	\frac{\partial u}{\partial t} + u \frac{\partial u}{\partial x}=\nu \frac{\partial^{2} u}{\partial x^{2}},
\end{equation}
where the solution is denoted by $u$, changing with time $t \ge 0$, space $x \in [0, 1]$, and viscosity $\nu$. The viscosity can be viewed as a parameter that controls the competing effects of diffusion and advection, resulting in a varying solution behavior for a wide range of $\nu$. We choose the following initial condition, such that the solution space at $t=0$ is also parameter-dependent:
\begin{equation}
	u(x, 0)=\frac{x}{1+\sqrt{\frac{1}{\kappa'}} \exp \left(\operatorname{Re} \frac{x^{2}}{4}\right)}
	; \qquad \kappa'=\exp (Re/8), \quad Re=1/\nu.
\end{equation}

\begin{figure}[!t] 
	\centering
	\begin{subfigure}[t]{0.3\textwidth}
		\centering
		\includegraphics[width=1\textwidth, trim=0 0 0 22, clip]{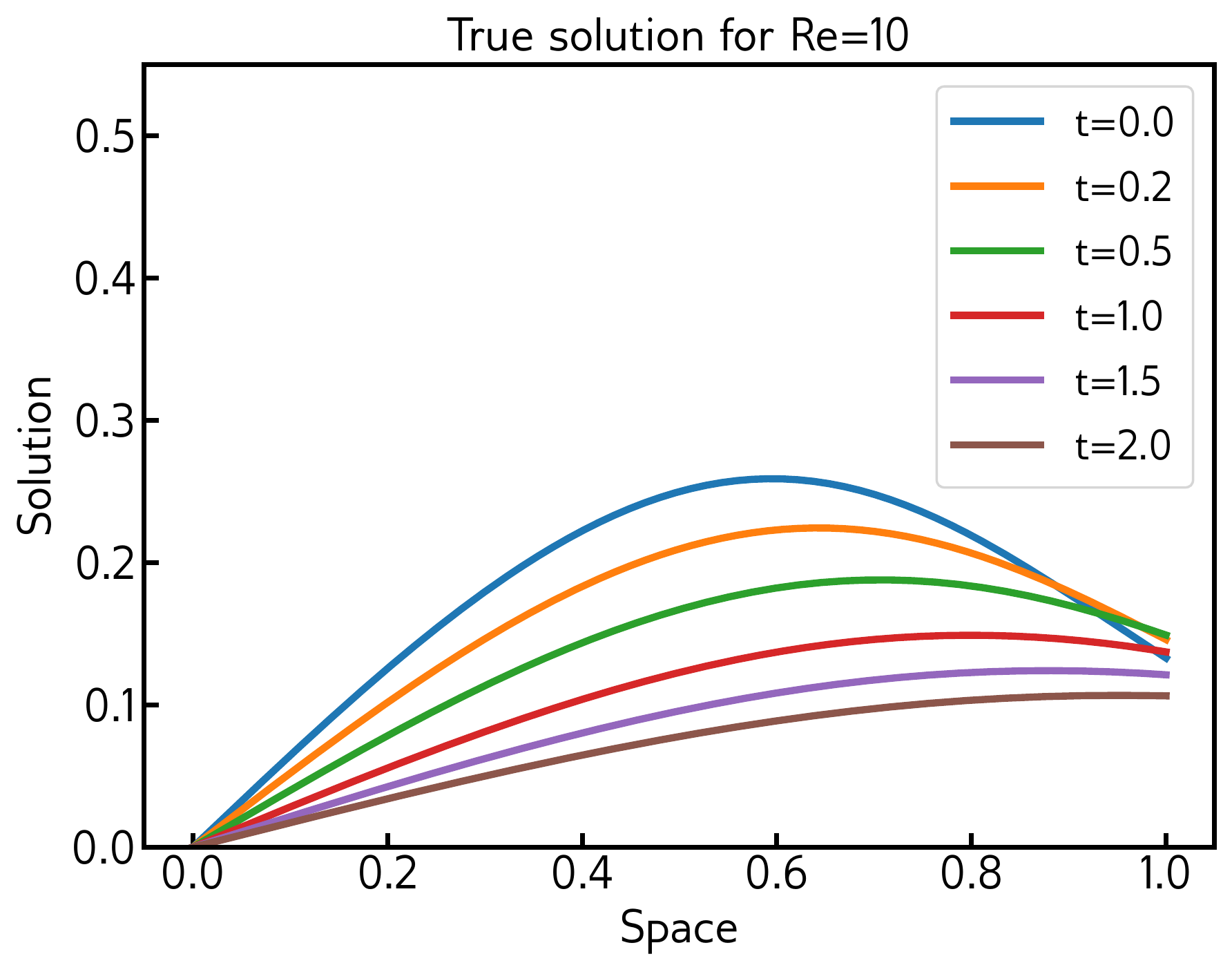}
		\caption{$Re=10$}
		\label{fig:burgers-setup-a}
	\end{subfigure}
	\begin{subfigure}[t]{0.3\textwidth}
		\centering
		\includegraphics[width=1\textwidth, trim=0 0 0 22, clip]{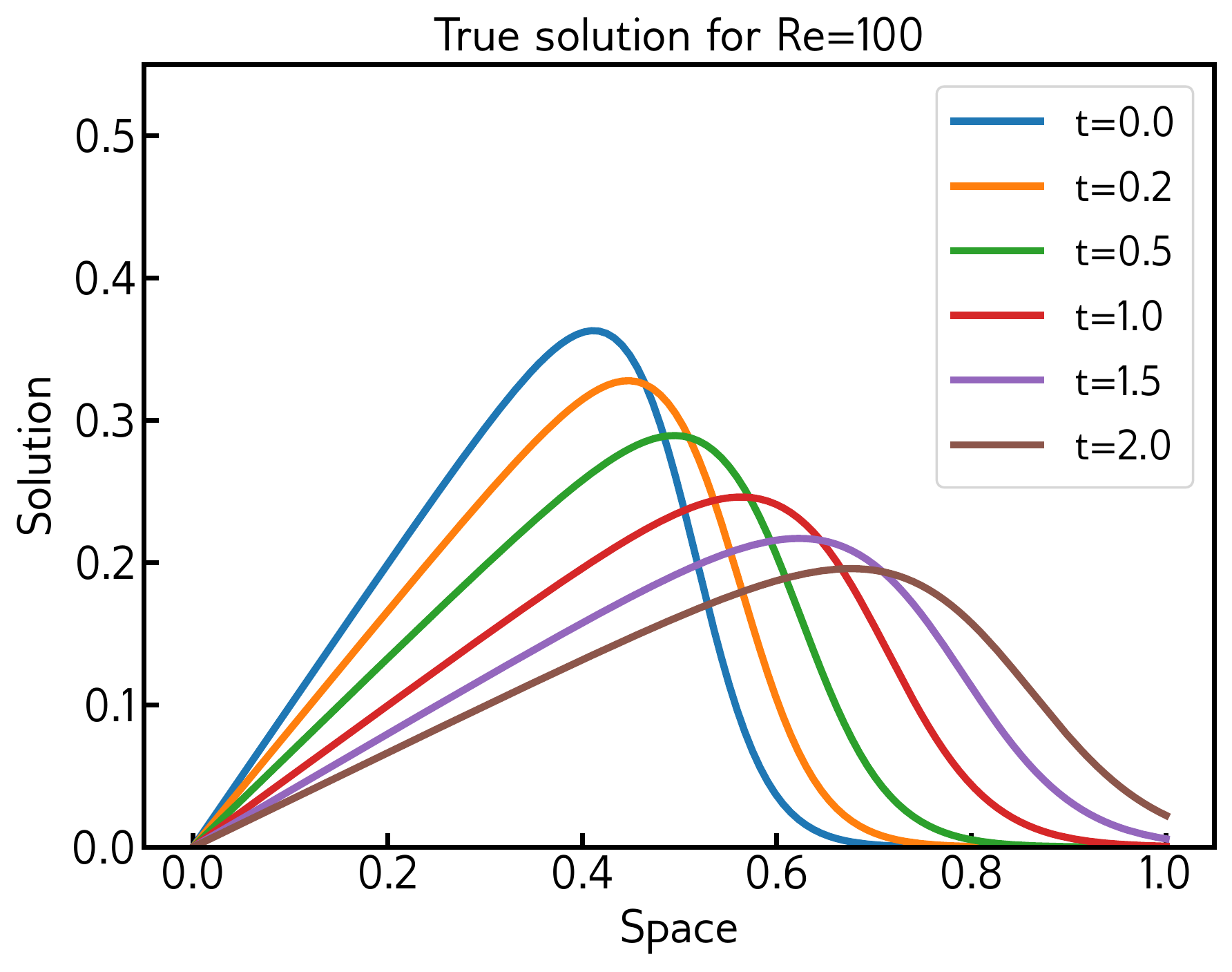}
		\caption{$Re=100$}
		\label{fig:burgers-setup-b}
	\end{subfigure}
	\begin{subfigure}[t]{0.3\textwidth}
		\centering
		\includegraphics[width=1\textwidth, trim=0 0 0 22, clip]{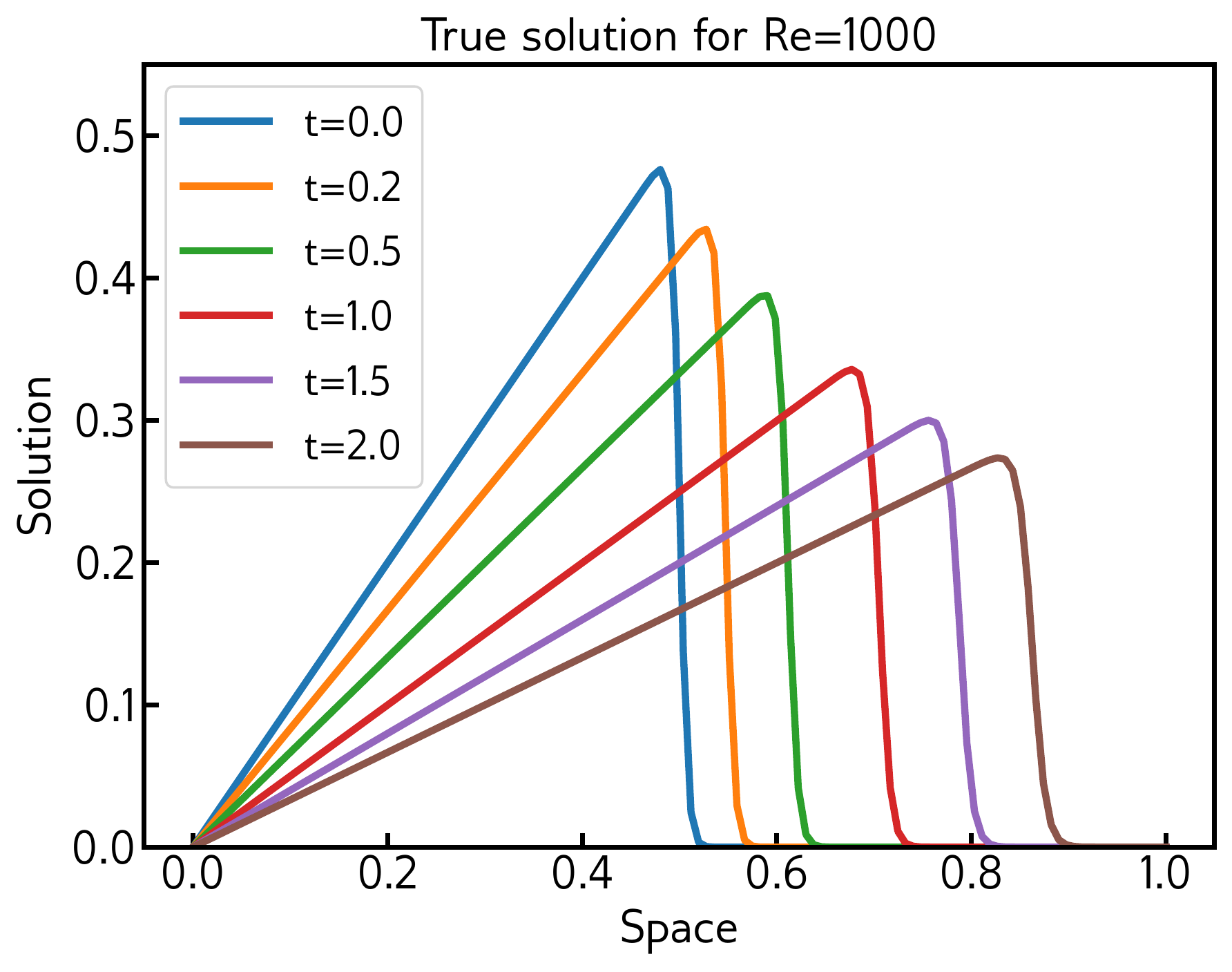}
		\caption{$Re=1000$}
		\label{fig:burgers-setup-c}
	\end{subfigure}
	\begin{subfigure}[t]{0.3\textwidth}
		\centering
		\includegraphics[width=1\textwidth, trim=0 0 0 22, clip]{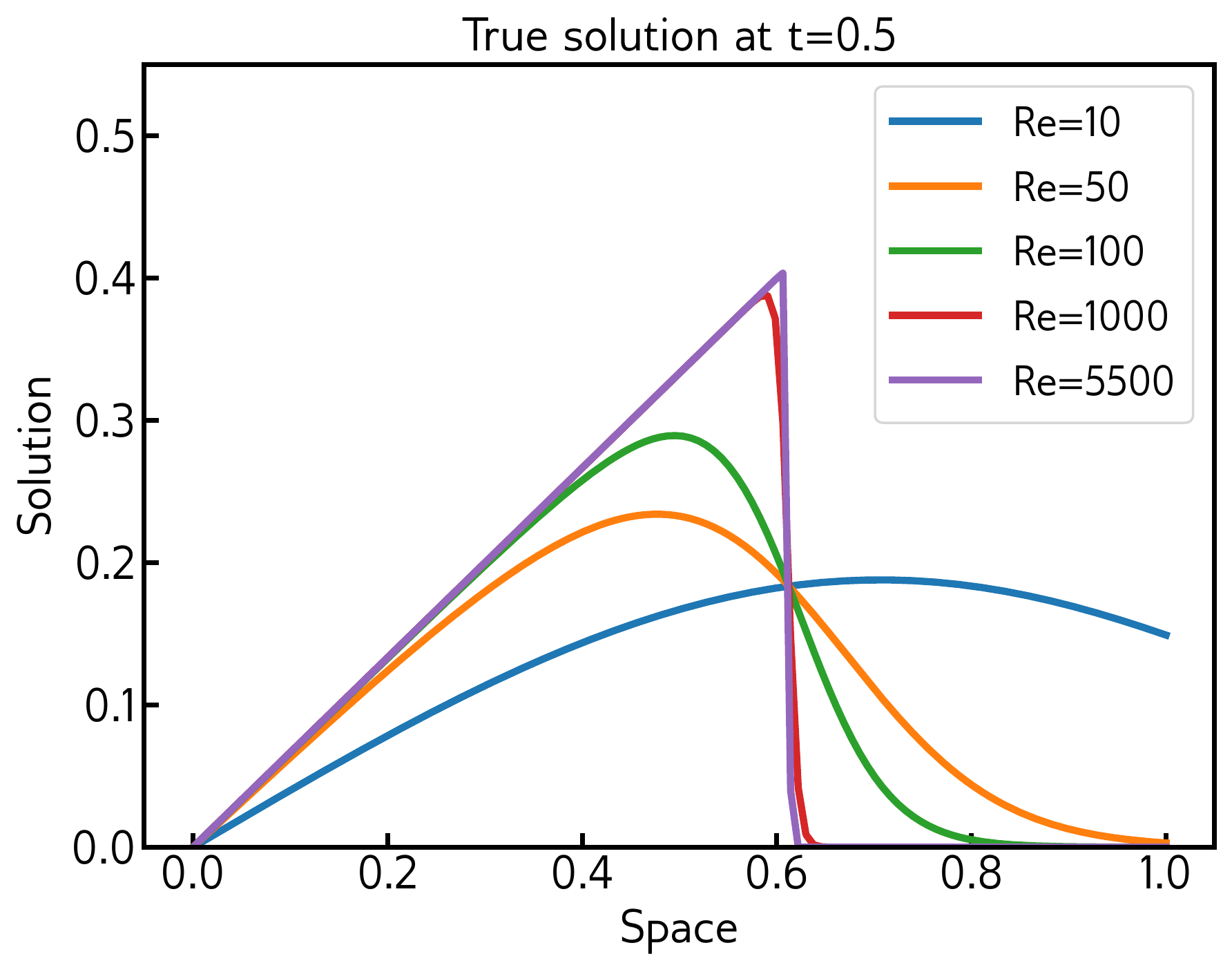}
		\caption{$t=0.5$}
		\label{fig:burgers-setup-d}
	\end{subfigure}
	\begin{subfigure}[t]{0.3\textwidth}
		\centering
		\includegraphics[width=1.026\textwidth, trim=0 0 0 0, clip]{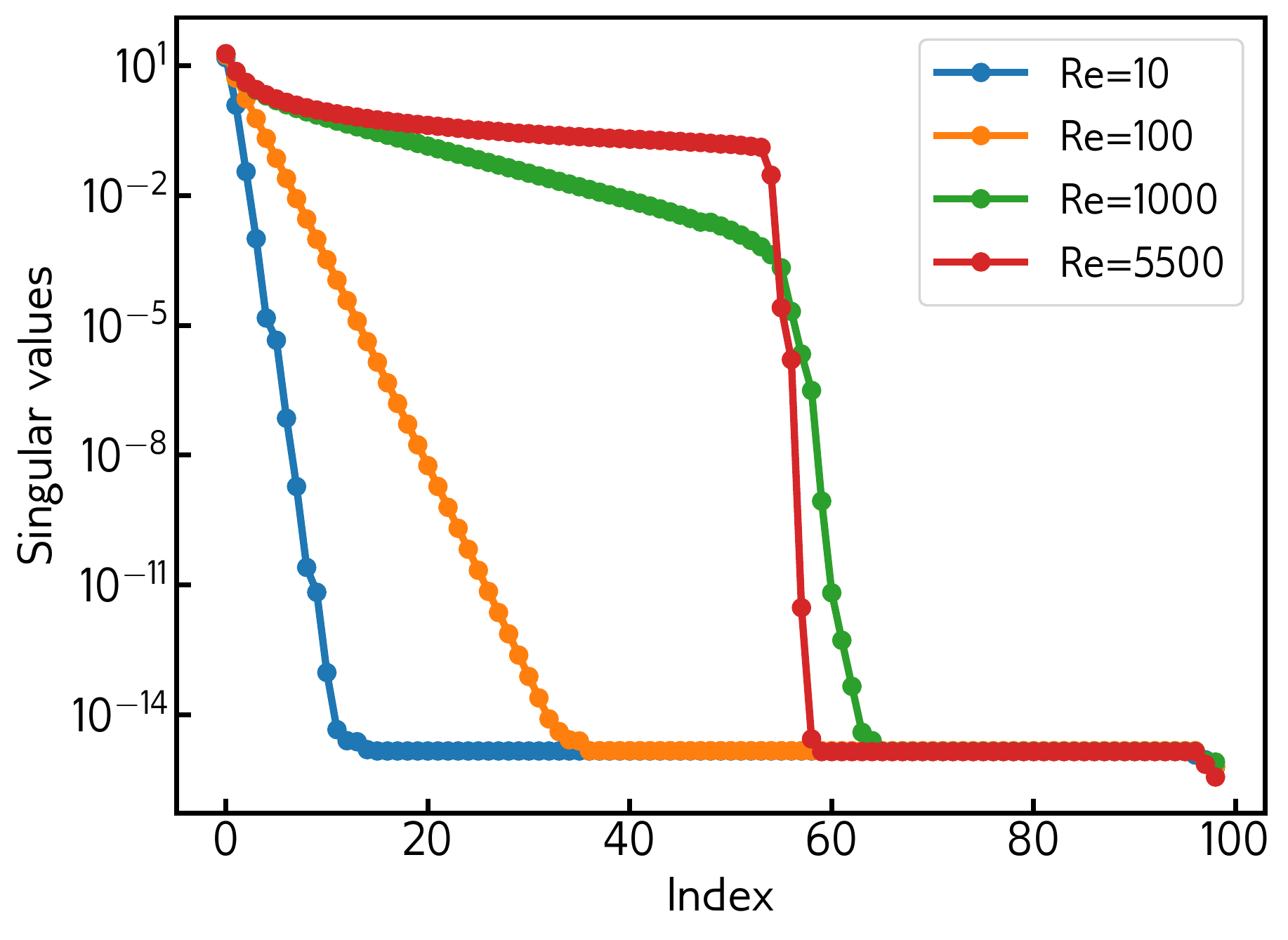}
		\caption{Decay of singular values}
		\label{fig:burgers-setup-e}
	\end{subfigure}
	\caption{Burgers' equation: Evolution of exact solution over time $t \in [0, 2]$ in (a), (b), (c) for various Reynolds numbers; Comparison of solution at $t=0.5$ in (d) for different Reynolds number; Comparison of the singular value decay in (e) for different Reynolds numbers.}
	\label{fig:burgers-setup}
\end{figure}

Instead of simulating the equation using a numerical discretization technique, we opt to utilize the exact solution by converting \cref{eqn:burgers-eqn} into a parabolic nonlinear PDE through the application of the Cole-Hopf transformation \cite{Eva10}. The exact solution takes the following form: 
\begin{equation}
	u(x, t)=\frac{\frac{x}{t+1}}{1+\sqrt{\frac{t+1}{\kappa'}} \exp \left(\operatorname{Re} \frac{x^{2}}{4 t+4}\right)}
	; \qquad \kappa'=\exp (Re/8), \quad Re=1/\nu,
\end{equation}
where we refer to $Re$ as the Reynolds number, denoting the inverse of viscosity.

\Cref{fig:burgers-setup-a,fig:burgers-setup-b,fig:burgers-setup-c,fig:burgers-setup-d} show the behavior of the solution for selected time instances $t \in [0,2]$ corresponding to various Reynolds numbers. To highlight the competing diffusive and convective effects of the equation, we select representative $Re$ values from different parametric regimes. Upon singular value decomposition of the snapshot matrices, we observe a progressively slow singular value decay in \Cref{fig:burgers-setup-e} as we go towards higher $Re$, which is due to the convective effects dominating the flow.

For our experiments, the spatial domain has $150$ grid nodes, and the time-domain is $t \in [0,2]$ with 100 time-steps. The parametric range for the Reynolds number is taken from $10$ to $5500$. The total number of discrete parameters we consider when accounting for both the candidate set $P^*$ and the parameter set $P$ are $100$. These $100$ values of $Re$ are picked by uniformly dividing the logarithmic $\nu$ values ($log_{10}(\frac{1}{5500})$ and $log_{10}(\frac{1}{10})$) into $99$ intervals, ensuring a decent pool of parameters.

\begin{figure}[!b] 
	\centering
	\begin{subfigure}[b]{0.355\textwidth}
		\centering
		\includegraphics[width=1.13\textwidth, trim=0 0 0 0, clip]{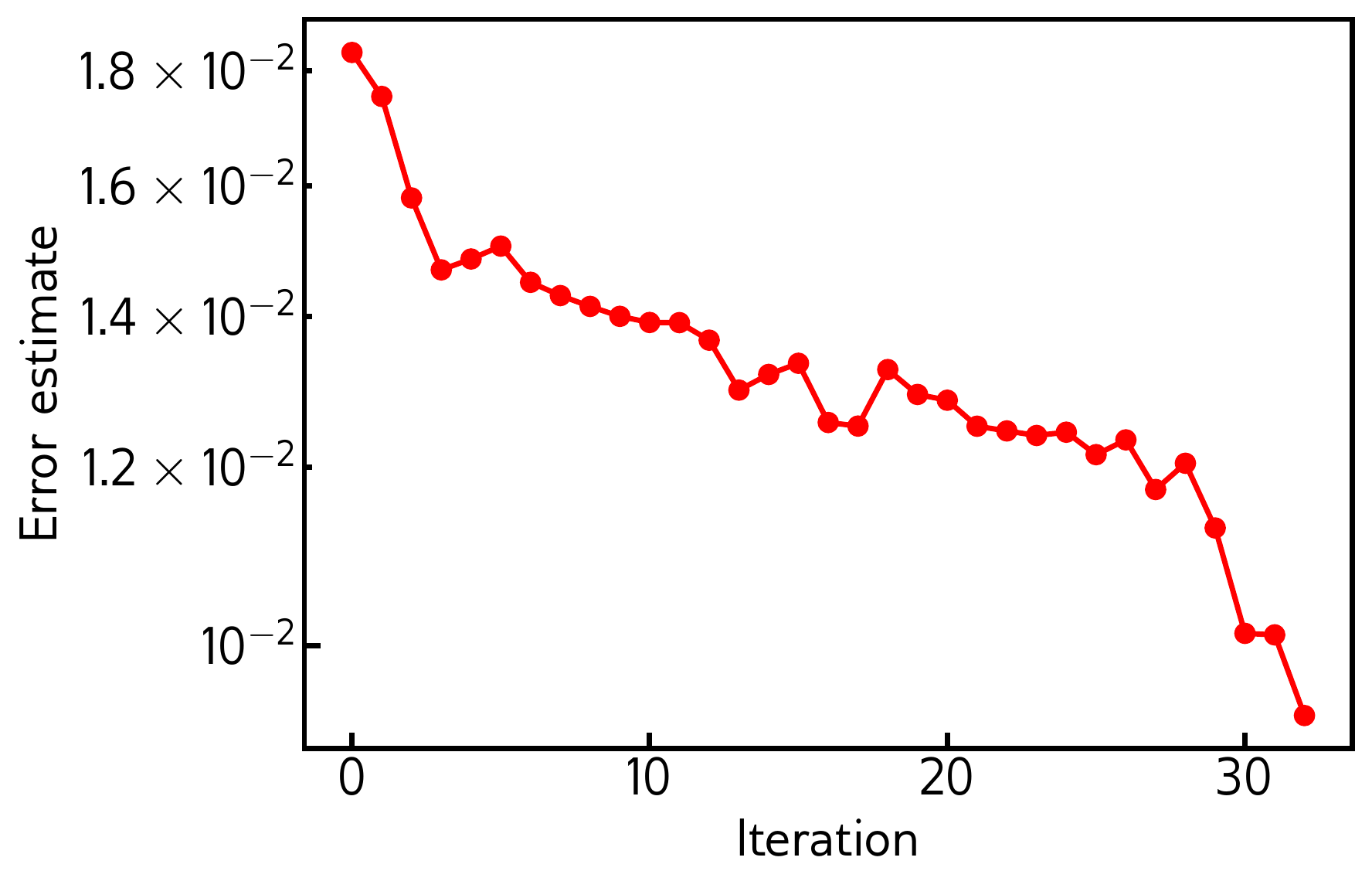}
		\caption{}
		\label{fig:burgers-active-learn-a}
	\end{subfigure}
	\hspace{2em}
	\begin{subfigure}[b]{0.35\textwidth}
		\centering
		\includegraphics[width=1\textwidth, trim=0 0 0 0, clip]{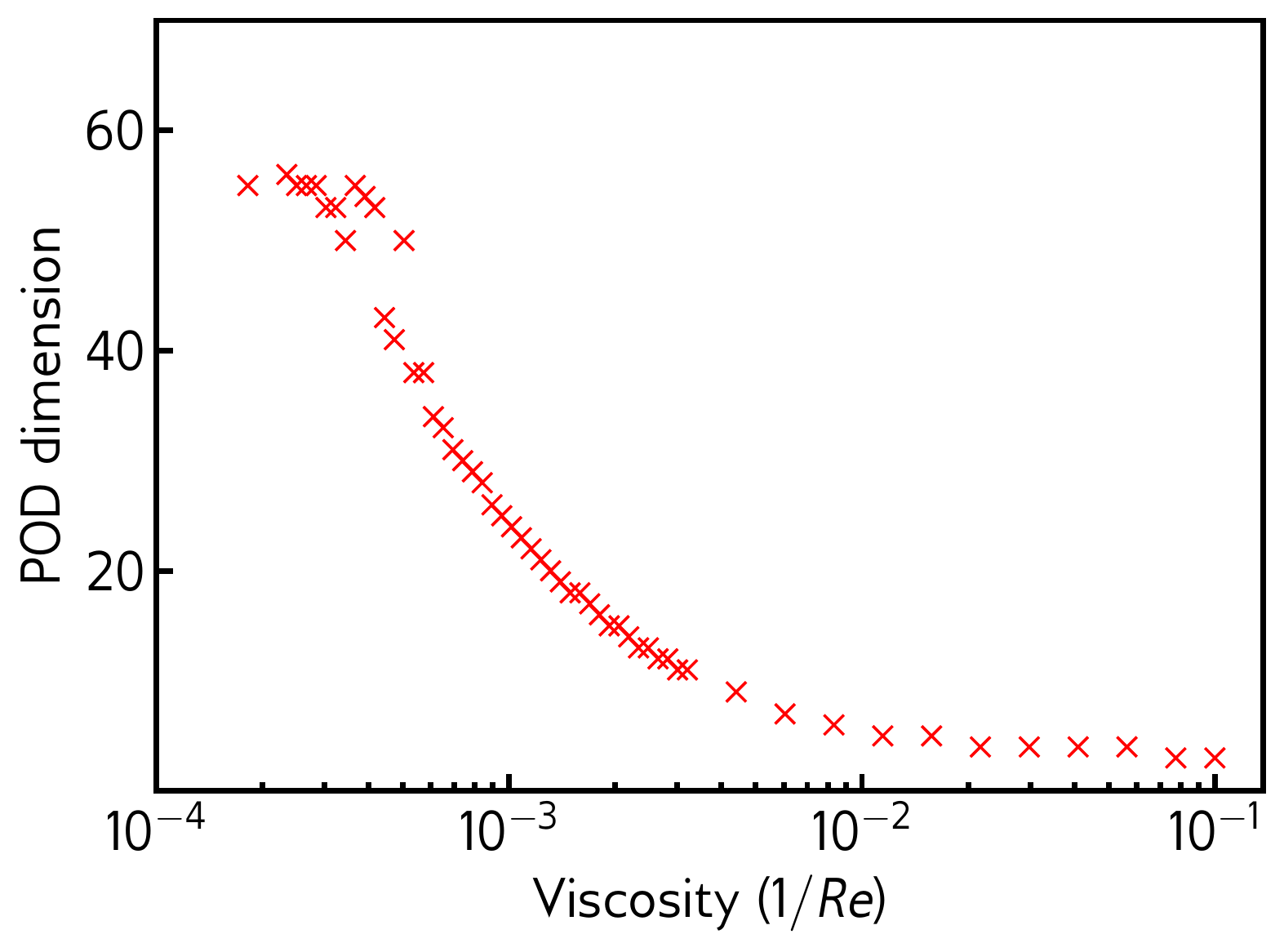}
		\caption{}
		\label{fig:burgers-active-learn-b}
	\end{subfigure}
	\caption{Active learning for Burgers' equation: (a)~shows the optimality criterion $\mathcal{E}^{(iter)}$ (refer \Cref{alg:summary-offline}) varying with greedy iterations of the active learning process; (b)~shows the final POD subspace dimension for each of the chosen parameter samples.}
	\label{fig:burgers-active-learn}
\end{figure}

\begin{figure}[H] 
	\centering
	\begin{subfigure}[b]{0.32\textwidth}
		\centering
		\includegraphics[width=0.9\textwidth, trim=0 0 0 200, clip]{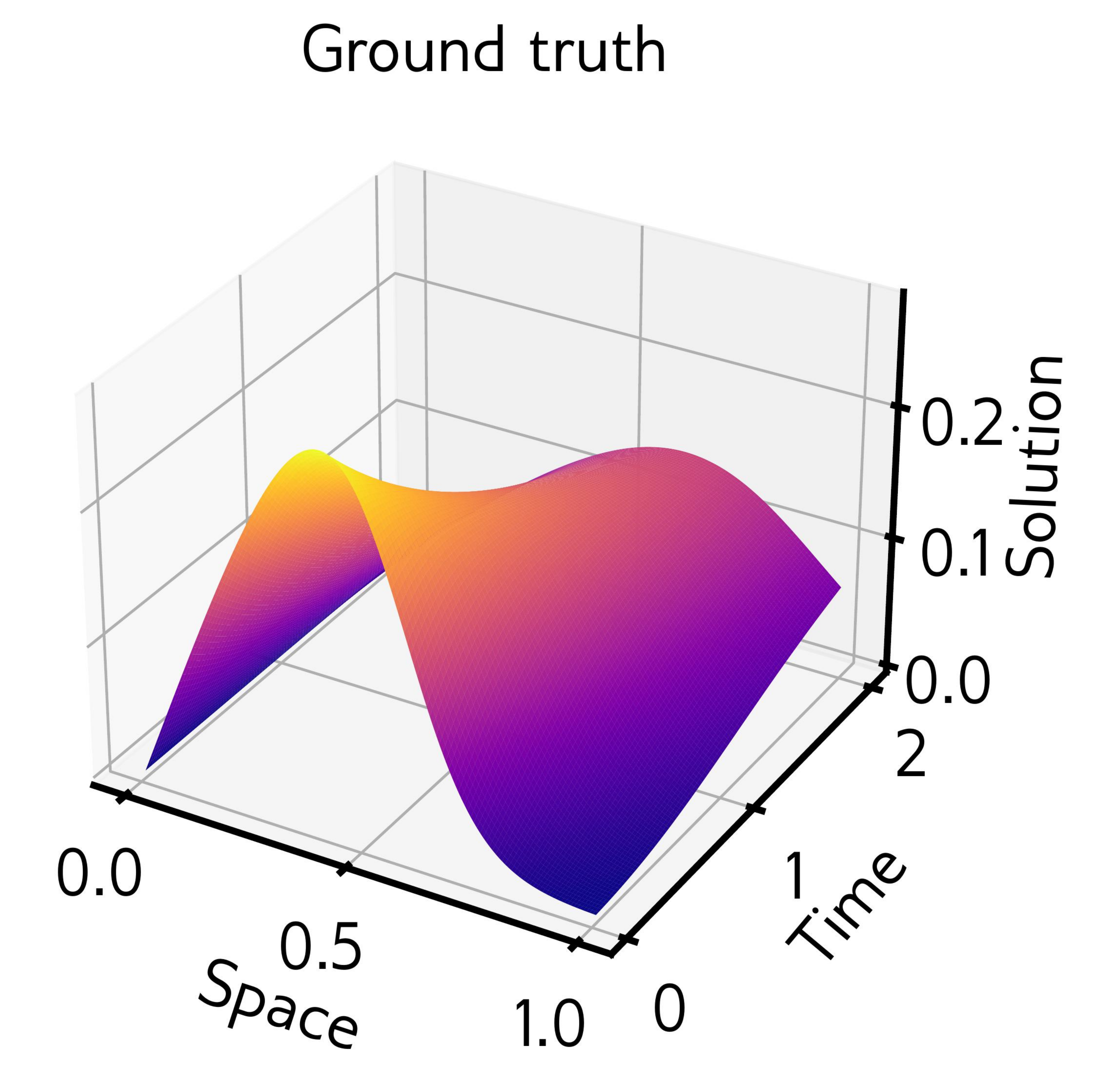}
	\end{subfigure}
	\begin{subfigure}[b]{0.32\textwidth}
		\centering
		\includegraphics[width=0.9\textwidth, trim=0 0 0 200, clip]{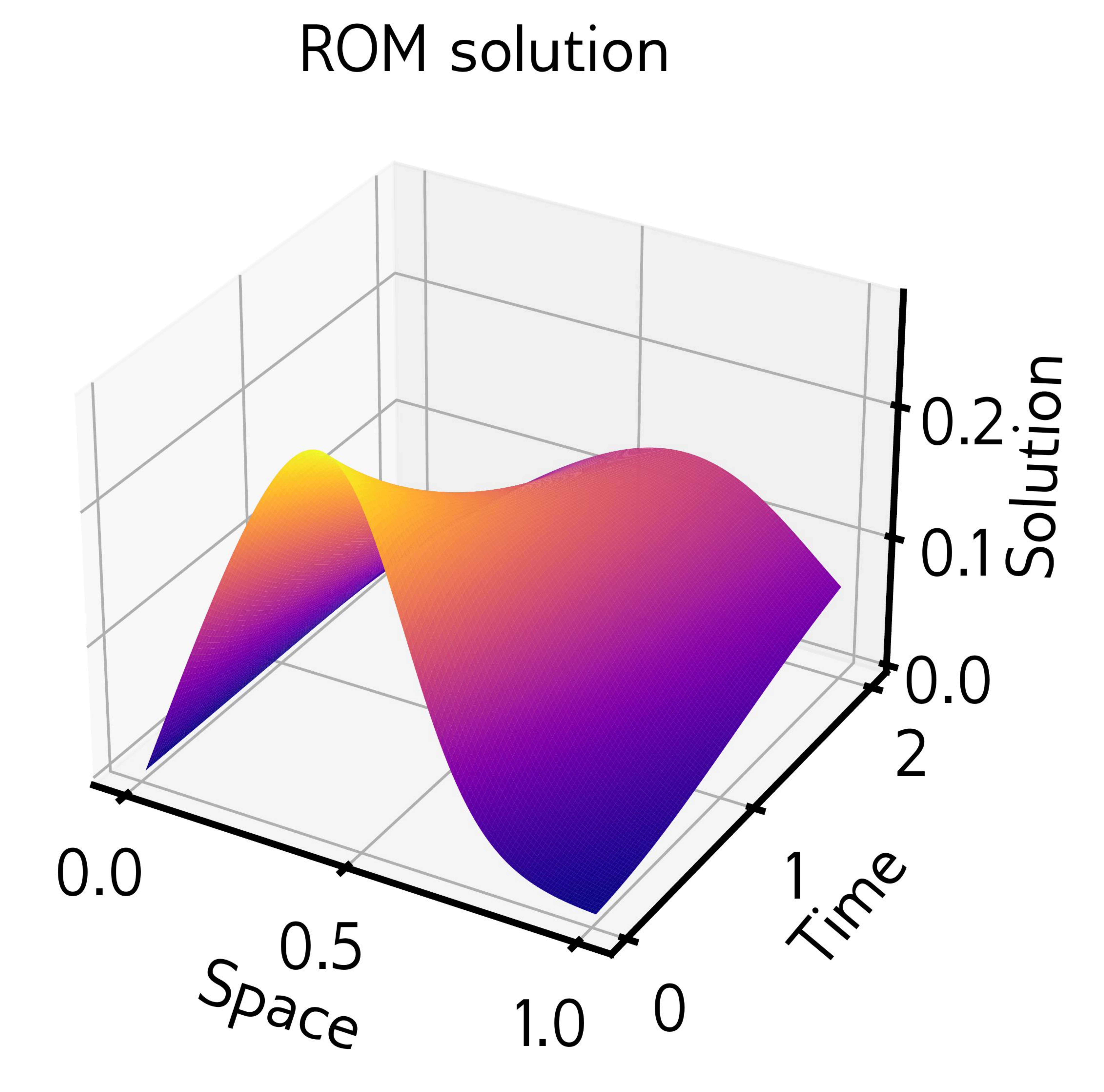}
	\end{subfigure}
	\begin{subfigure}[b]{0.32\textwidth}
		\centering
		\includegraphics[width=0.85\textwidth, trim=0 0 0 200, clip]{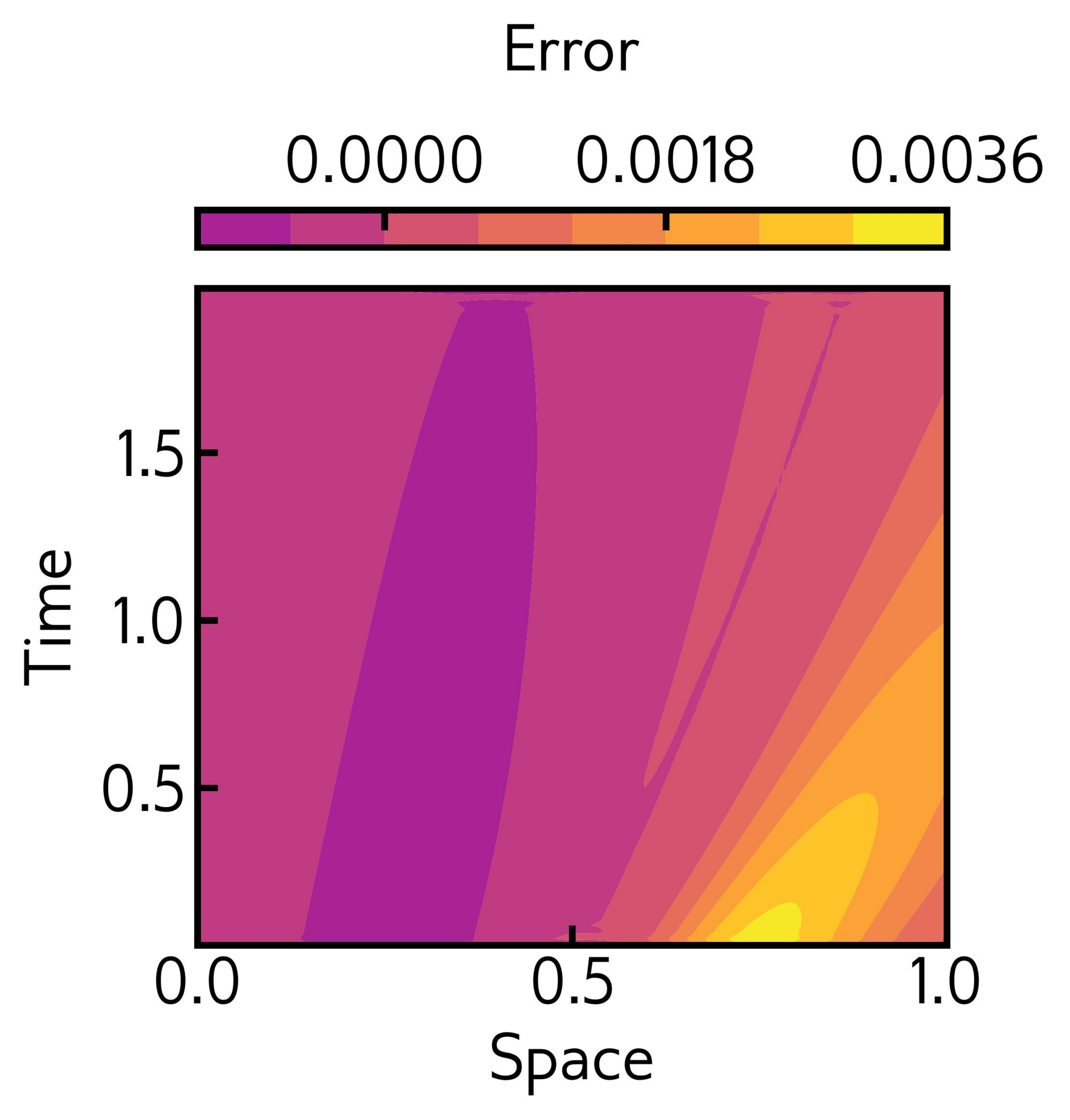}
	\end{subfigure}
	\begin{subfigure}[b]{0.32\textwidth}
		\centering
		\includegraphics[width=0.9\textwidth, trim=0 0 0 200, clip]{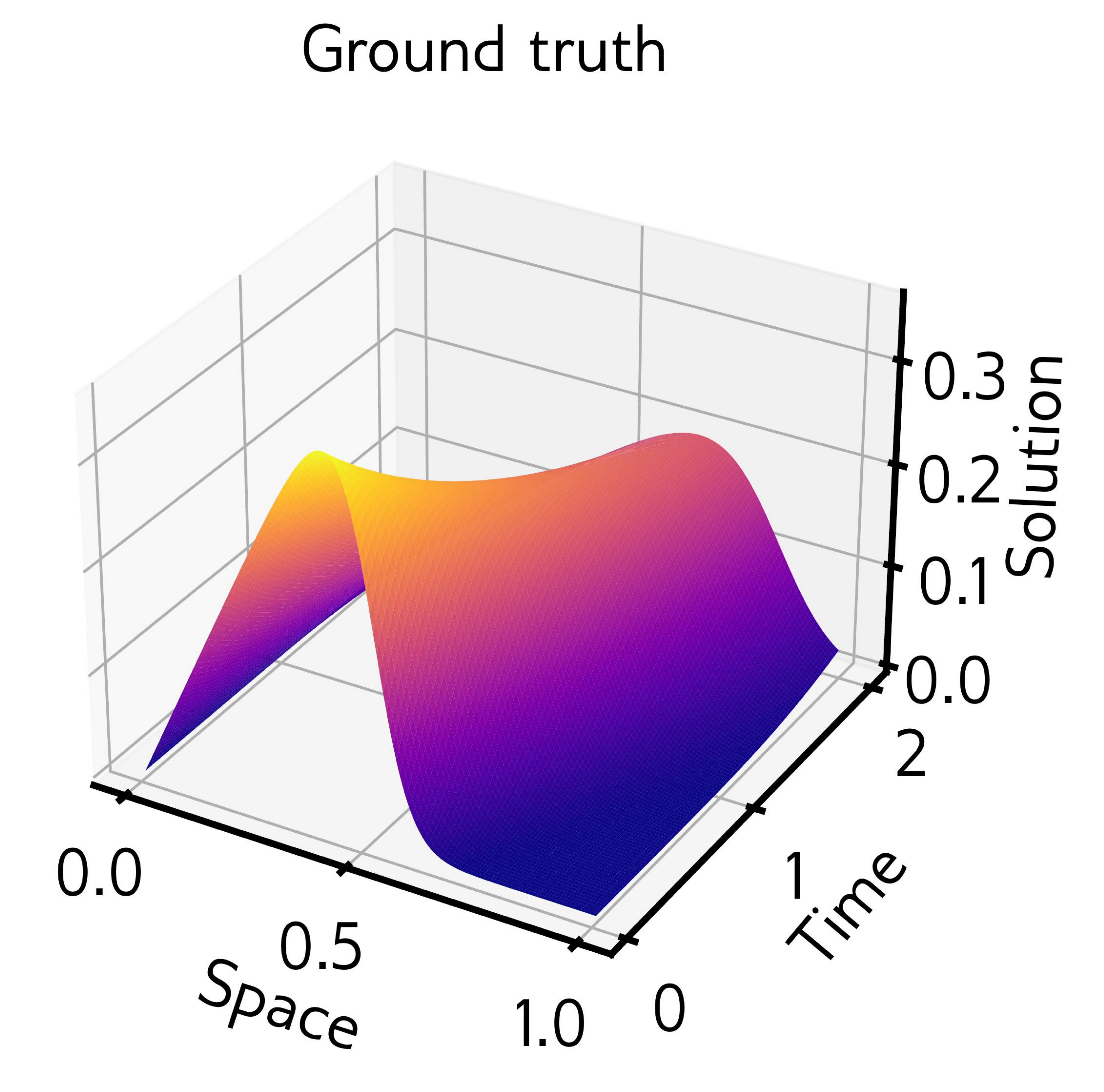}
	\end{subfigure}
	\begin{subfigure}[b]{0.32\textwidth}
		\centering
		\includegraphics[width=0.9\textwidth, trim=0 0 0 200, clip]{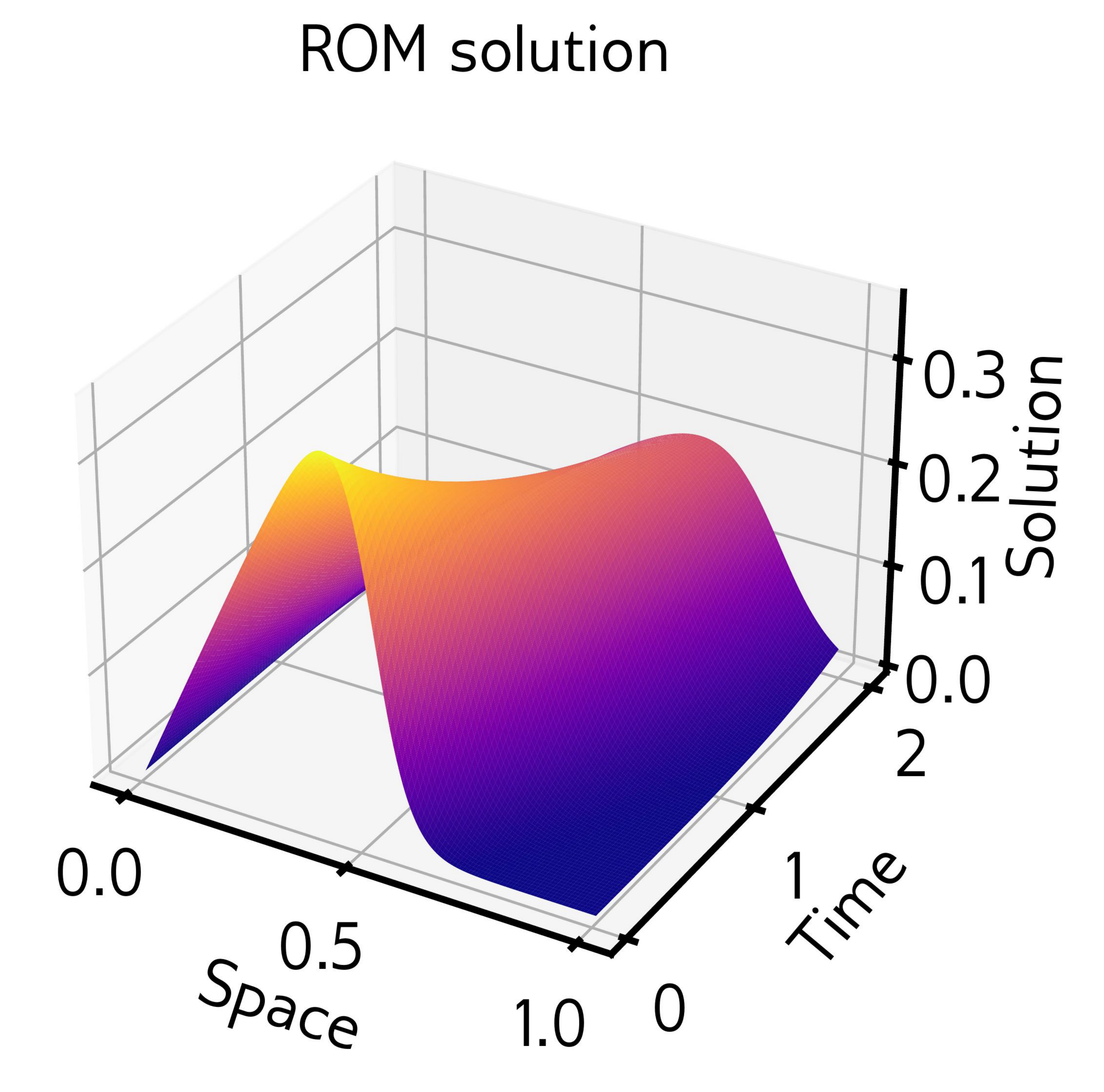}
	\end{subfigure}
	\begin{subfigure}[b]{0.32\textwidth}
		\centering
		\includegraphics[width=0.85\textwidth, trim=0 0 0 200, clip]{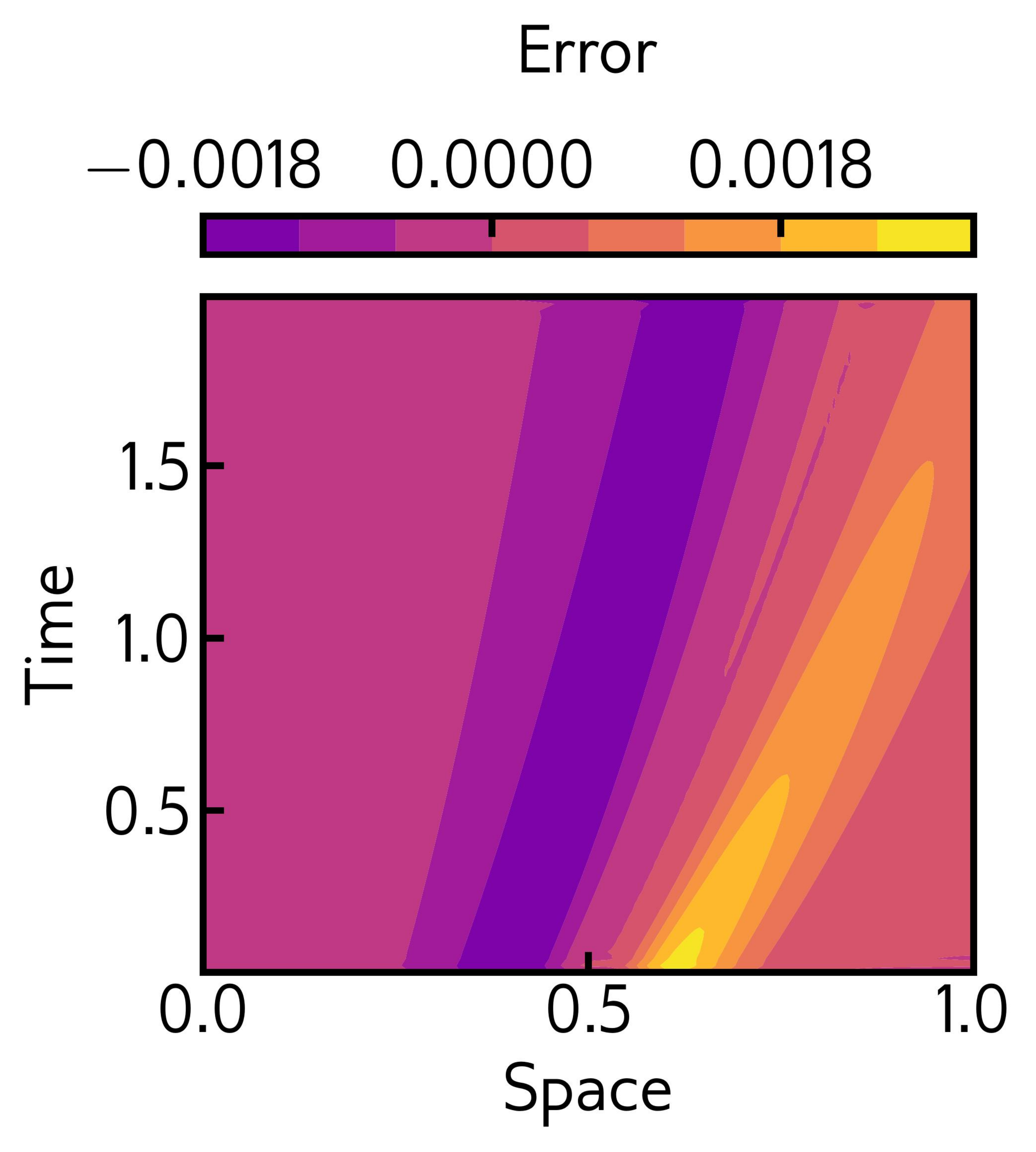}
	\end{subfigure}
	\begin{subfigure}[b]{0.32\textwidth}
		\centering
		\includegraphics[width=0.9\textwidth, trim=0 0 0 200, clip]{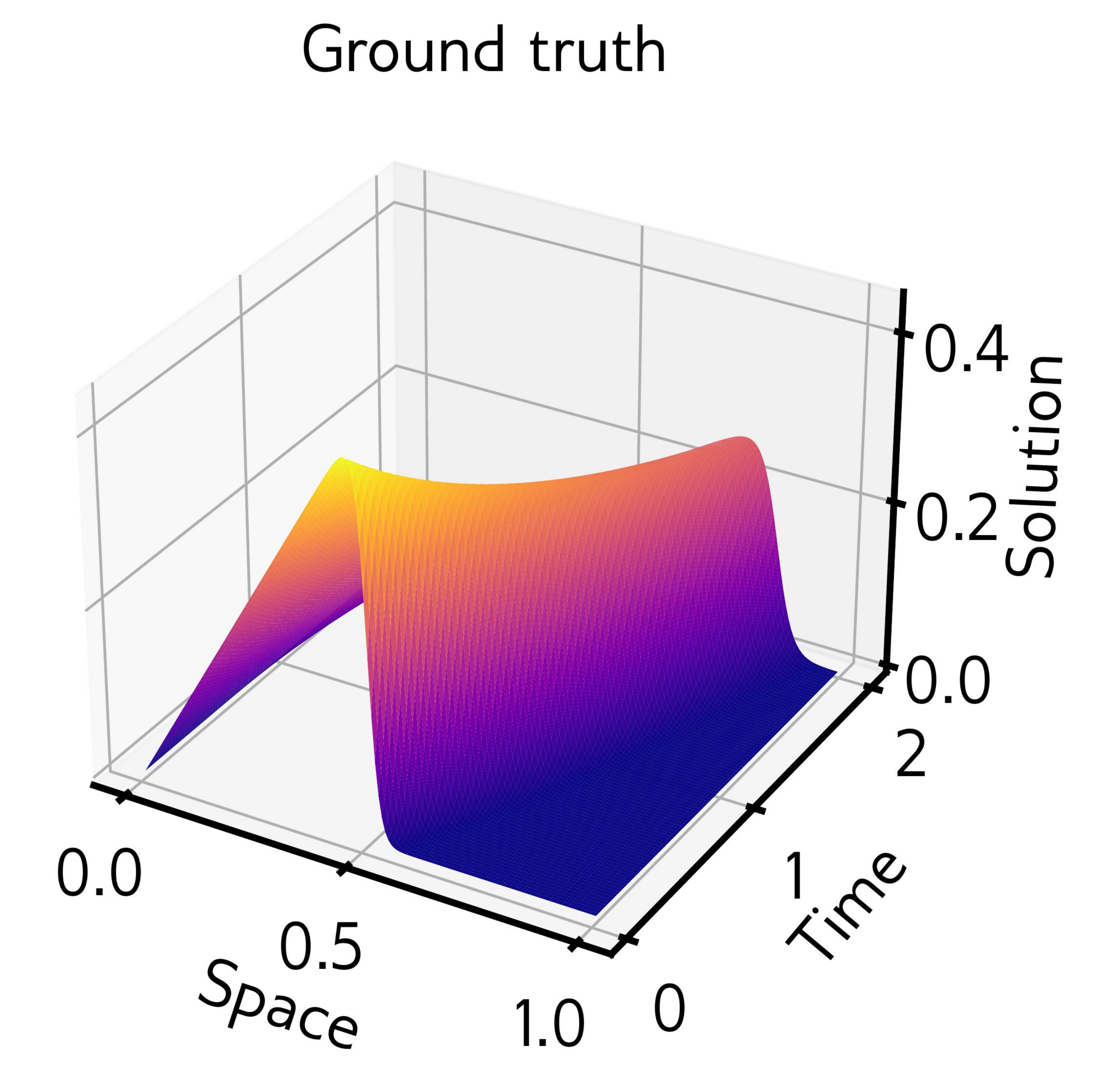}
	\end{subfigure}
	\begin{subfigure}[b]{0.32\textwidth}
		\centering
		\includegraphics[width=0.9\textwidth, trim=0 0 0 200, clip]{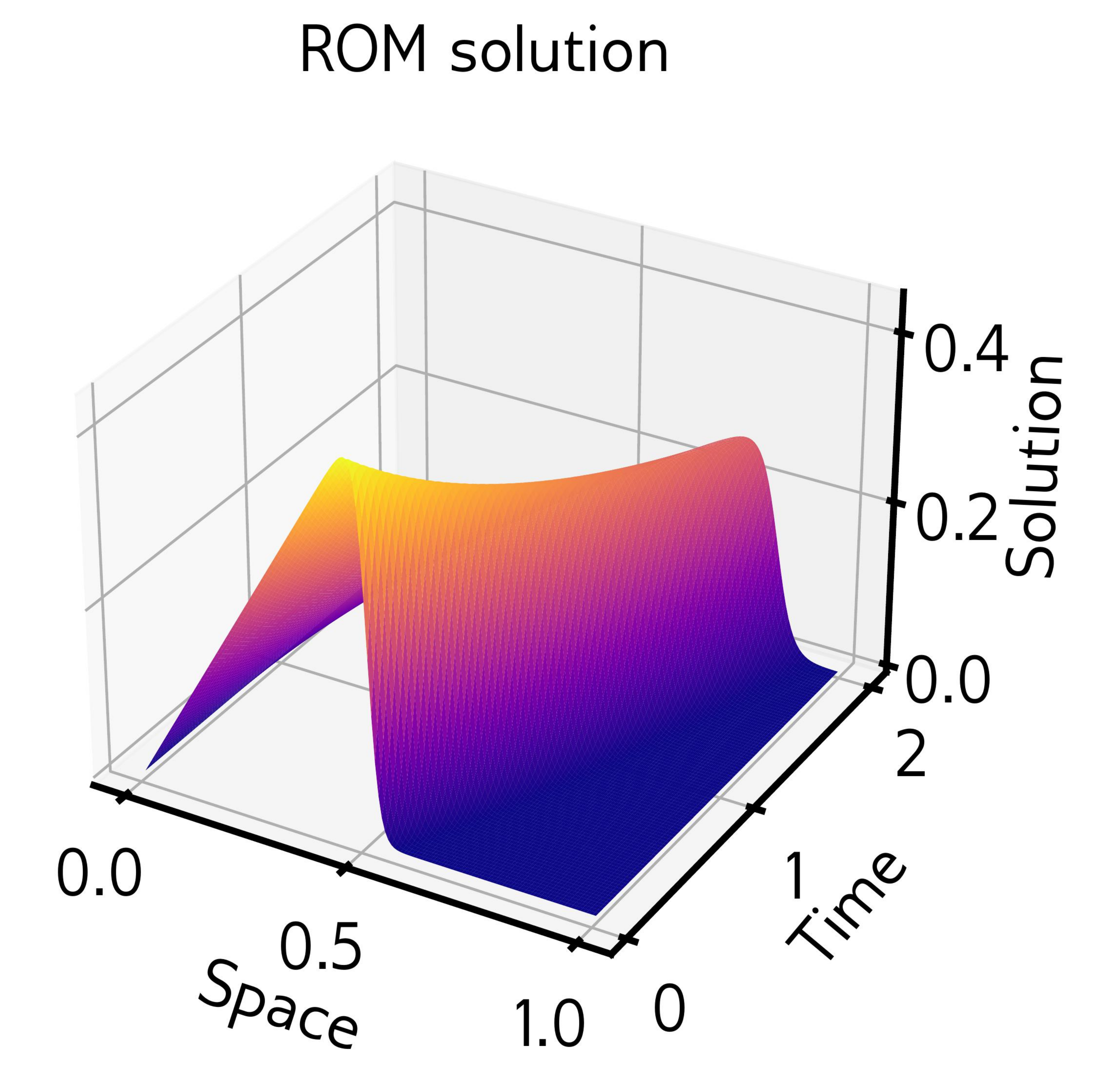}
	\end{subfigure}
	\begin{subfigure}[b]{0.32\textwidth}
		\centering
		\includegraphics[width=0.85\textwidth, trim=0 0 0 200, clip]{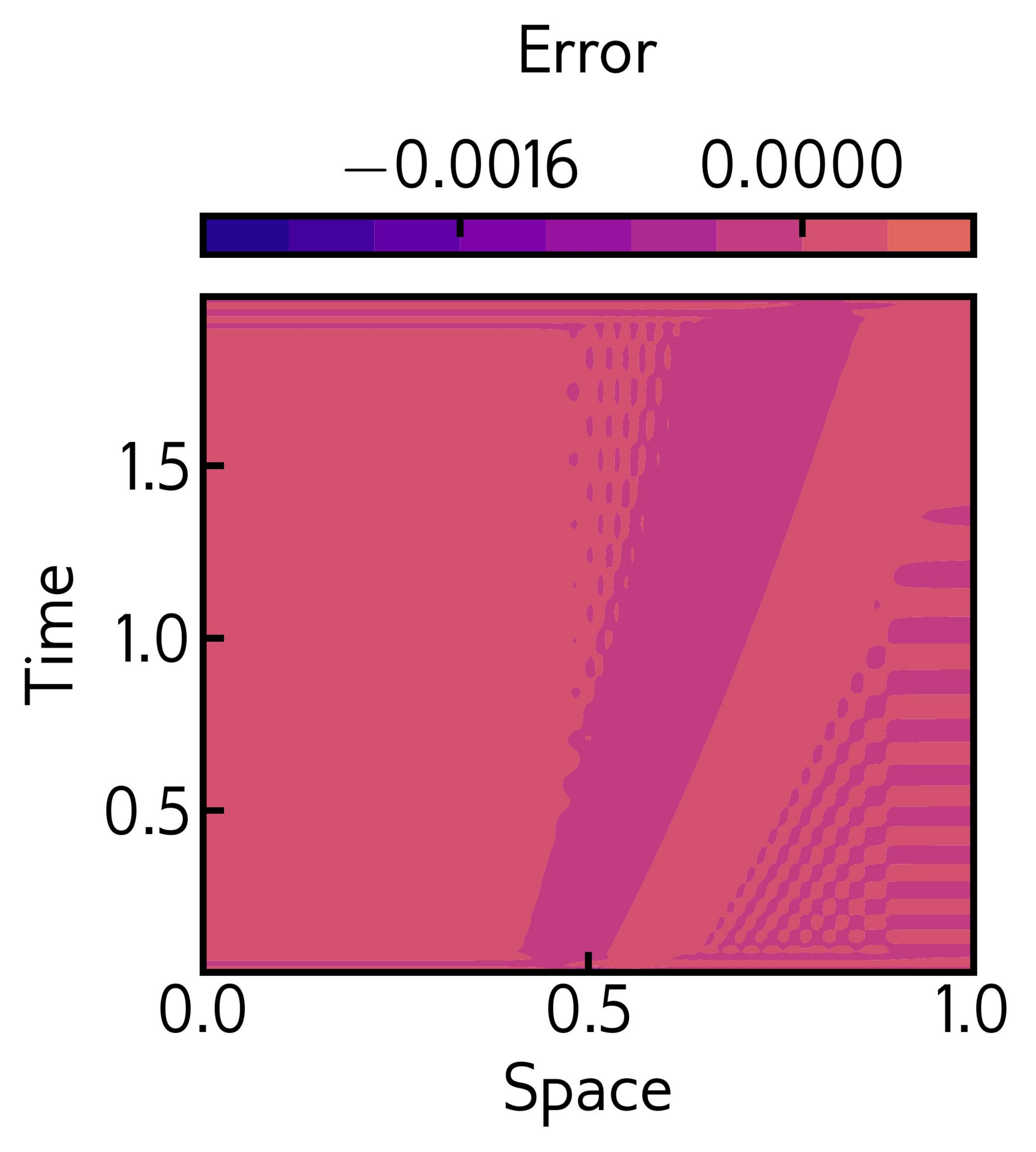}
	\end{subfigure}
	\begin{subfigure}[b]{0.32\textwidth}
		\centering
		\includegraphics[width=0.9\textwidth, trim=0 0 0 200, clip]{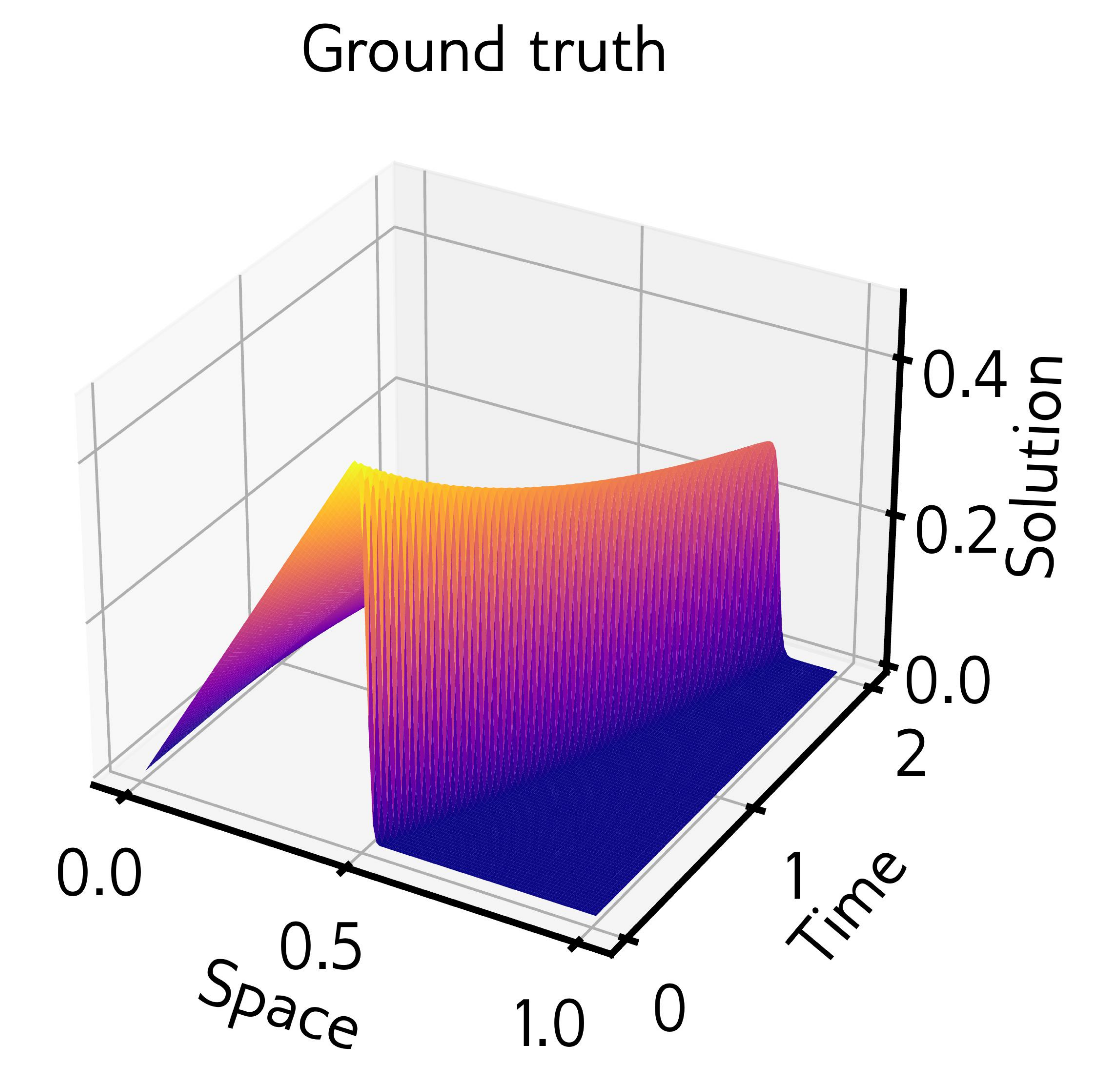}
	\end{subfigure}
	\begin{subfigure}[b]{0.32\textwidth}
		\centering
		\includegraphics[width=0.9\textwidth, trim=0 0 0 200, clip]{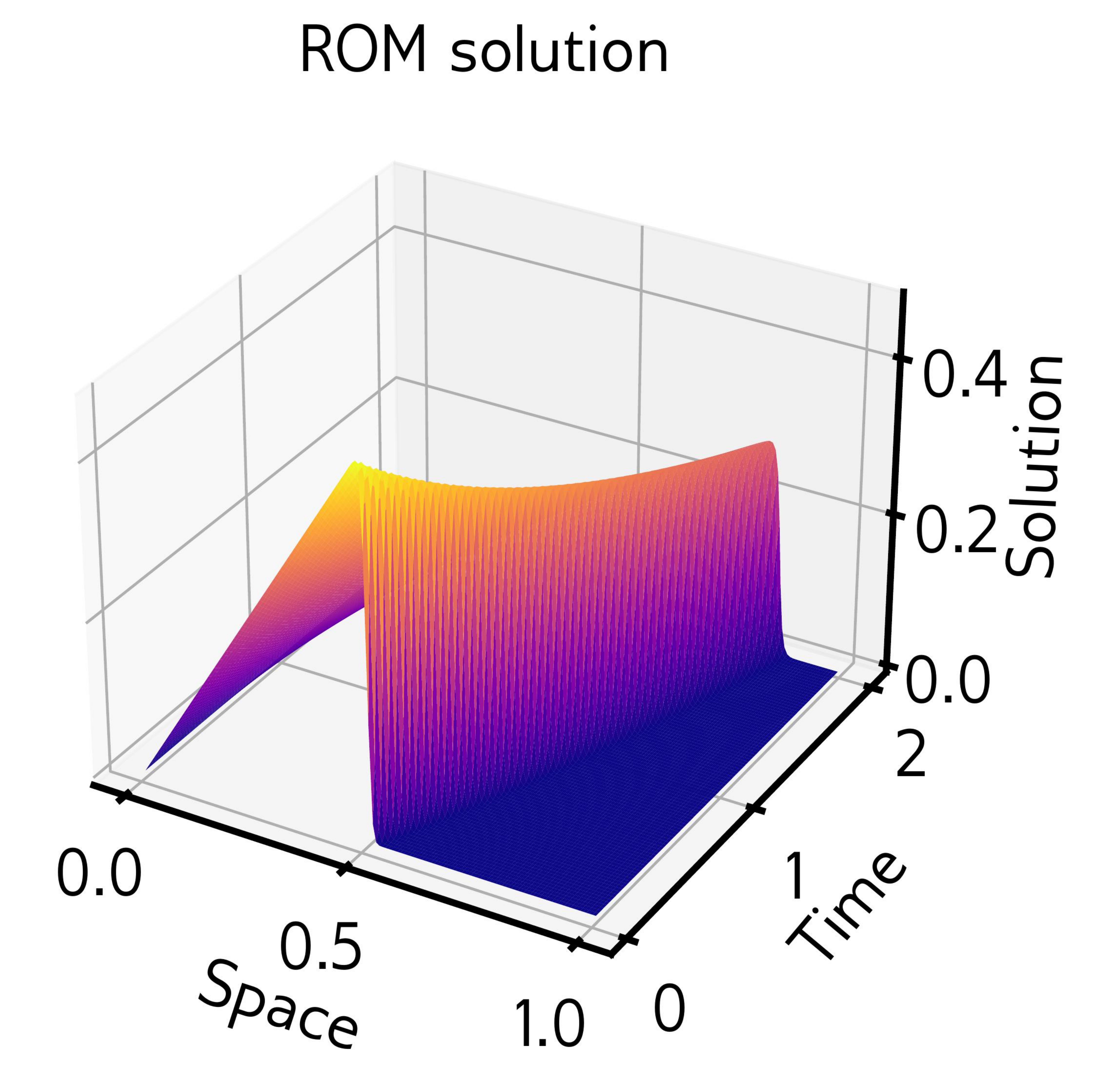}
	\end{subfigure}
	\begin{subfigure}[b]{0.32\textwidth}
		\centering
		\includegraphics[width=0.85\textwidth, trim=0 0 0 200, clip]{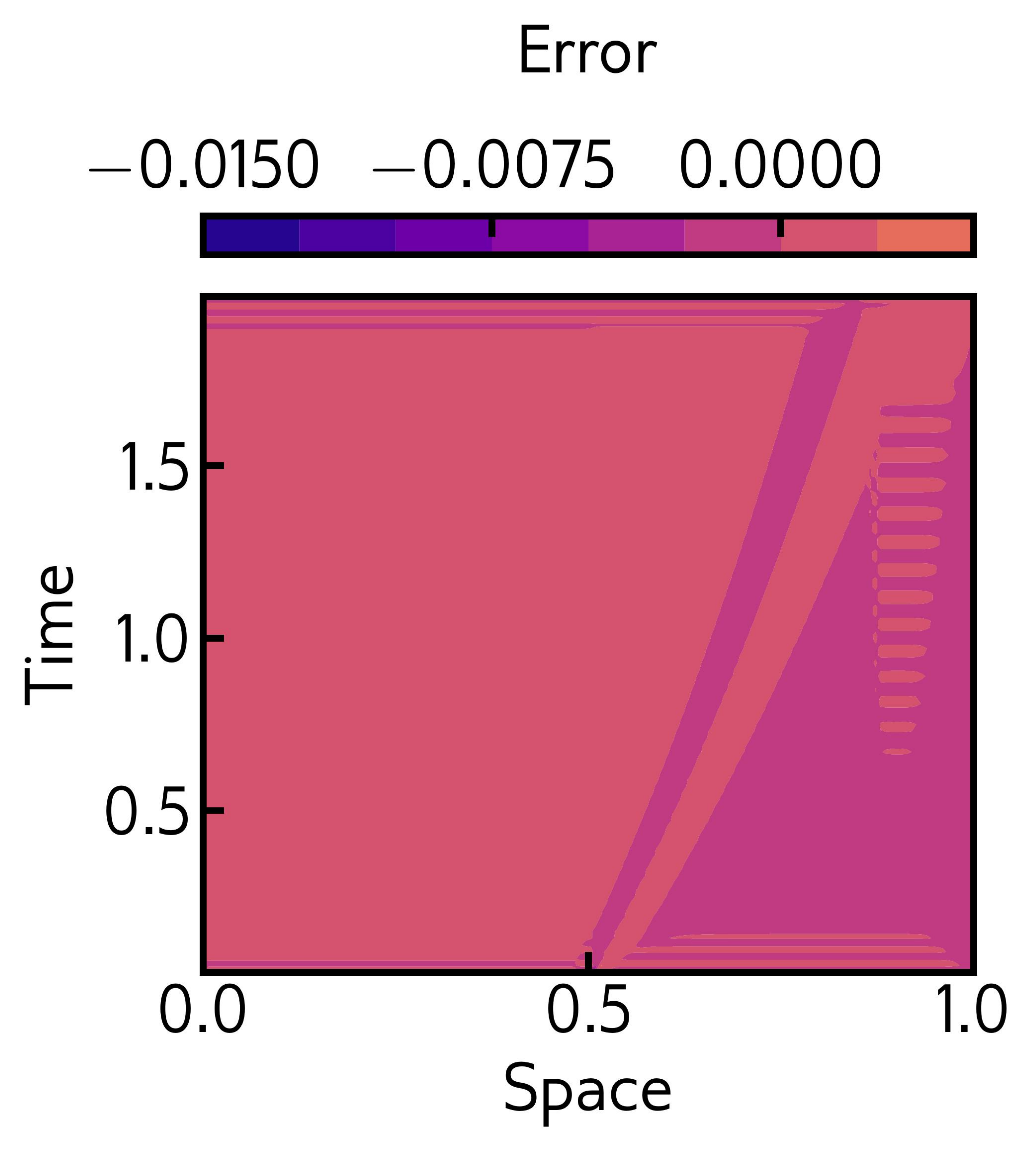}
	\end{subfigure}
	\begin{subfigure}[b]{0.32\textwidth}
		\centering
		\includegraphics[width=0.9\textwidth, trim=0 0 0 200, clip]{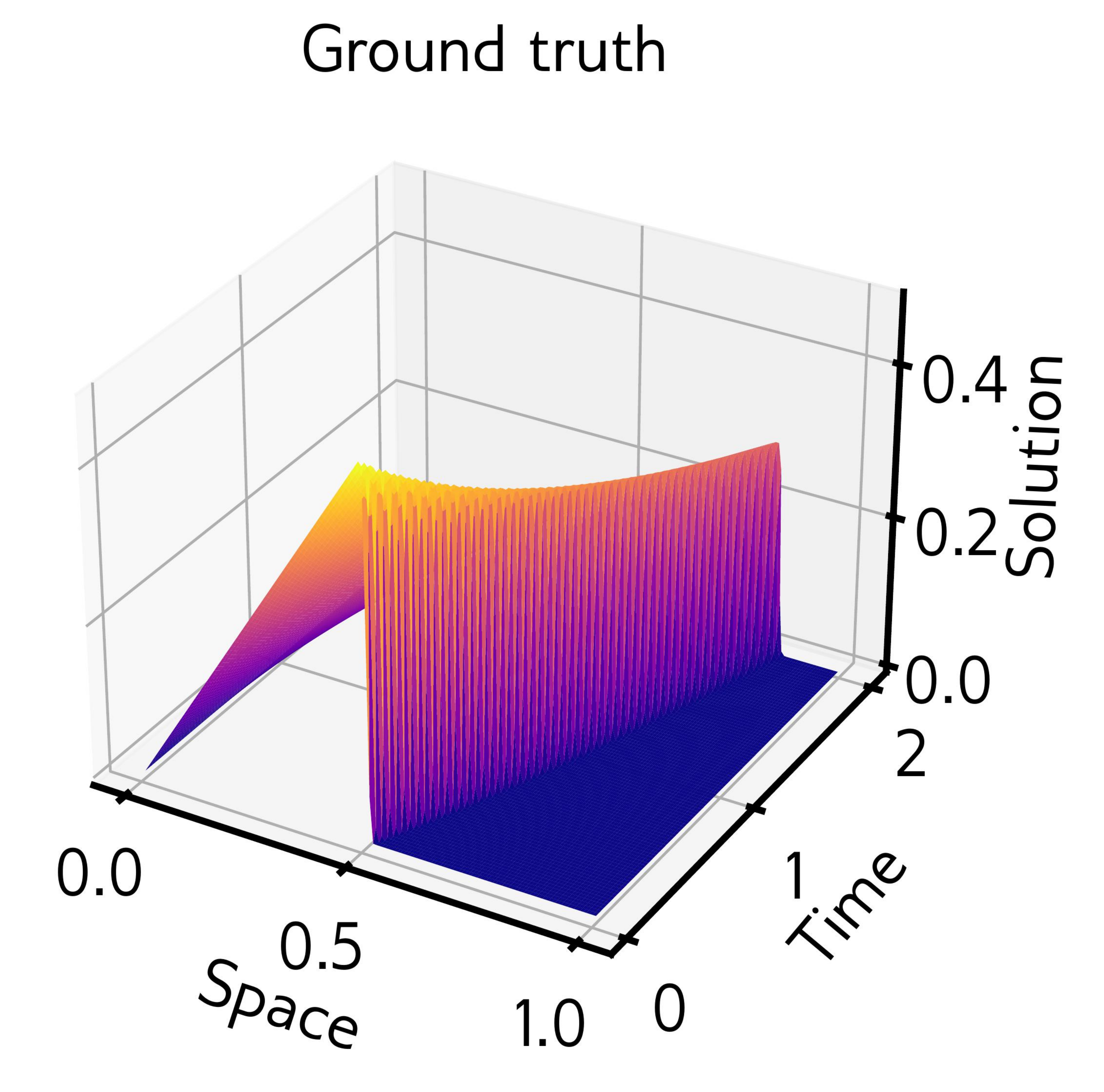}
	\end{subfigure}
	\begin{subfigure}[b]{0.32\textwidth}
		\centering
		\includegraphics[width=0.9\textwidth, trim=0 0 0 200, clip]{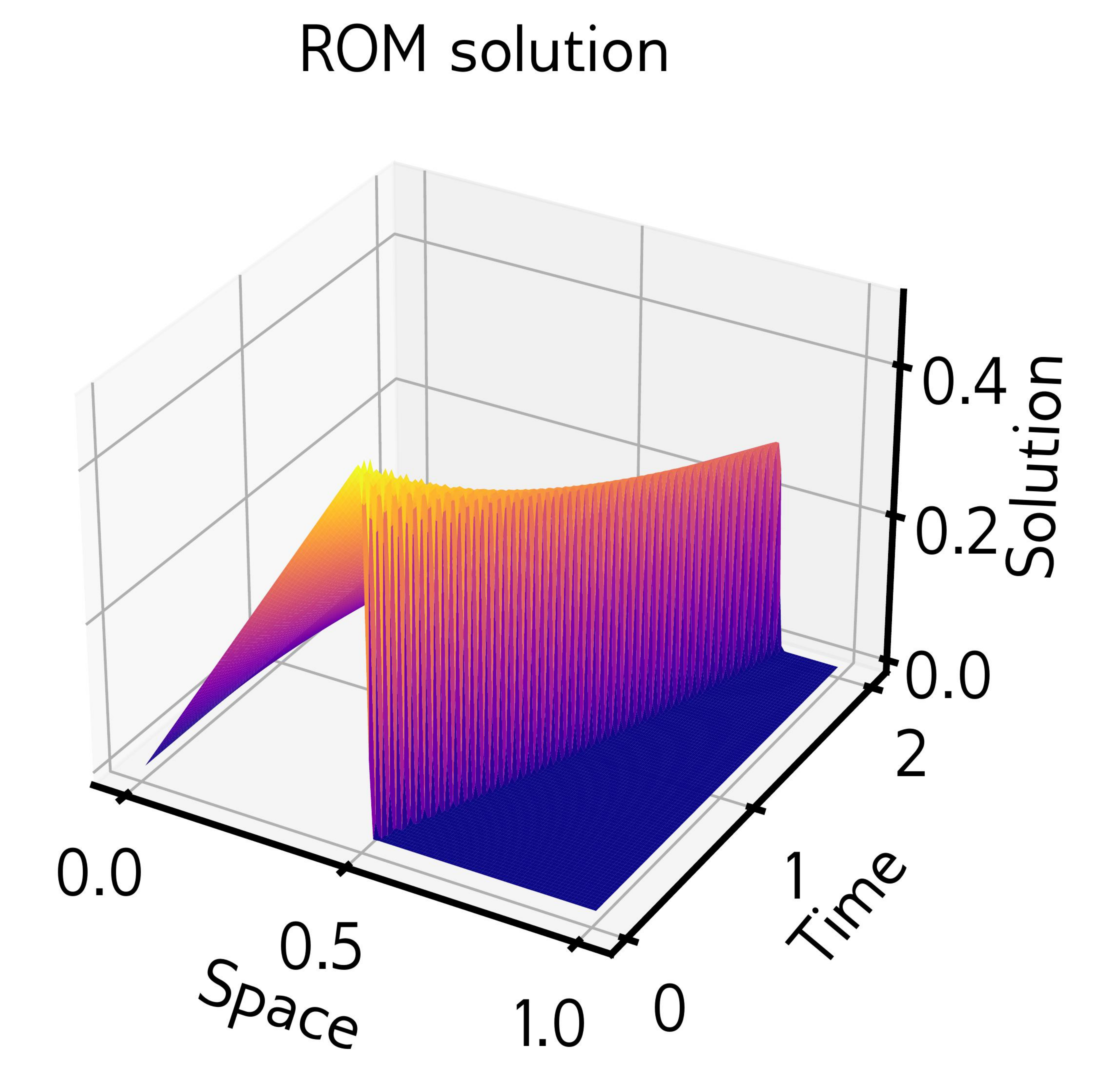}
	\end{subfigure}
	\begin{subfigure}[b]{0.32\textwidth}
		\centering
		\includegraphics[width=0.85\textwidth, trim=0 0 0 200, clip]{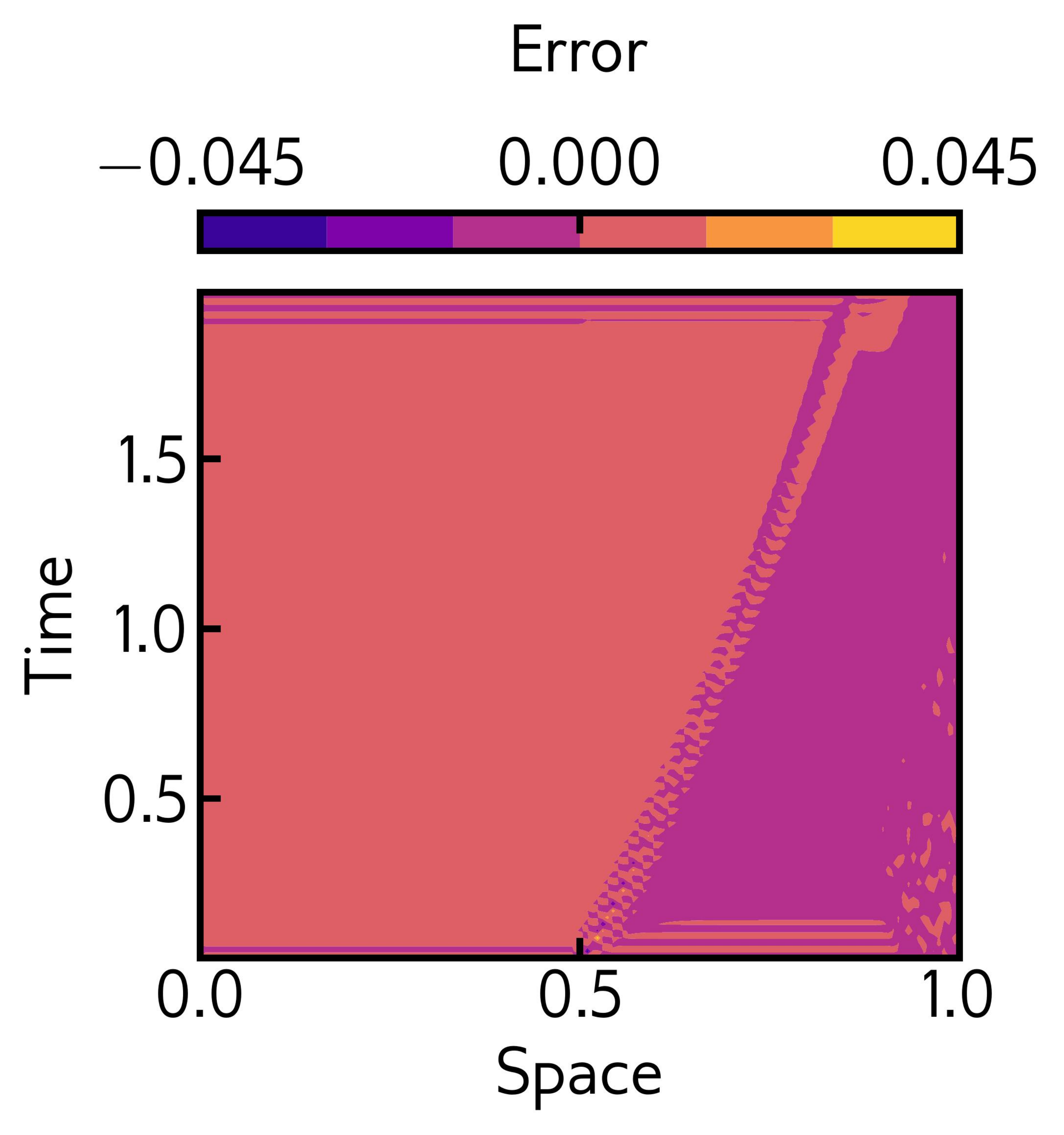}
	\end{subfigure}
	\caption{Burgers' equation: The true solution (shown in first column), ActLearn-POD-KSNN solution (shown in second column), and the solution error (shown in third column). The error values correspond to the point-wise difference in the space-time domain between the ActLearn-POD-KSNN solution and the true solution. The $Re$ values going from top to bottom in the rows are in the following order: $\{40, 100, 350, 1250, 3000\}$. All these Re values and discrete time instances required to generate the plots are outside of the training set.}
	\label{fig:burgers-rom-sol-surf}
\end{figure}

\pagebreak

\begin{figure}[!t] 
	\centering
	\begin{subfigure}[b]{0.32\textwidth}
		\centering
		\includegraphics[width=1\textwidth, trim=0 0 0 0, clip]{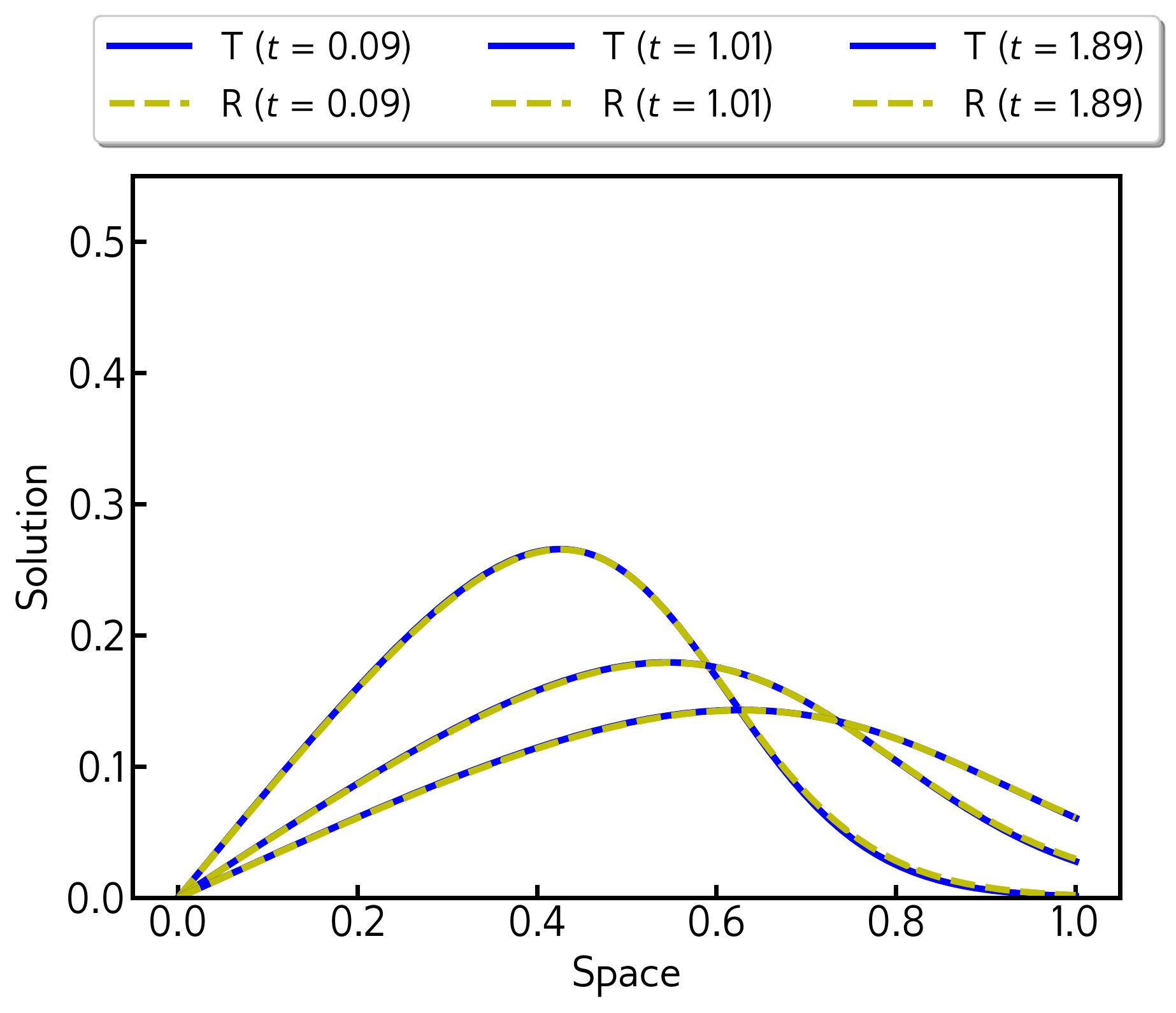}
		\caption{$Re = 40$}
		\label{fig:burgers-rom-sol-a}
	\end{subfigure}
	\begin{subfigure}[b]{0.32\textwidth}
		\centering
		\includegraphics[width=1\textwidth, trim=0 0 0 0, clip]{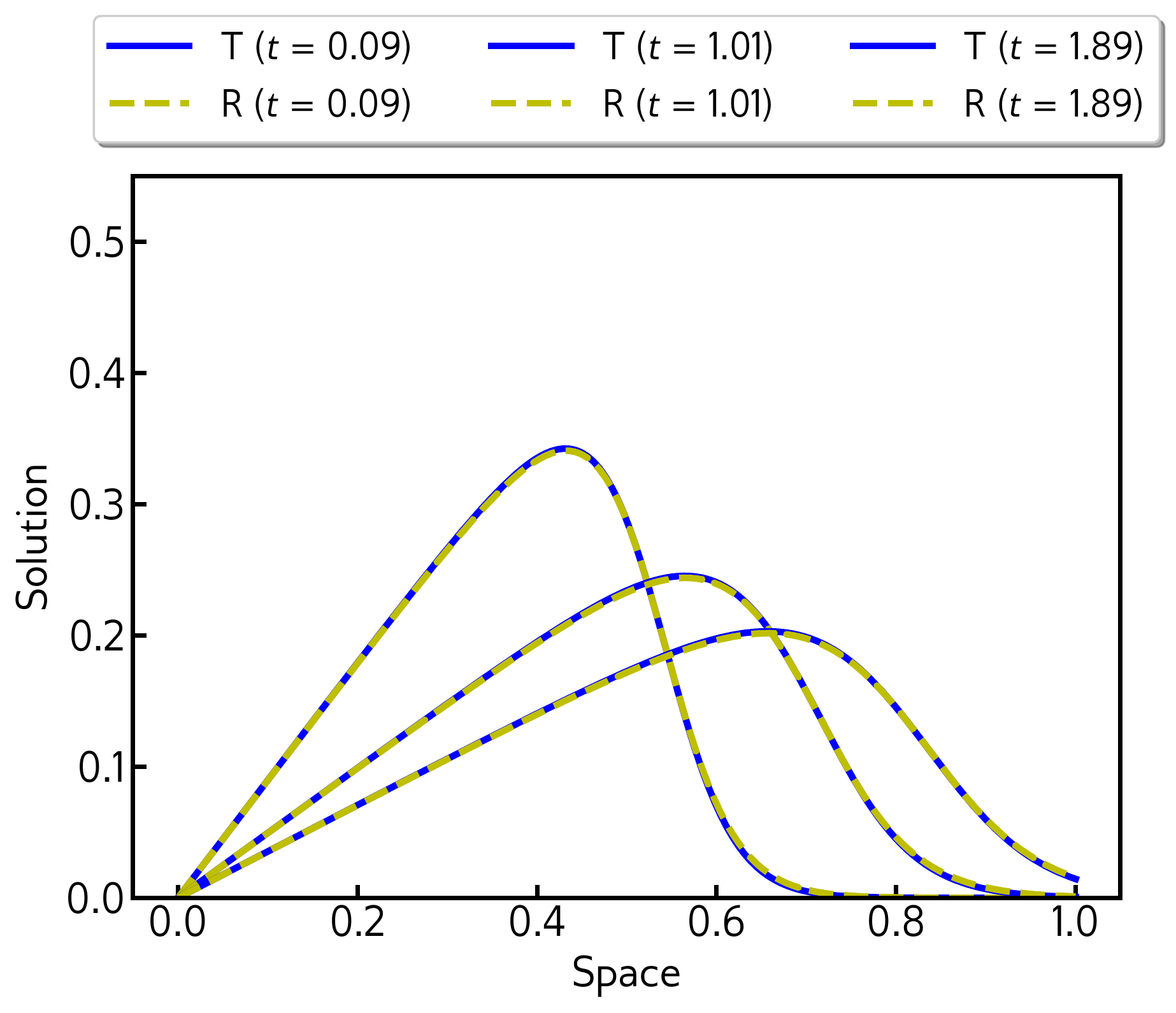}
		\caption{$Re = 100$}
		\label{fig:burgers-rom-sol-b}
	\end{subfigure}
	\begin{subfigure}[b]{0.32\textwidth}
		\centering
		\includegraphics[width=1\textwidth, trim=0 0 0 0, clip]{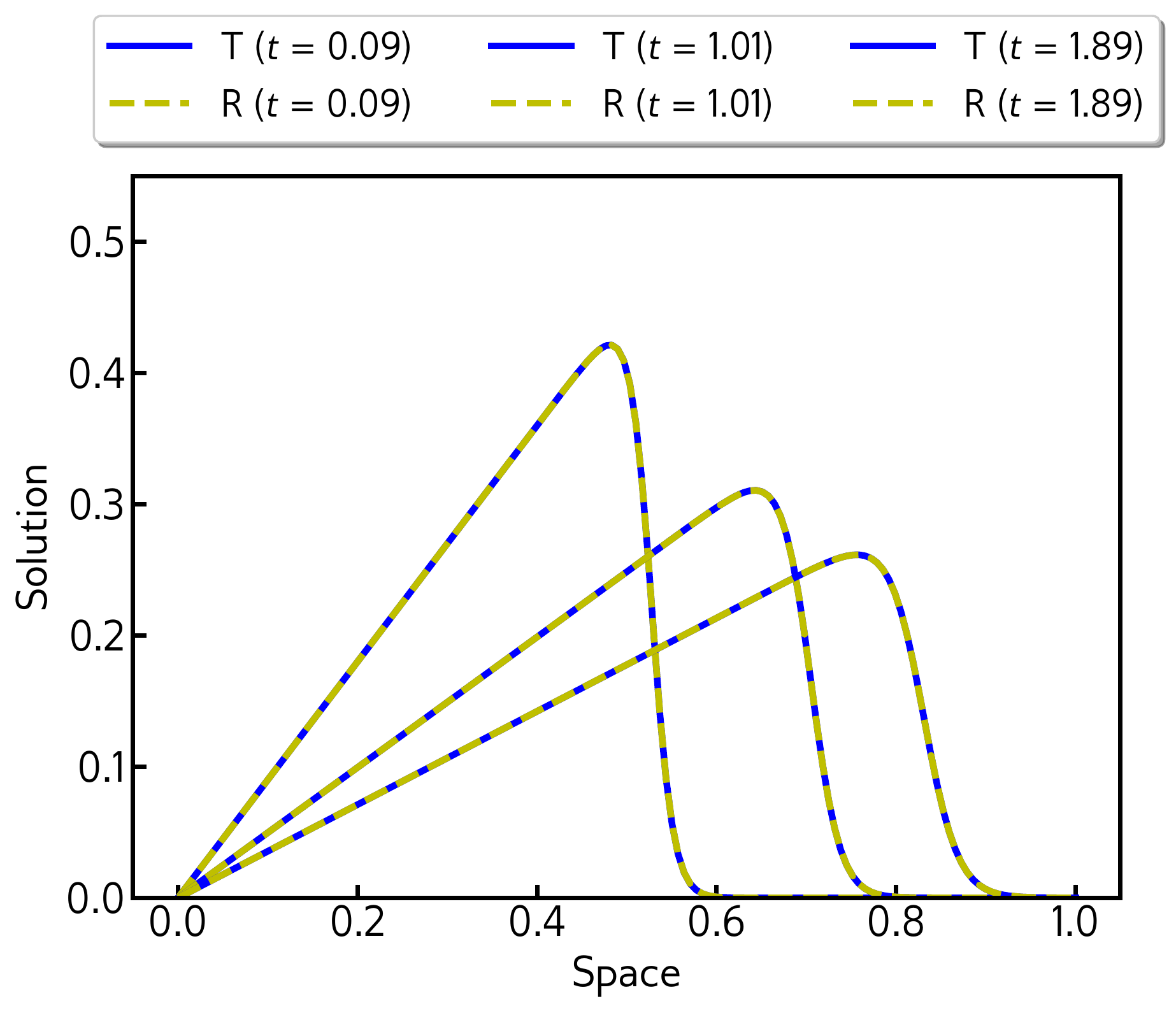}
		\caption{$Re = 350$}
		\label{fig:burgers-rom-sol-c}
	\end{subfigure}
	\begin{subfigure}[b]{0.32\textwidth}
		\centering
		\includegraphics[width=1\textwidth, trim=0 0 0 0, clip]{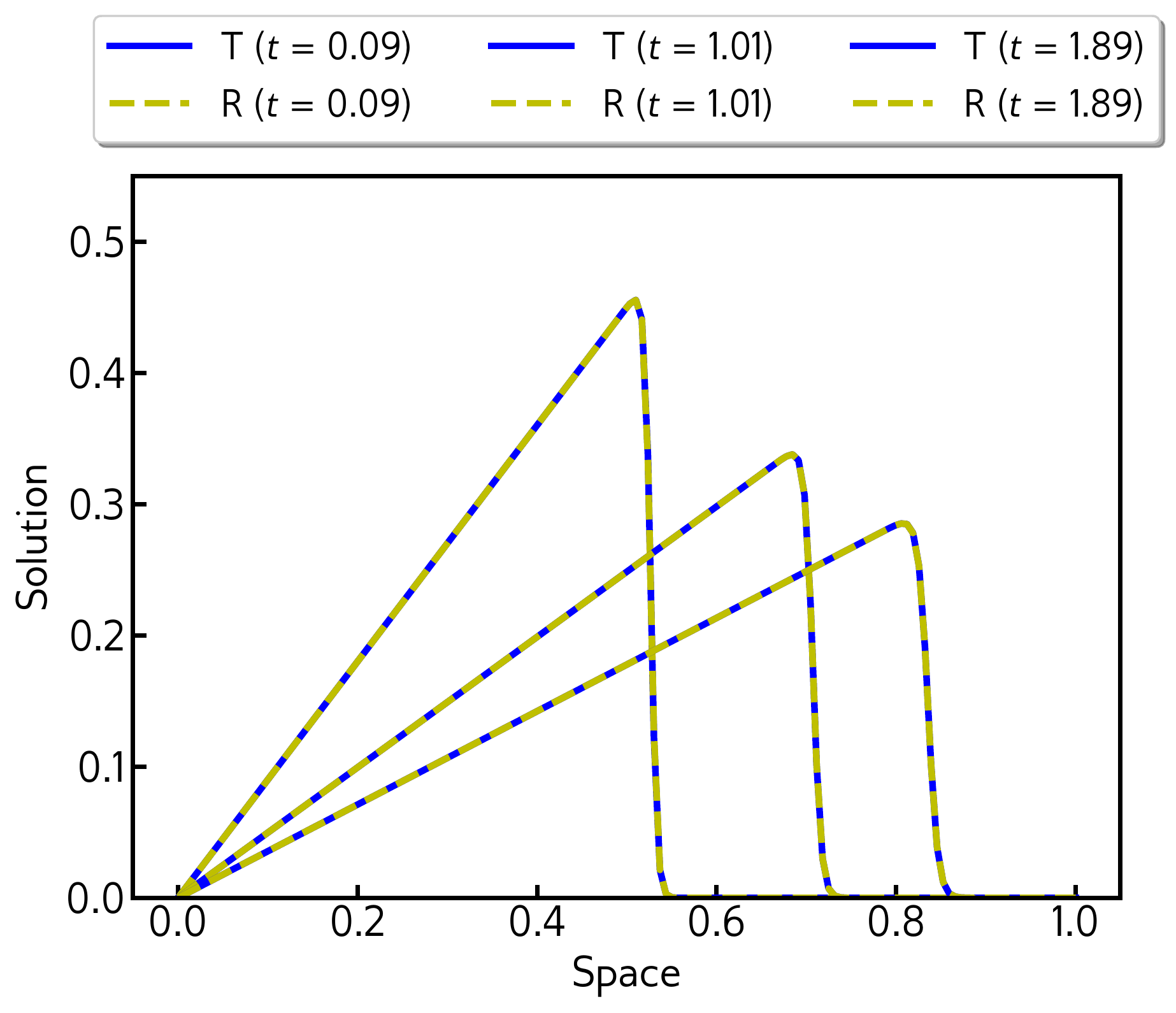}
		\caption{$Re = 1250$}
		\label{fig:burgers-rom-sol-d}
	\end{subfigure}
	\begin{subfigure}[b]{0.32\textwidth}
		\centering
		\includegraphics[width=1\textwidth, trim=0 0 0 0, clip]{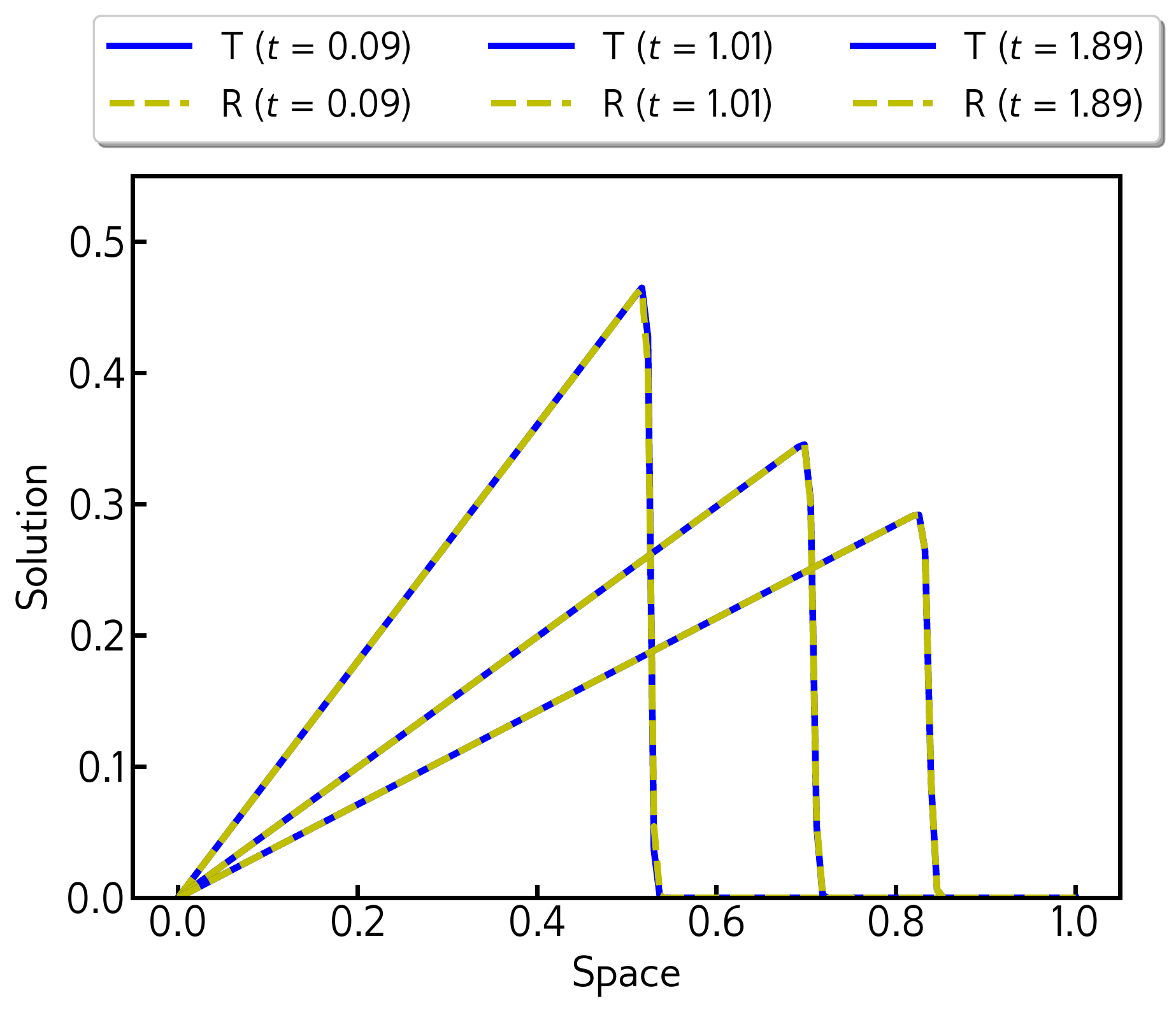}
		\caption{$Re = 3000$}
		\label{fig:burgers-rom-sol-e}
	\end{subfigure}
	\caption{Comparison of the ActLearn-POD-KSNN solution (denoted by R) and true solution (denoted by T) for Burgers' equation. All the Reynolds numbers and time instances~($t$) are outside of the training set.}
	\label{fig:burgers-rom-sol}
\end{figure}

To begin the active learning procedure, the parameter set $P$ is initiated by $21$ viscosity values corresponding to the following indices,
\begin{equation*}
	\{0,99,10,20,30,40,50,60,70,80,90,5,15,25,35,45,55,65,75,85,95\}.
\end{equation*}
Here, the $\nu$ values are ordered from lowest to highest and the index starts from $0$ when counting the $100$ values. We report the indices instead of exact values for ease of readability. High-fidelity snapshots are generated for all the viscosities in $P$ at $100$ instances of $t \in [0,2]$. Using these snapshots, POD-approximate solutions are computed such that the POD subspace retains $99.99 \%$ of the energy, i.e., $\eta(\nu) = 10^{-4}$, $\forall \nu \in P$.

\begin{figure}[!b] 
	\centering
	\begin{subfigure}[b]{0.47\textwidth}
		\centering
		\includegraphics[width=0.85\textwidth, trim=0 0 0 0, clip]{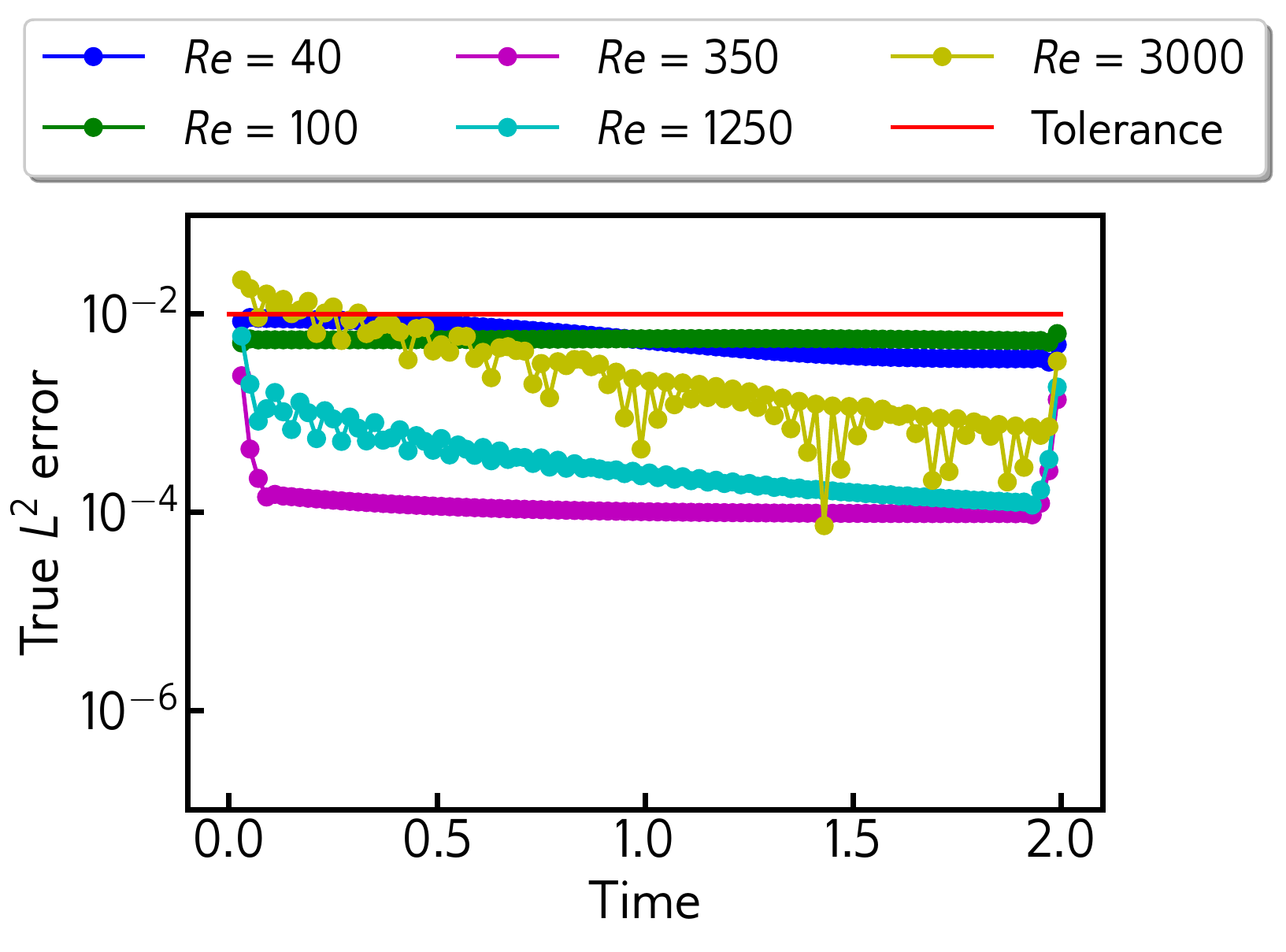}
		\caption{}
		\label{fig:burgers-error-plot-a}
	\end{subfigure}
	\begin{subfigure}[b]{0.47\textwidth}
		\centering
		\includegraphics[width=0.85\textwidth, trim=0 0 0 0, clip]{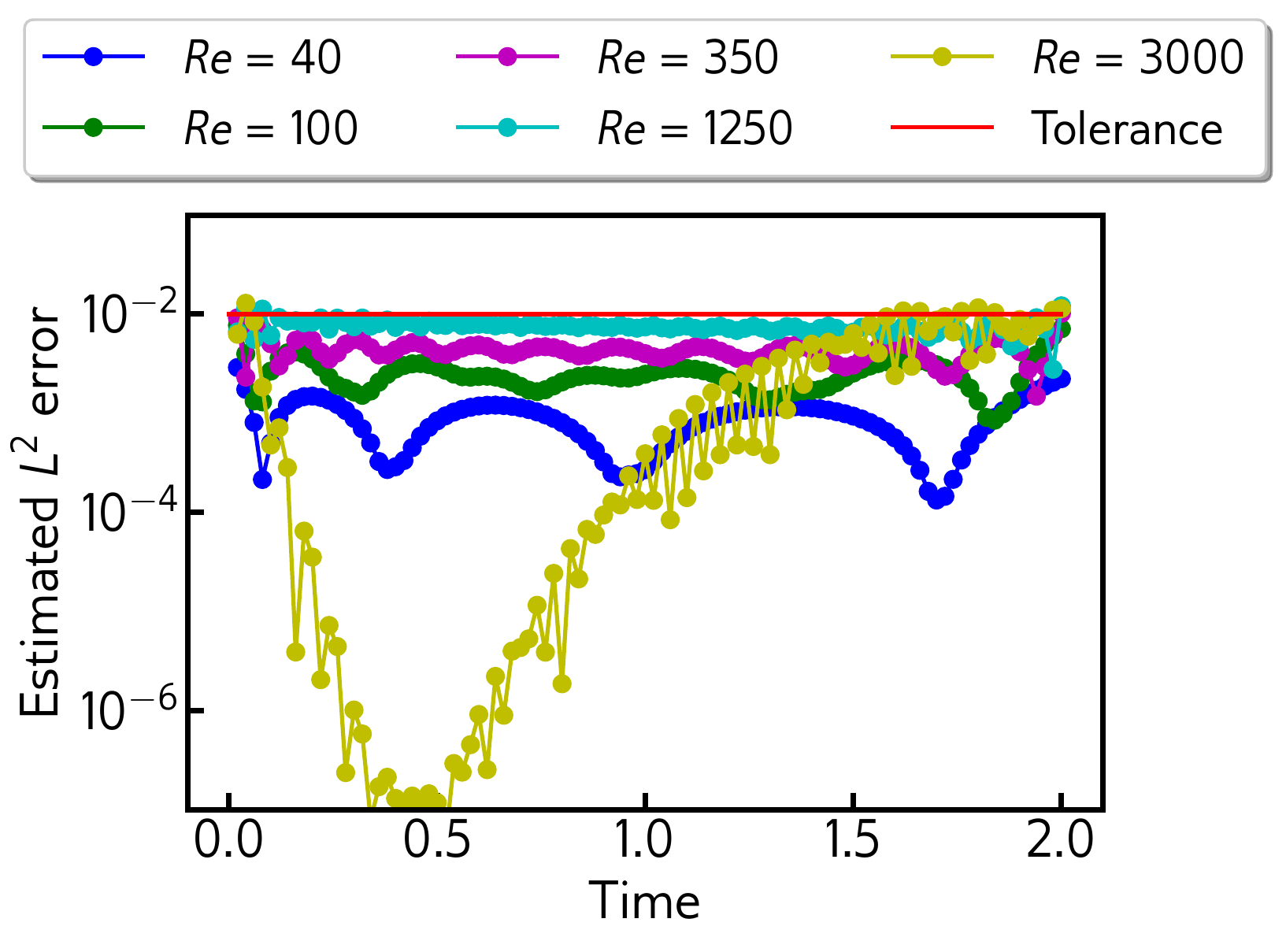}
		\caption{}
		\label{fig:burgers-error-plot-b}
	\end{subfigure}
	\caption{Plot (a)~shows the true error of the ActLearn-POD-KSNN solution for the Burgers' equation on a new time grid corresponding to several out-of-training Reynolds numbers. The tolerance used for termination of the active learning procedure is also shown for comparison. Plot (b)~shows the estimated error on the original time grid.}
	\label{fig:burgers-error-plot}
\end{figure}

During the active learning loop, we sample from the candidate set $P^*$ at each iteration. The initial state of $P^*$ for the reported results comprises $79$ candidate values -- upon exclusion of $21$ values present in $P$ (initially) from the total $100$ values. \Cref{fig:burgers-active-learn-a} shows the estimated error $\mathcal{E}^{(iter)}$ (see \Cref{alg:summary-offline}) varying with greedy iterations of the active learning process. The error decreases over iterations, and we stop the loop after a tolerance of $10^{-2}$ is met. So, $33$ new parameter points are selected in a greedy fashion and their corresponding estimated error values are reported in \Cref{fig:burgers-active-learn-a}.

The ultimate choice of viscosity values and their corresponding POD subspace dimensions are shown in \Cref{fig:burgers-active-learn-b}. The reported dimensions also account for the POD space adaptation (refinement) when required through the iterations, as discussed in \Cref{subsec:active-learn}. For low viscosities, the subspace dimension is comparatively higher. And new selections are mostly concentrated in regions where the nature of the POD subspace changes significantly. Among all the individual parametric POD subspaces, the lowest energy criterion $\hat \eta$ in \cref{eqn:new-energy-criterion} is $2.676 \times 10^{-11}$. This is used as a condition for deciding the POD subspace dimension for any newly queried parameter.

\begin{figure}[!t] 
	\centering
	\includegraphics[width=\textwidth, trim=0 0 0 0, clip]{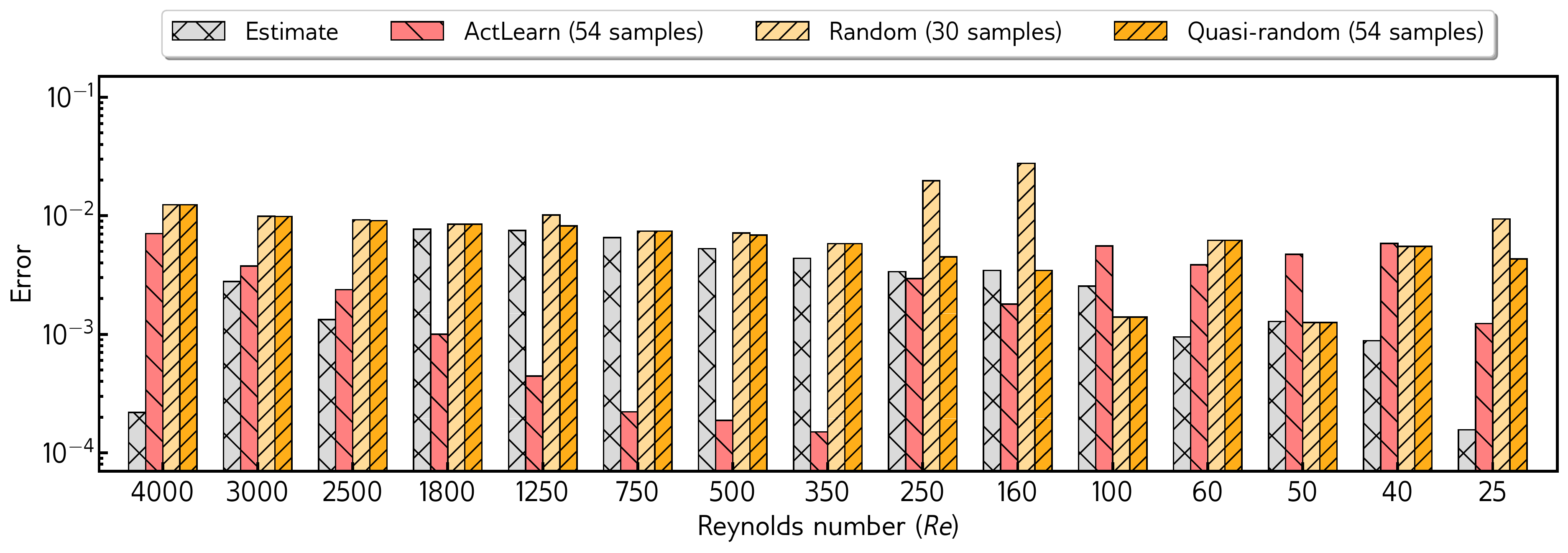}
	\caption{Burgers' equation:  Error comparison between the ActLearn-POD-KSNN solution and the POD-KSNN solutions upon a random (30 samples, seed 10) and quasi-random (54 samples, seed 10) selection of parametric training data. The values labeled 'ActLearn', 'Random', and 'Quasi-random' are the time-averaged (over the test time grid) relative $l^2$ errors in the spatial domain. The values labeled 'Estimate' are the time-average (over the training time grid) of the error estimate values given by \cref{eqn:error-est-time-each-entry}. All the reported Reynolds numbers are outside of the training set.}
	\label{fig:burgers-random-comparison}
\end{figure}

In~\Cref{fig:burgers-rom-sol-surf}, the solution to the ActLearn-POD-KSNN surrogate model is compared with the true solution over the entire space-time domain. This allows us to see the evolution of the solutions in time. The Reynolds numbers are taken outside the training set -- $\{40, 100, 350, 1250, 3000\}$. The solution is computed on a new test time grid with $99$ instances starting from $0.01$ with a step size of $0.02$. We can see that the ActLearn-POD-KSNN solution agrees well with the ground truth. To further visualize and compare the solution with the truth, solutions are plotted in \Cref{fig:burgers-rom-sol} for three representative time instances from the start, middle, and end of the time domain. The error contours in the third column of \Cref{fig:burgers-rom-sol-surf} show the point-wise difference in the space-time domain between the surrogate solution and the ground truth. The solution error stays reasonable for the entire range of $Re$ values. 

\begin{figure}[!t] 
	\centering
	\begin{subfigure}[b]{\textwidth}
		\centering
		\includegraphics[width=\textwidth, trim=0 0 0 0, clip]{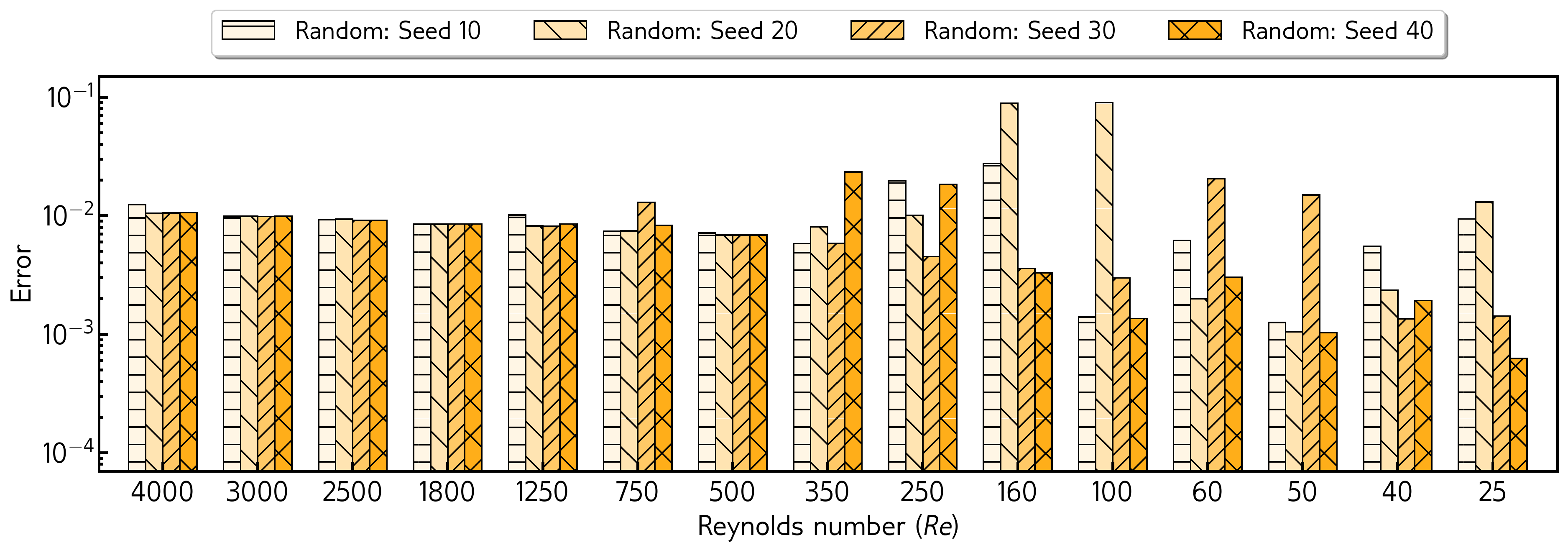}
		\caption{Error in the solution upon random selection of parametric training data.}
		\label{fig:burgers-random-seed-comparison-a}
	\end{subfigure}
	\begin{subfigure}[b]{\textwidth}
		\centering
		\includegraphics[width=\textwidth, trim=0 0 0 -20, clip]{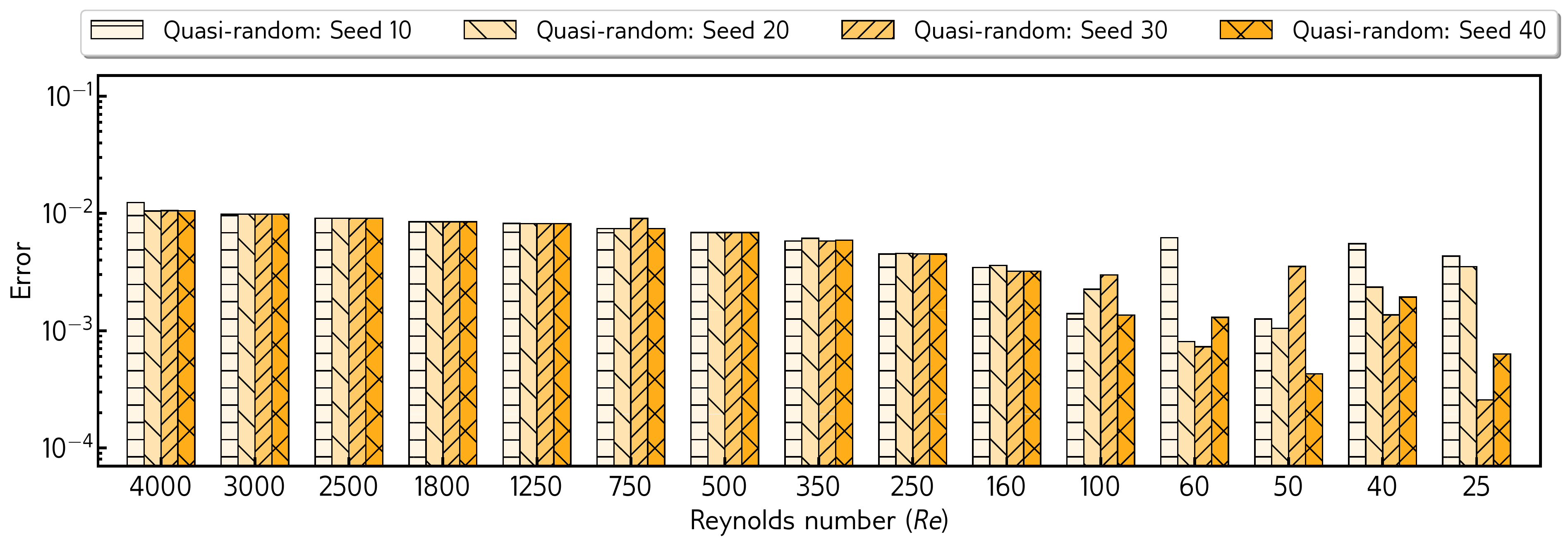}
		\caption{Error in the solution upon quasi-random selection of parametric training data.}
		\label{fig:burgers-random-seed-comparison-b}
	\end{subfigure}
	\caption{Burgers' equation:  Error comparison between the POD-KSNN solutions upon a random (30 samples) and quasi-random (54 samples) selection of parametric training data with four different starting seeds: $\{10,20,30,40\}$. The error values are the time-averaged (over the test time grid) relative $l^2$ errors in the spatial domain. All the reported Reynolds numbers are outside of the training set.}
	\label{fig:burgers-random-seed-comparison}
\end{figure}

To obtain an estimate of the relative solution error in the surrogate's approximation at a newly queried parameter $\mu^*$, we train the KSNNs using the relative error data given by \cref{eqn:rel-error} for all $\mu_i \in P$. Doing this, we obtain an error estimate $\boldsymbol{\tilde{\varepsilon}}(Re)$ for all the time instances, which is given by $\boldsymbol{\tilde{\varepsilon}}(\boldsymbol{\mu}^*)$ in \cref{eqn:error-est-time}. Such an estimation of the error is shown in \Cref{fig:burgers-error-plot-b} for the training (original) time grid corresponding to several out-of-training Reynolds number. Whereas, \Cref{fig:burgers-error-plot-a} shows the true relative error values of the ActLearn-POD-KSNN solution. Here, the time instances are different from the training time grid. For the most part, the true relative errors are bounded by the tolerance criterion $10^{-2}$ which is used for the active learning loop.

In~\Cref{fig:burgers-random-comparison}, we provide a comparative study between ActLearn-POD-KSNN solution error and POD-KSNN solution errors that are obtained by randomly picking $30$ and $54$ parameter samples for preparing the training snapshots of the surrogate. We label the surrogate solution error obtained by training with $54$ random samples as quasi-random because this choice is actually informed by the total number of parameter samples ($21 + 33 = 54$) upon termination of the active learning loop. The random selection from a pool of $100$ values of $Re$ (which are the same as described before during the preparation of sets $P$ and $P^*$) is done with a fixed representative seed value of $10$ using the NumPy library written for Python programming language. The surrogate solutions $(\mathbf{u}_s)_k(Re) := \mathbf{u}_s(t_k,Re)$ are first computed over the previously described test time grid ($t_k$ with $k=1,\ldots, 99$, $t_1=0.01$, a uniform step size of 0.02, and the total discrete time instances are $\tilde{N}_t = 99$) after which the relative $l^2$ error values in space are obtained,
\begin{equation}
	\varepsilon_k(Re) := \frac{\|\mathbf{u}_k (Re) - (\mathbf{u}_s)_k(Re)\|_2}{\|\mathbf{u}_k (Re)\|_2}
\end{equation}
We then take the average of these $l^2$ errors over all the discrete time instances, i.e., Error $:= (1/\tilde{N}_t) \sum_k \varepsilon_k (Re)$, and report the final result for active, random, and quasi-random sampling in~\Cref{fig:burgers-random-comparison}. In a similar fashion, we report the time-average of the error estimate values $\tilde \varepsilon_j (Re)$ defined in \cref{eqn:error-est-time-each-entry} on the training time grid ($t_j$ with $j=1,\ldots, 100$, $t_1=0.0$, a uniform step size of 0.02, and the total discrete time instances are $N_t = 100$), i.e., Estimate $:=(1/N_t) \sum_j \tilde \varepsilon_j (Re)$. We observe that the error in the ActLearn-POD-KSNN solution is bounded by the tolerance of $10^{-2}$ specified for termination of the active learning procedure.

In~\Cref{fig:burgers-random-seed-comparison}, we compare the POD-KSNN solution error for scenarios when $30$ and $54$ parameters are randomly picked with four different starting seed values: $\{10,20,30,40\}$. Similar to \Cref{fig:burgers-random-comparison}, the reported error values are the time-average (over the test time grid) of the relative $l^2$ error in space. \Cref{fig:burgers-random-seed-comparison-a} shows that with $30$ random samples the error goes up to $10^{-1}$, and there is also significant difference among the solution error values corresponding to certain testing Reynolds numbers for different starting seeds. We observe from \Cref{fig:burgers-random-seed-comparison-b} that the error values reduce by adding more samples, but there is still a noticeable variation between the values for different starting seeds. This situation is addressed by the active learning framework. Moreover, the procedure provides an idea about the number of parameter samples required for preparing the snapshot training data such that the constructed surrogate model approximates the solution (at any new parameter and time instance) up to some predefined accuracy level, as specified by the tolerance value.


\subsection{Shallow Water Equations}%
\label{subsec:swe}

The free-surface flows in water bodies like channels or rivers can be modeled using the shallow water equations \cite{Cha04}. They are obtained from the incompressible Navier-Stokes equations under the condition that the fluid flow's vertical extent is significantly smaller than its horizontal extent. The conservation of mass and momentum takes the following form:
\begin{align}
	\partial_t h + \partial_x (h u) &= 0, \\
	\partial_t(h u) + \partial_x \Big( h u^2 + \frac{1}{2} g h^2 \Big) &= - \frac{\nu}{\lambda} u,
\end{align}
where $h(t,x)$ is the depth of the channel, $u(t,x)$ is the depth-averaged velocity along the length of the channel, $\nu$ is the dynamic viscosity, $\lambda$ is the mean-free path, $g$ is the gravitational acceleration. This form of the shallow water equations is also known as the Saint-Venant system, which naturally allows us to capture a constant vertical velocity profile in a channel. To carry out the experiments, the initial condition of the height is taken as a smooth bump described by the following nonlinear function:
\begin{equation}
	h(0,x) = 1 + \exp(3 \cos(\pi (x + 0.5)) - 4).
\end{equation}
The initial velocity is taken to be constant along $x$,
\begin{equation}
	u(0,x) = 0.25.
\end{equation}

\begin{figure}[!b] %
	\centering
	\begin{subfigure}[b]{0.45\textwidth}
		\centering
		\includegraphics[width=.8\textwidth]{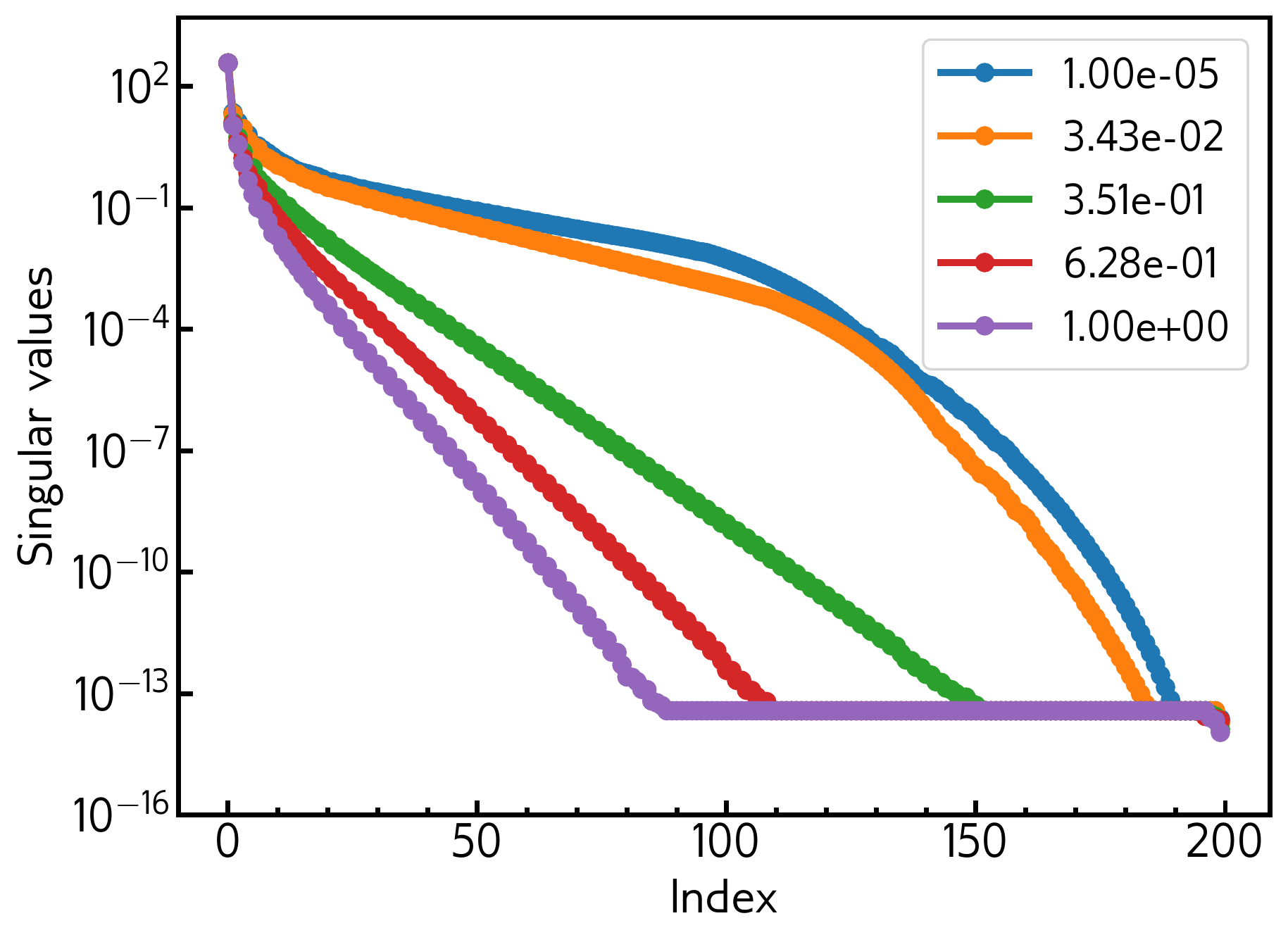}
		\caption{Singular value decay for the fluid height.}
		\label{fig:swe-sing-val-a}
	\end{subfigure}
	\hspace{2em}
	\begin{subfigure}[b]{0.45\textwidth}
		\centering
		\includegraphics[width=.8\textwidth]{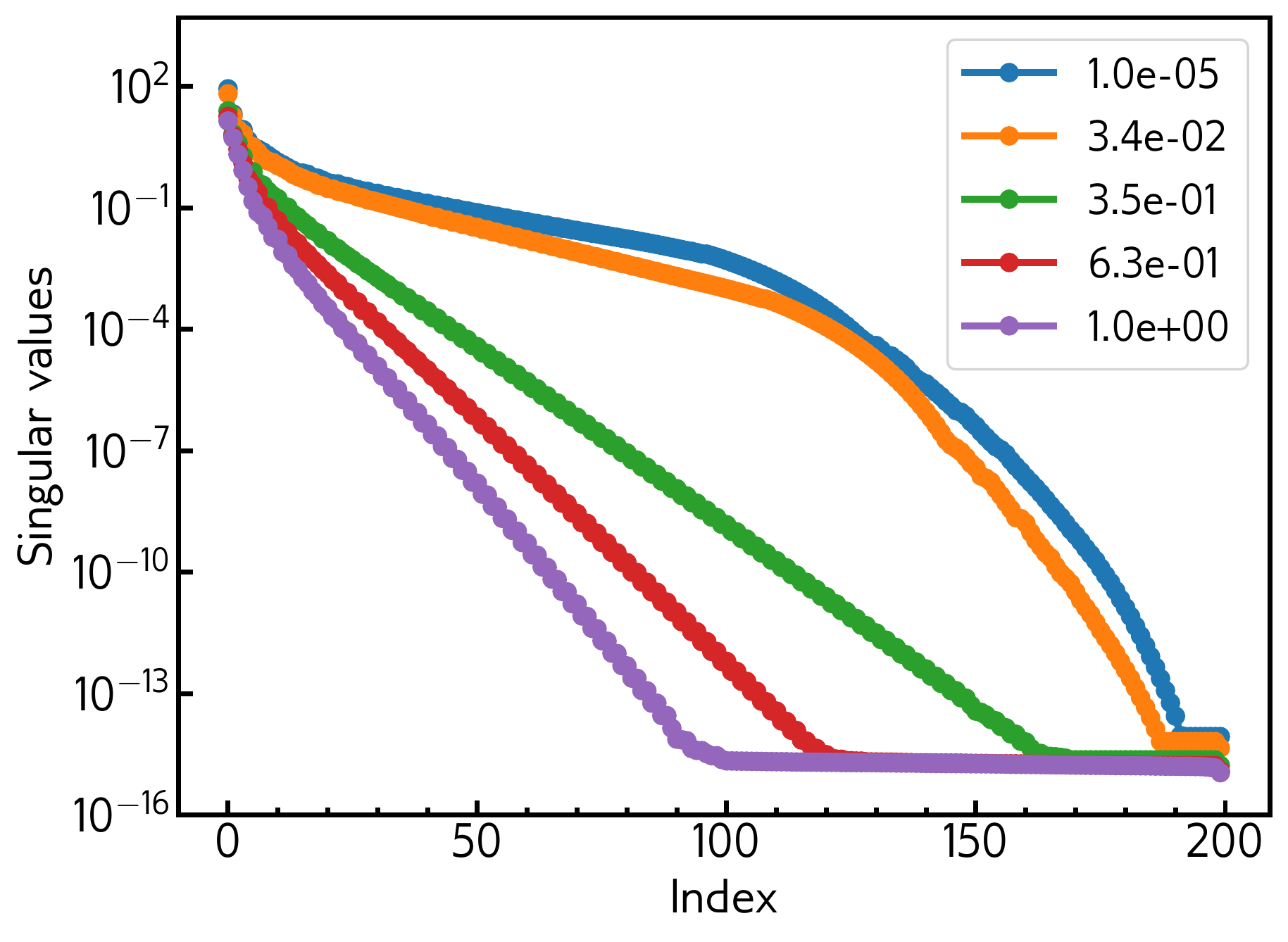}
		\caption{Singular value decay for the fluid velocity.}
		\label{fig:swe-sing-val-b}
	\end{subfigure}
	\caption{Shallow water equations: The decay in singular values for various viscosity samples $\nu$.}
	\label{fig:swe-sing-val}
\end{figure}

\begin{figure}[!t] %
	\centering
	\begin{subfigure}[b]{0.355\textwidth}
		\centering
		\includegraphics[width=1.1\textwidth, trim=0 0 0 0, clip]{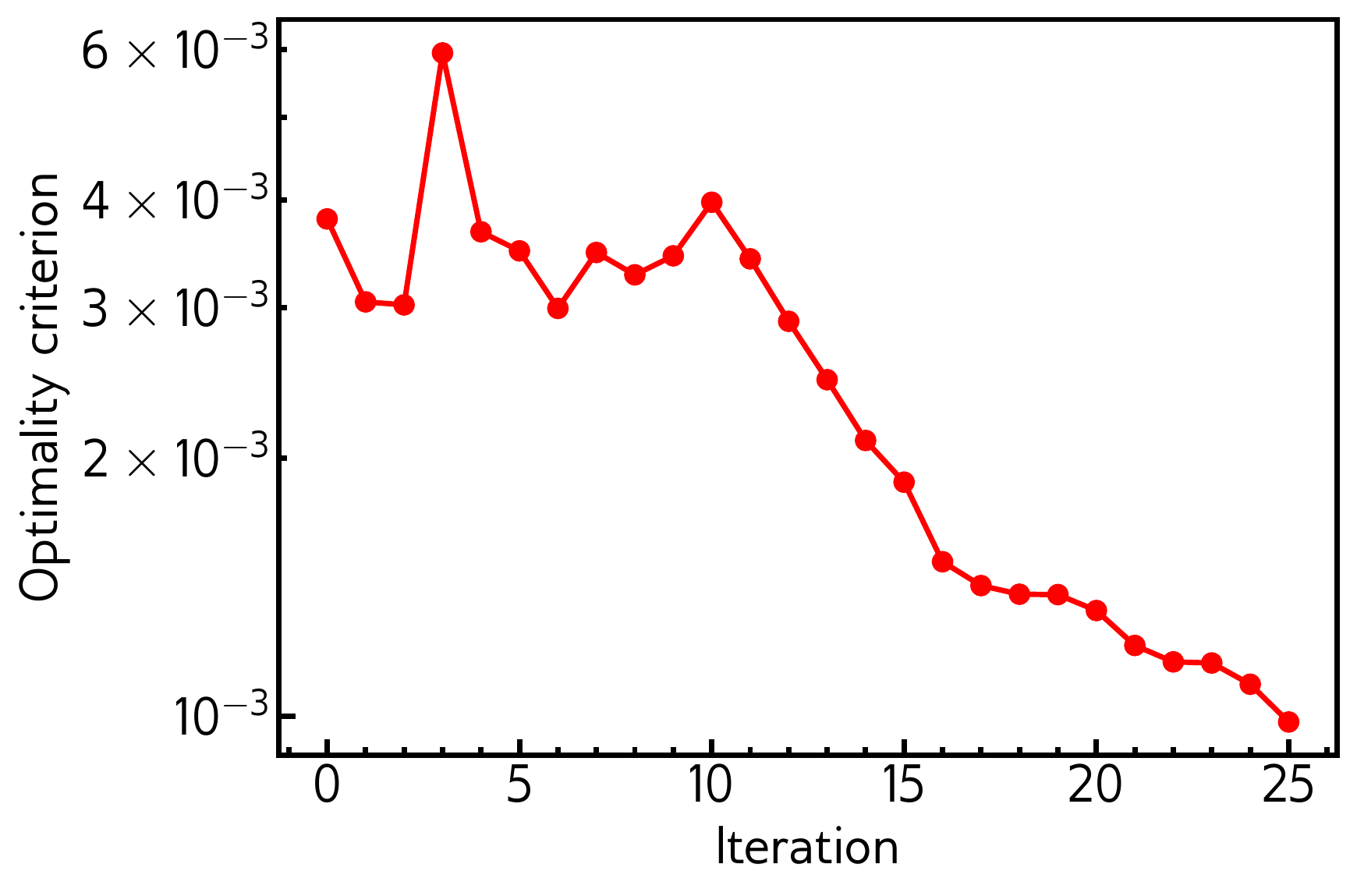}
		\caption{}
		\label{fig:swe-active-learn-a}
	\end{subfigure}
	\hspace{2em}
	\begin{subfigure}[b]{0.35\textwidth}
		\centering
		\includegraphics[width=1\textwidth, trim=0 0 0 0, clip]{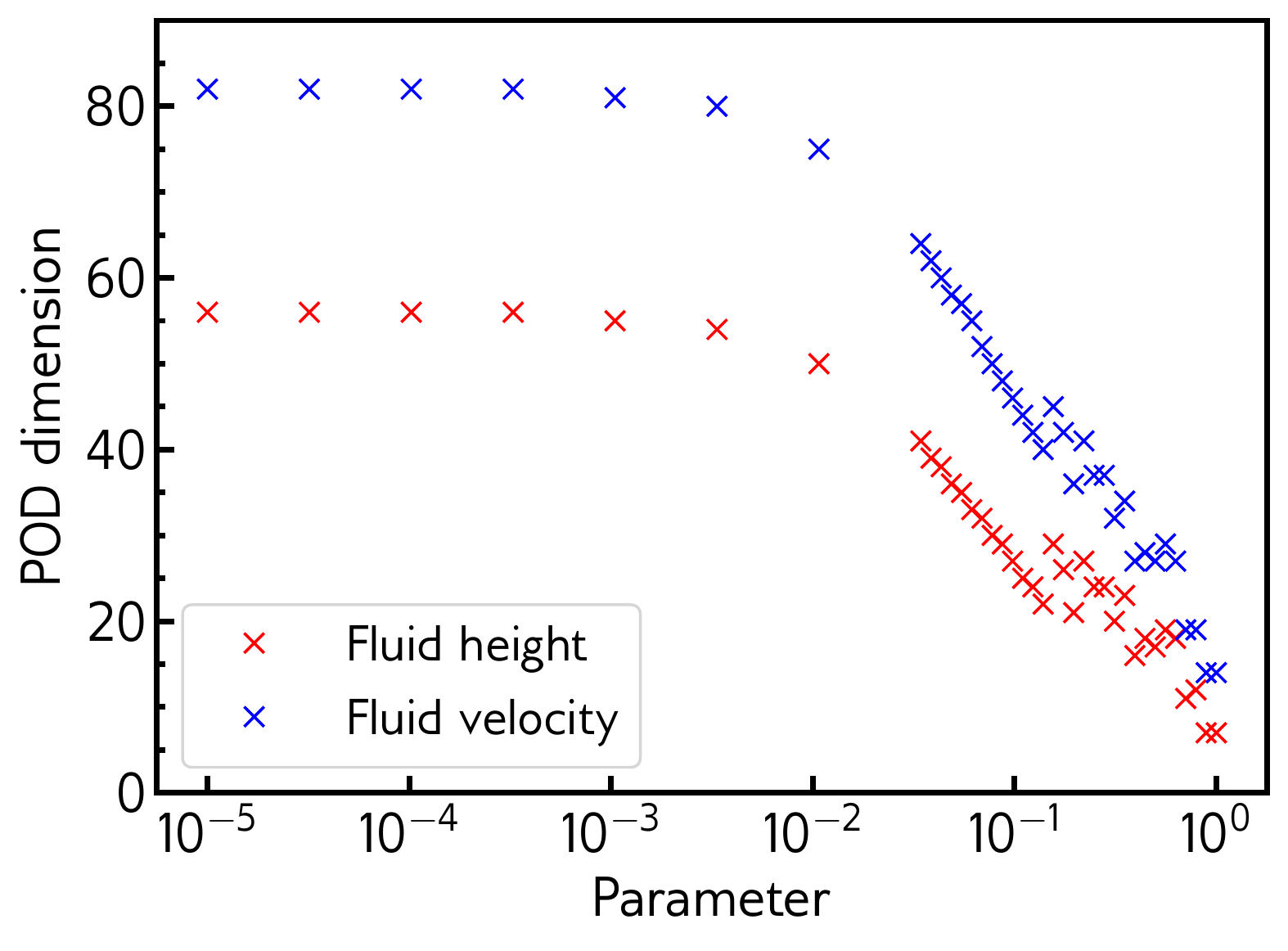}
		\caption{}
		\label{fig:swe-active-learn-b}
	\end{subfigure}
	\caption{Active learning for shallow water equations: (a)~shows the optimality criterion $\mathcal{E}^{(iter)}$ (refer \Cref{alg:summary-offline}) varying with greedy iterations of the active learning process; (b)~shows the final POD subspace dimension for each of the chosen parameter samples.}
	\label{fig:swe-active-learn}
\end{figure}

The periodic spatial domain is $\Omega \in [-1,1]$ with $601$ grid nodes, and the time-domain is $t \in [0,2]$ in which the solution is stored at 200 uniform time steps. The high-fidelity solution of the system of equations is computed using a discontinuous Galerkin solver with local polynomial reconstruction of degree~$1$~\cite{kapadia2019}. The Riemann solver utilized to compute the numerical flux is local Lax-Friedrichs. For integration in time, we use a second-order strong stability preserving Runge-Kutta scheme. The equations are solved in non-dimensional form, and all the reported parameter values are non-dimensional. For details about the conversion to dimensionless form, we suggest the reader to refer~\cite{KowT18}. We perform a singular value decomposition of the parameter-specific snapshot matrices for the fluid height and velocity and plot their respective singular value decays in \Cref{fig:swe-sing-val}. The decay is already not very fast for higher viscosity values, which gets exacerbated further as the viscosity values are decreased. The hyperbolic nature of these equations renders an even greater challenge for constructing an efficient reduced-order surrogate.

For our experiments, we fix $\lambda = 0.1$, and vary the viscosity $\nu$ from $1$ to $10^{-5}$. The total number of discrete parameters we consider when accounting for both the candidate set $P^*$ and the parameter set $P$ are $100$. These $100$ samples of $\nu$ are picked by uniformly dividing the logarithmic $\nu$ values ($log_{10}(10^{-5})$ and $log_{10}(1)$) into $99$ intervals, ensuring a decent pool of parameters.

To begin the active learning procedure, the parameter set $P$ is initiated by $11$ viscosity values corresponding to the following indices,
\begin{equation*}
	\{0,99,10,20,30,40,50,60,70,80,90\}.
\end{equation*}
Here, the $\nu$ values are ordered from lowest to highest and the index starts from $0$ when counting the $100$ values. Like done for Burgers' equation, we report the indices instead of exact values for ease of readability. High-fidelity snapshots are generated for all the viscosities in $P$ for $200$ time instances. Using these snapshots, POD-approximate solutions are computed such that the POD subspace retains $99.99997 \%$ of the energy.

During the active learning loop, we sample from the candidate set $P^*$ at each iteration. The initial state of $P^*$ for the reported results comprises $89$ candidate values---upon exclusion of $11$ values present in $P$ (initially) from the total $100$ values. \Cref{fig:swe-active-learn-a} shows variation of the estimated error $\mathcal{E}^{(iter)}$ (see \Cref{alg:summary-offline}) with greedy iterations of the active learning process. The error decreases over iterations, and we stop the loop after a tolerance of $10^{-3}$ is met. So, $26$ new viscosity samples are picked in a greedy fashion for which the estimated error values are reported in \Cref{fig:swe-active-learn-a}.

The ultimate choice of viscosity values and their corresponding POD subspace dimensions are shown in \Cref{fig:swe-active-learn-b}. Like for the Burgers' equation, here as well the reported dimensions account for the POD subspace refinement through the iterations. Also like Burgers' equation, for low values of viscosities, the subspace dimension is comparatively higher. New selections are mostly concentrated in regions where the nature of the POD subspace changes significantly. Here, this is towards the moderate to high viscosity regions. Among all the individual parametric POD subspaces, the lowest energy criterion $\hat \eta$ in \cref{eqn:new-energy-criterion} for the fluid height is $2.989 \times 10^{-10}$, and for the fluid velocity is $6.362 \times 10^{-10}$. These are used as conditions for deciding the POD subspace dimension for the fluid height and velocity at any newly queried value of $\nu$ in the online phase (refer \Cref{alg:summary-online}).

\begin{figure}[H]
	\centering
	\begin{subfigure}[b]{0.32\textwidth}
		\centering
		\includegraphics[width=0.9\textwidth, trim=0 0 0 200, clip]{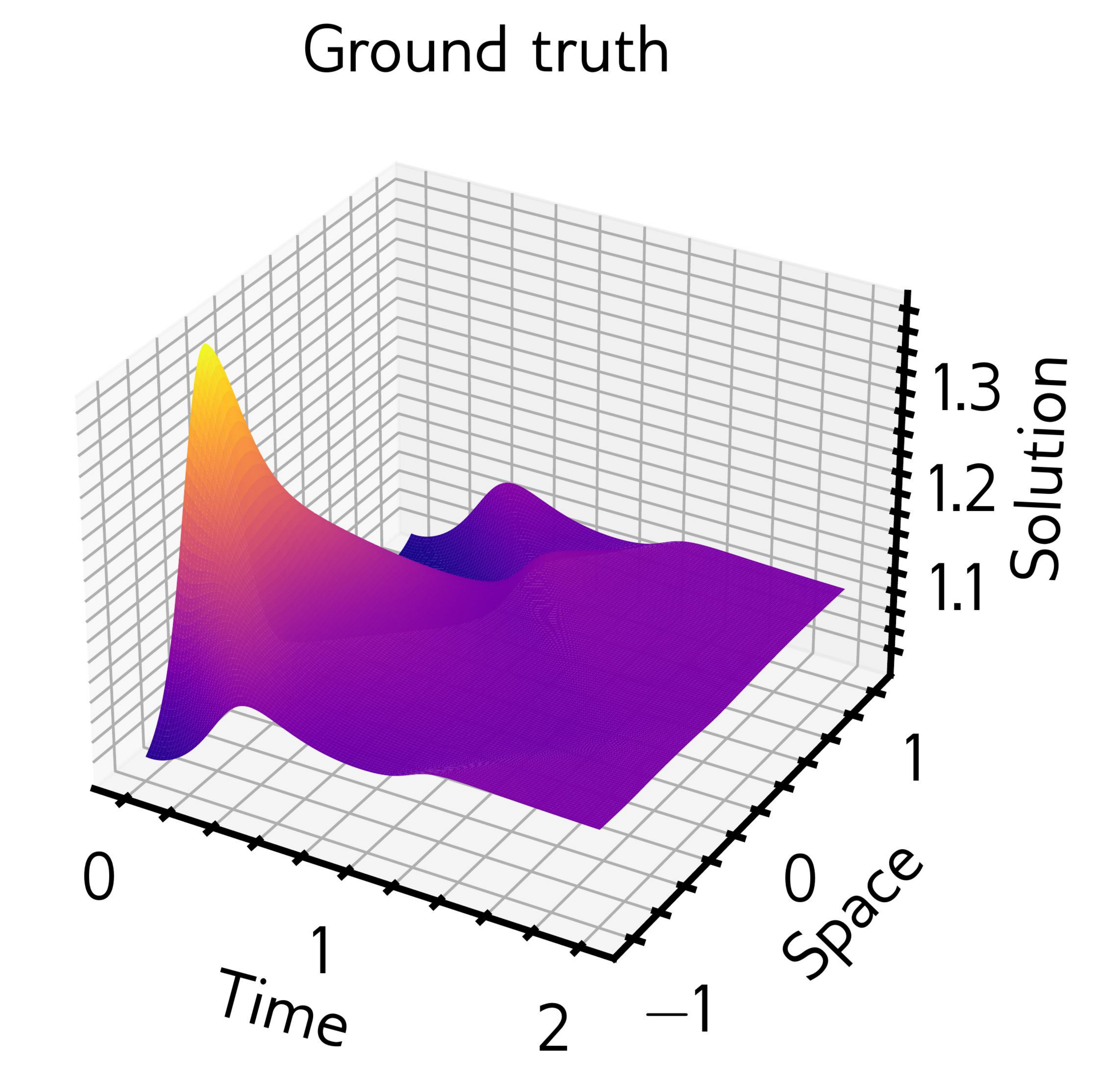}
	\end{subfigure}
	\begin{subfigure}[b]{0.32\textwidth}
		\centering
		\includegraphics[width=0.9\textwidth, trim=0 0 0 200, clip]{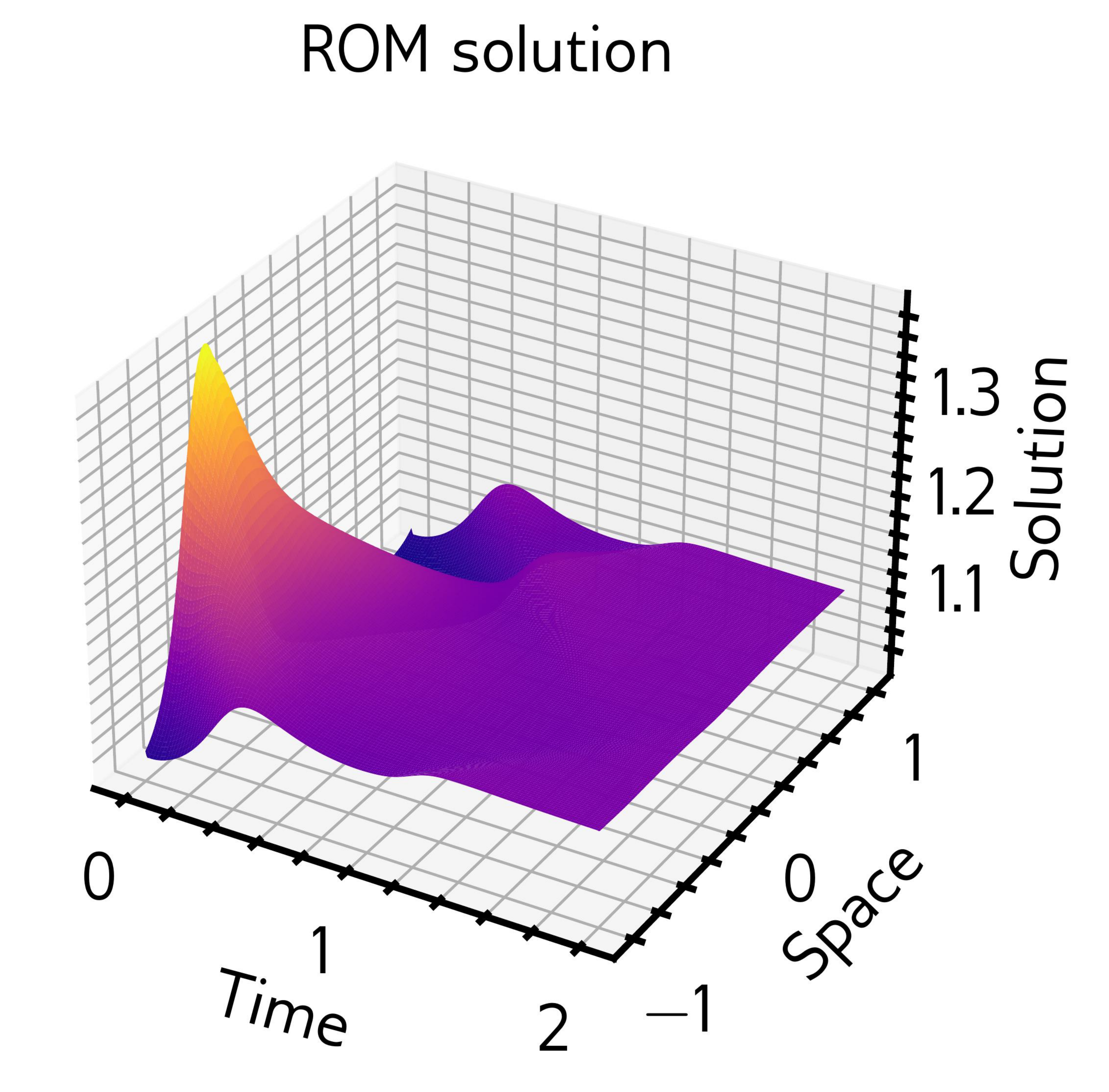}
	\end{subfigure}
	\begin{subfigure}[b]{0.32\textwidth}
		\centering
		\includegraphics[width=0.85\textwidth, trim=0 0 0 200, clip]{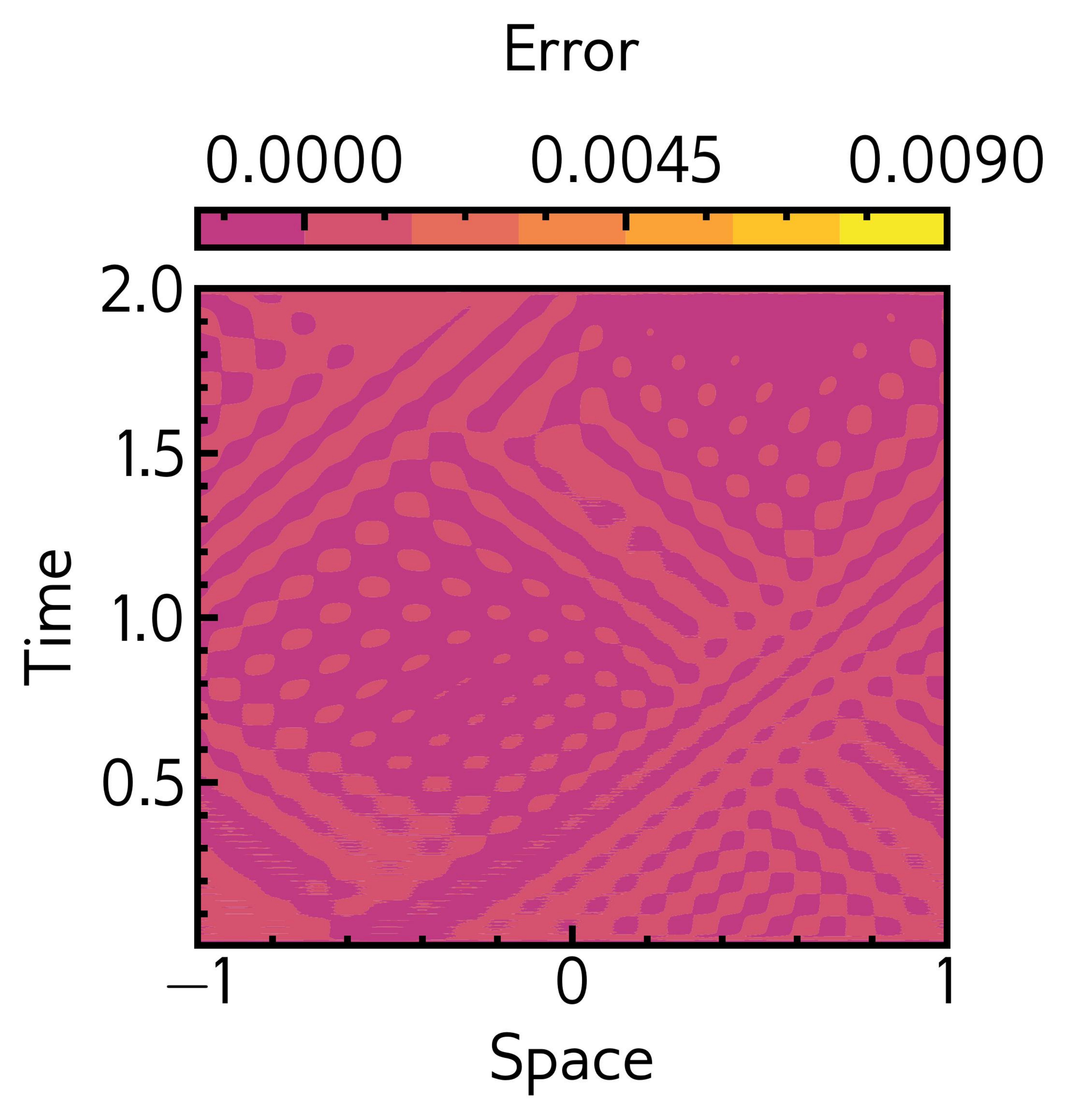}
	\end{subfigure}
	\begin{subfigure}[b]{0.32\textwidth}
		\centering
		\includegraphics[width=0.9\textwidth, trim=0 0 0 200, clip]{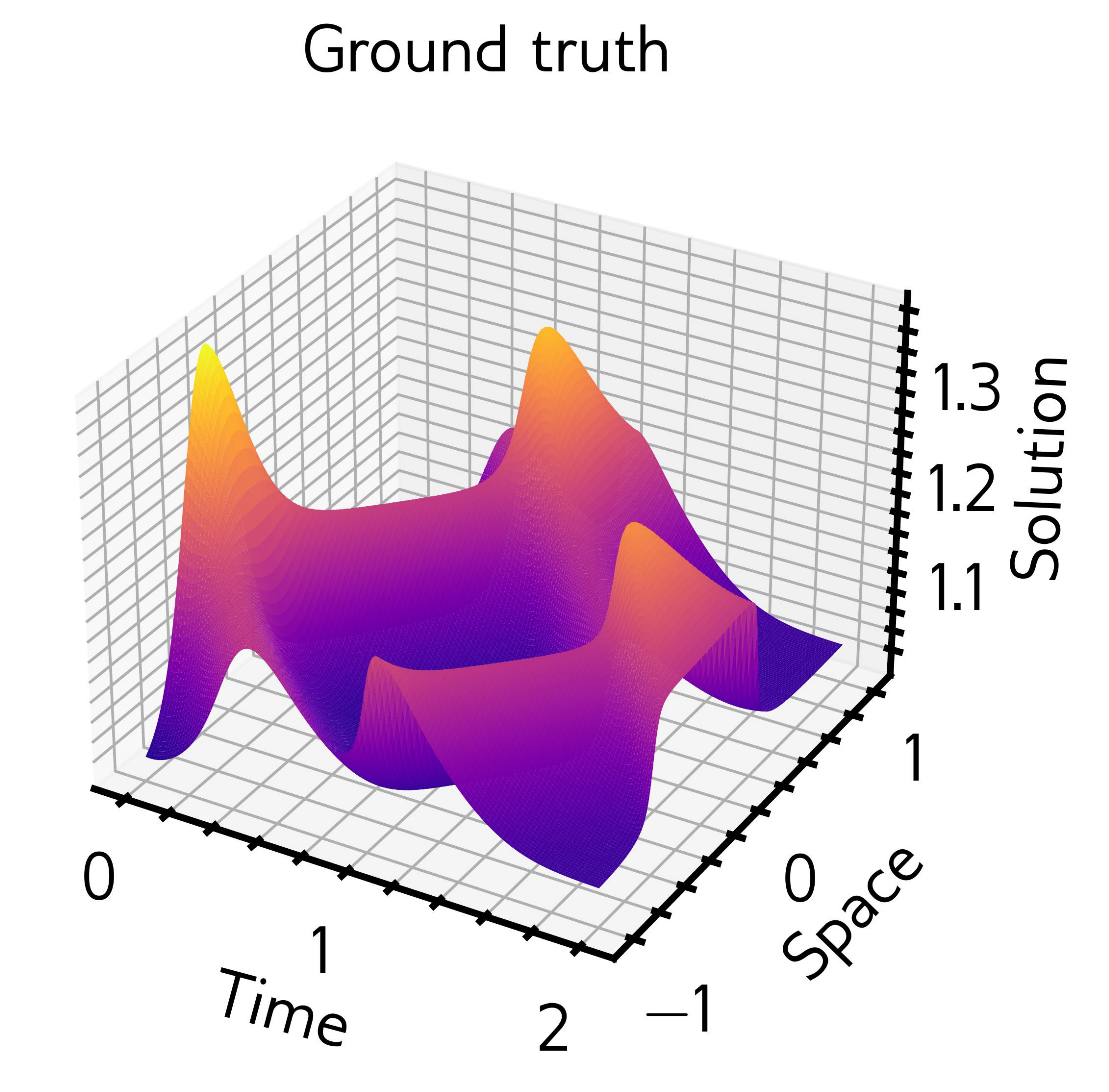}
	\end{subfigure}
	\begin{subfigure}[b]{0.32\textwidth}
		\centering
		\includegraphics[width=0.9\textwidth, trim=0 0 0 200, clip]{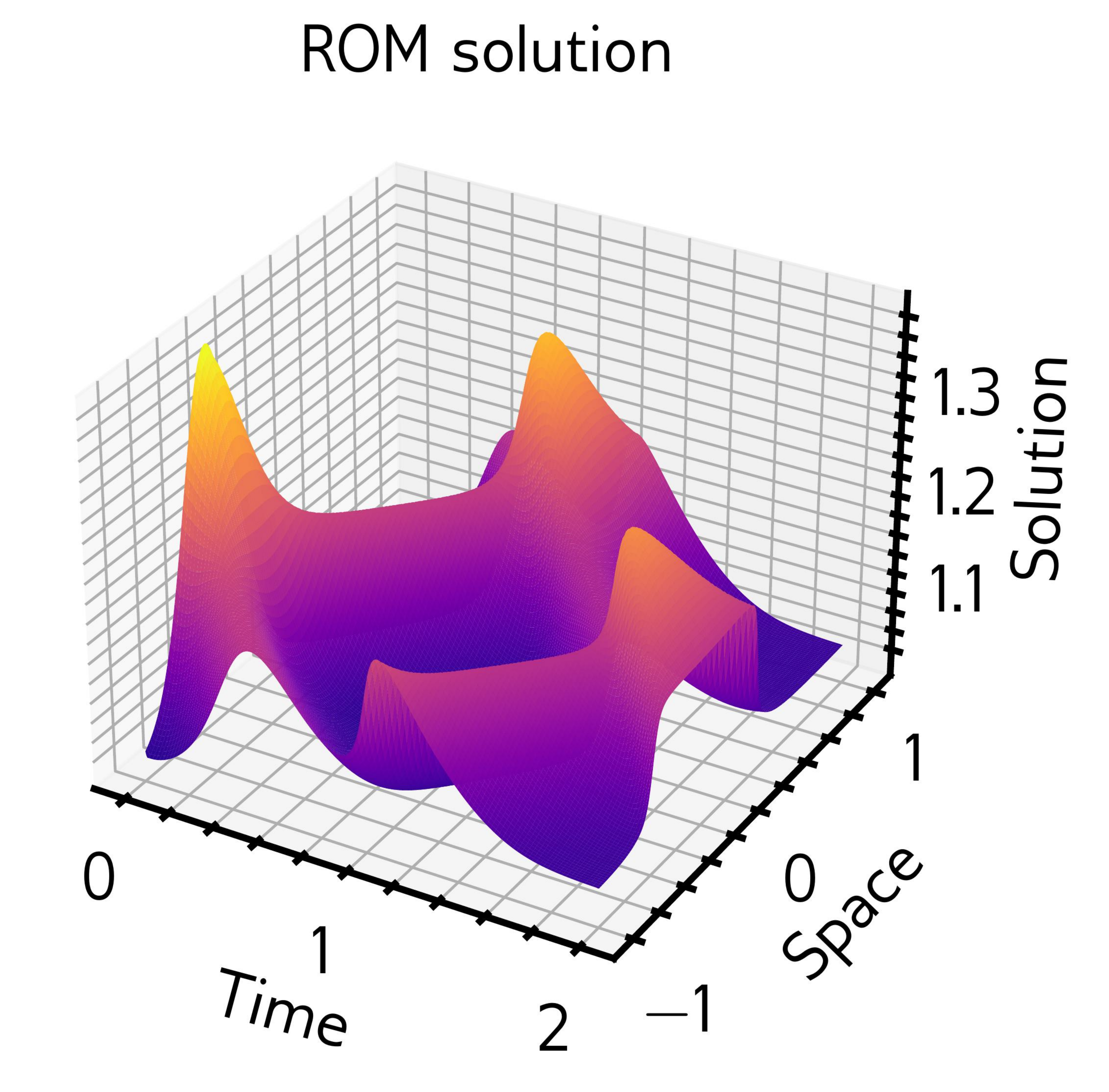}
	\end{subfigure}
	\begin{subfigure}[b]{0.32\textwidth}
		\centering
		\includegraphics[width=0.85\textwidth, trim=0 0 0 200, clip]{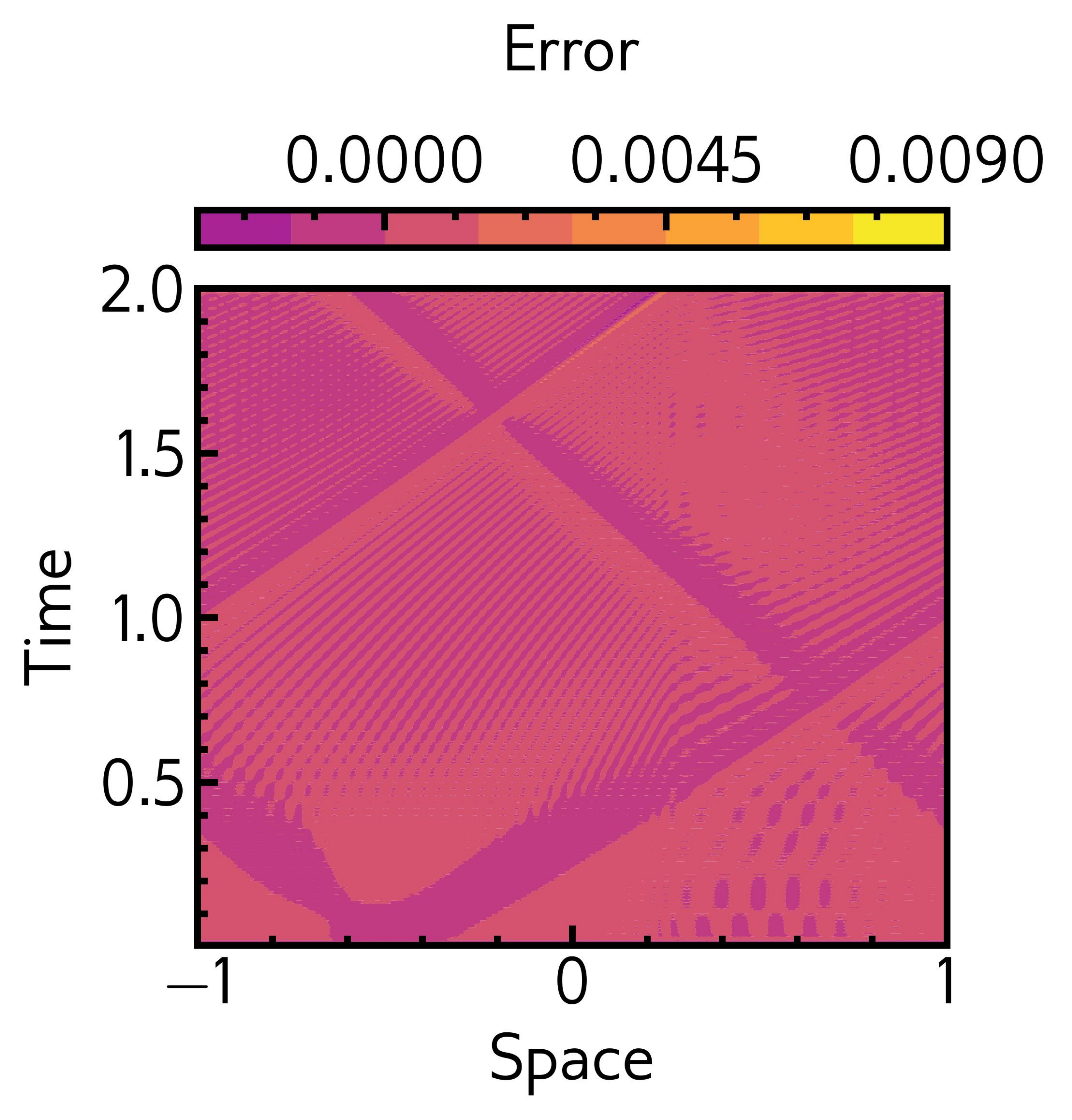}
	\end{subfigure}
	\begin{subfigure}[b]{0.32\textwidth}
		\centering
		\includegraphics[width=0.9\textwidth, trim=0 0 0 200, clip]{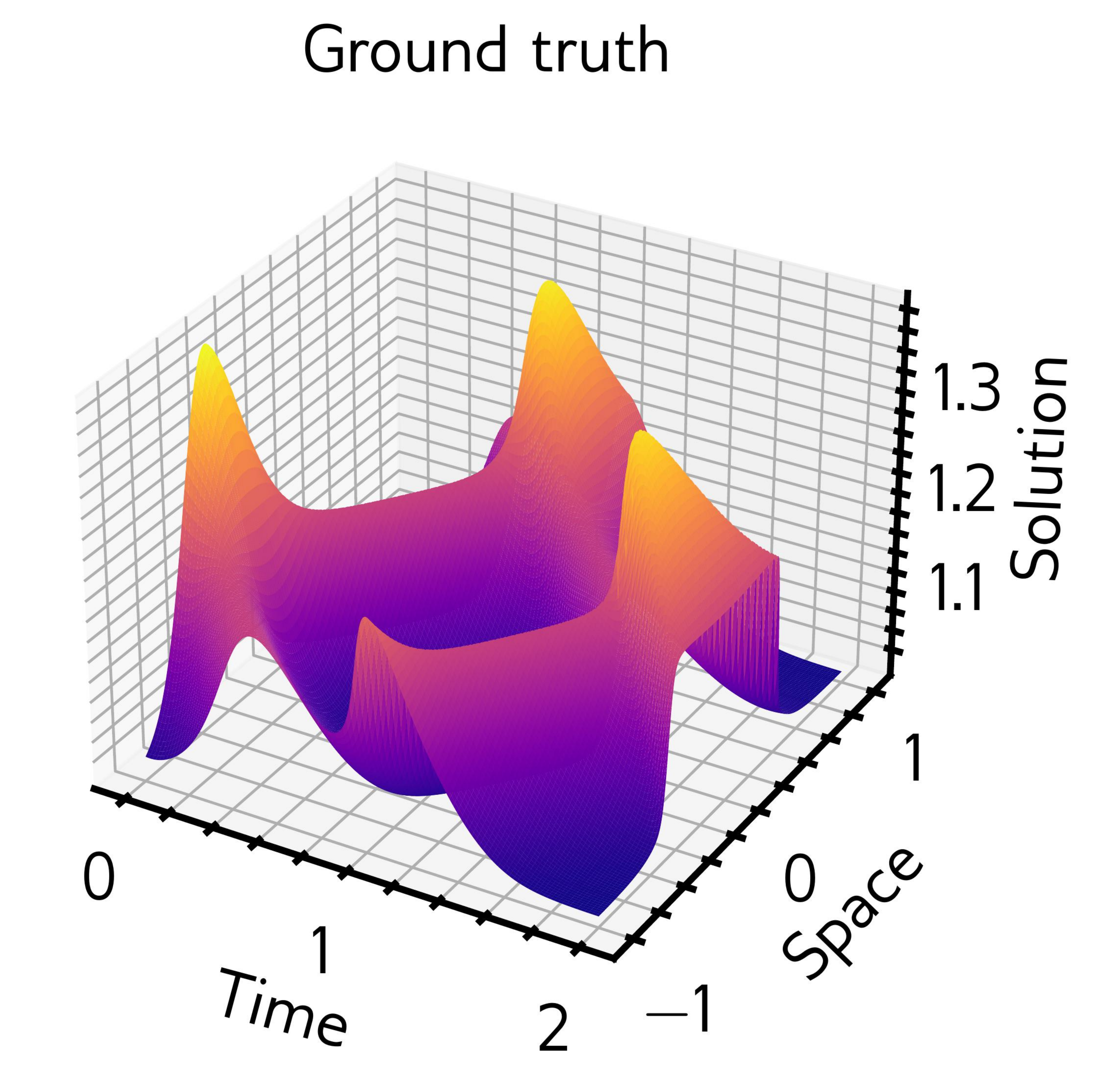}
	\end{subfigure}
	\begin{subfigure}[b]{0.32\textwidth}
		\centering
		\includegraphics[width=0.9\textwidth, trim=0 0 0 200, clip]{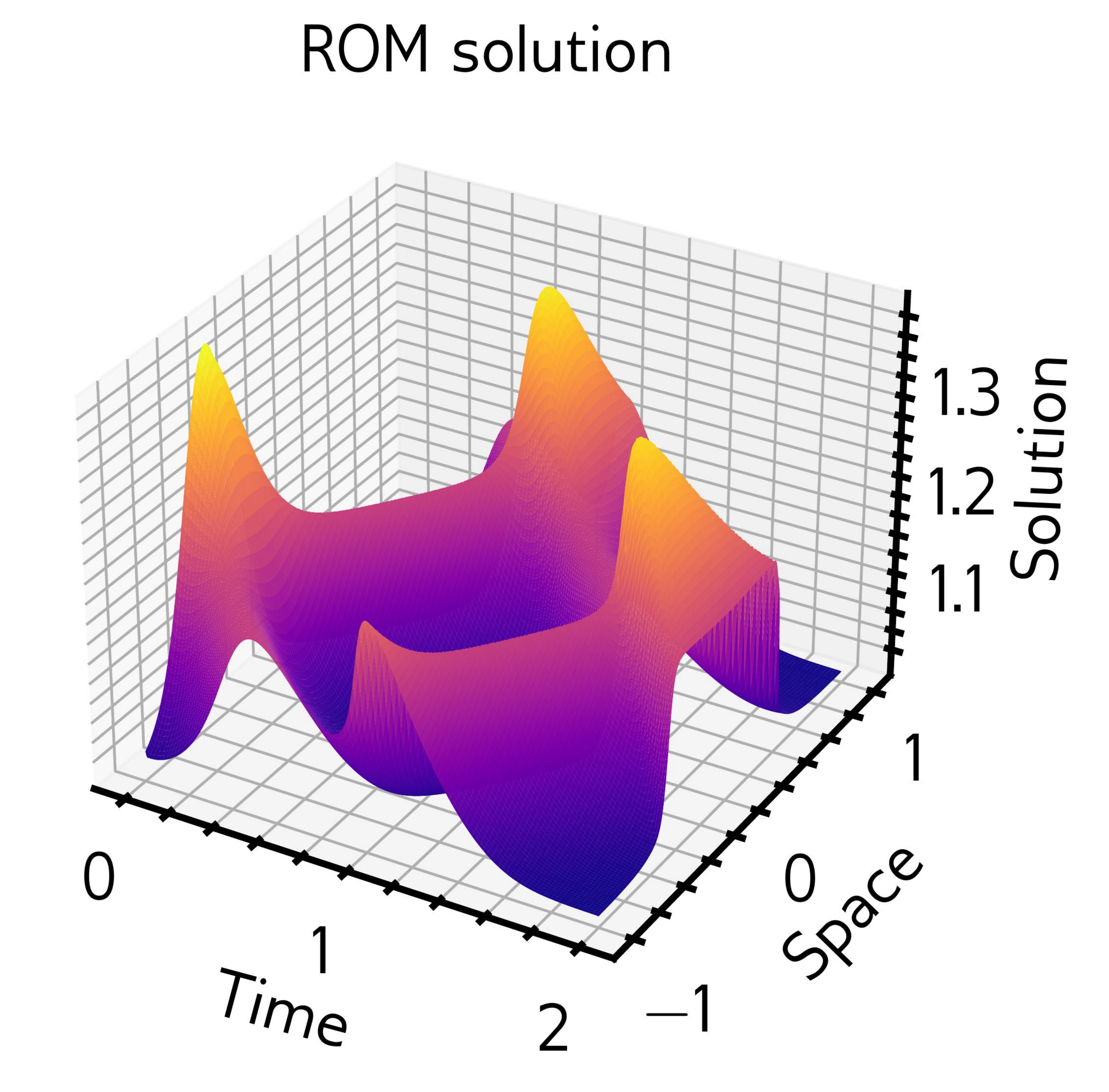}
	\end{subfigure}
	\begin{subfigure}[b]{0.32\textwidth}
		\centering
		\includegraphics[width=0.85\textwidth, trim=0 0 0 200, clip]{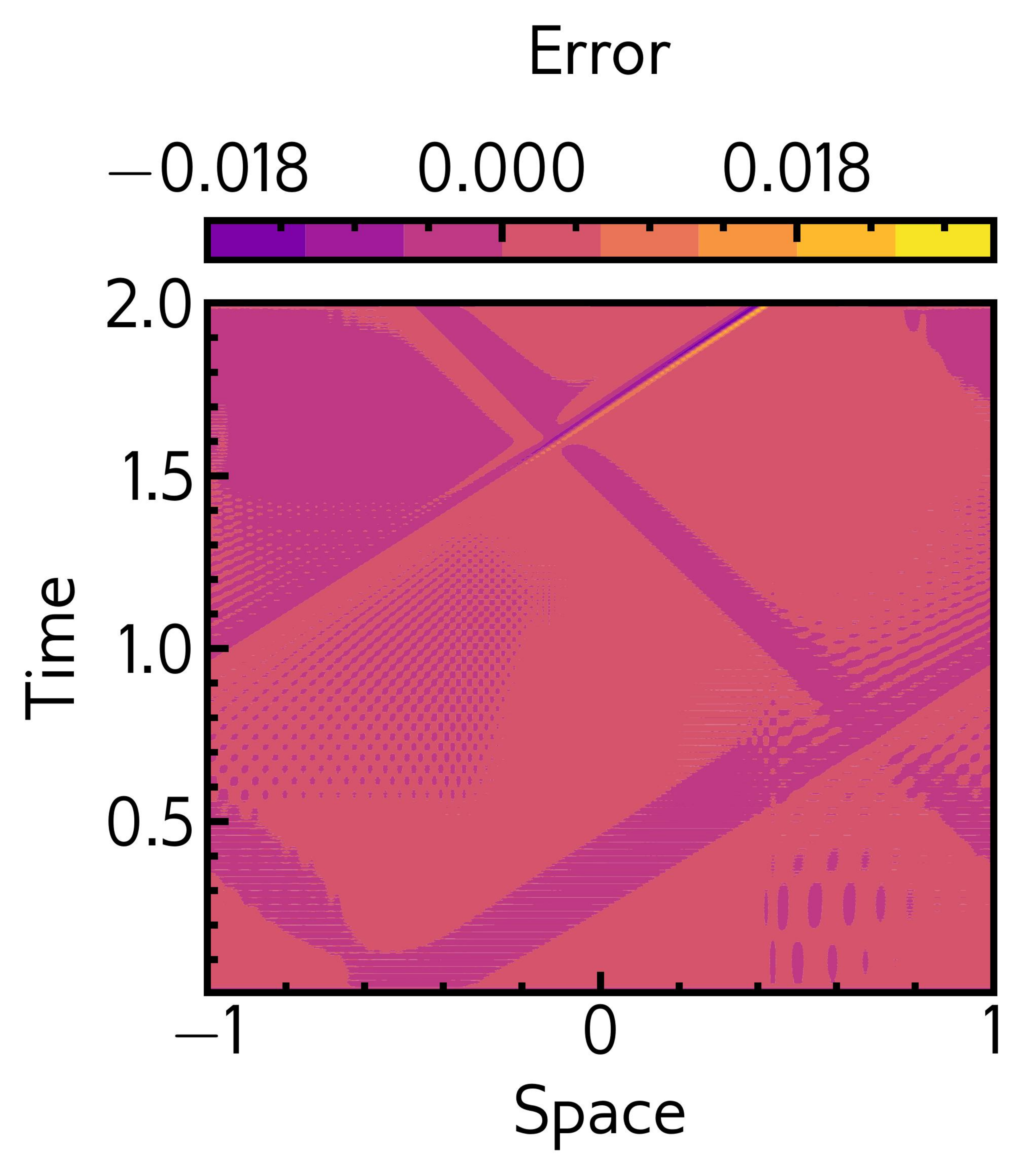}
	\end{subfigure}
	\begin{subfigure}[b]{0.32\textwidth}
		\centering
		\includegraphics[width=0.9\textwidth, trim=0 0 0 200, clip]{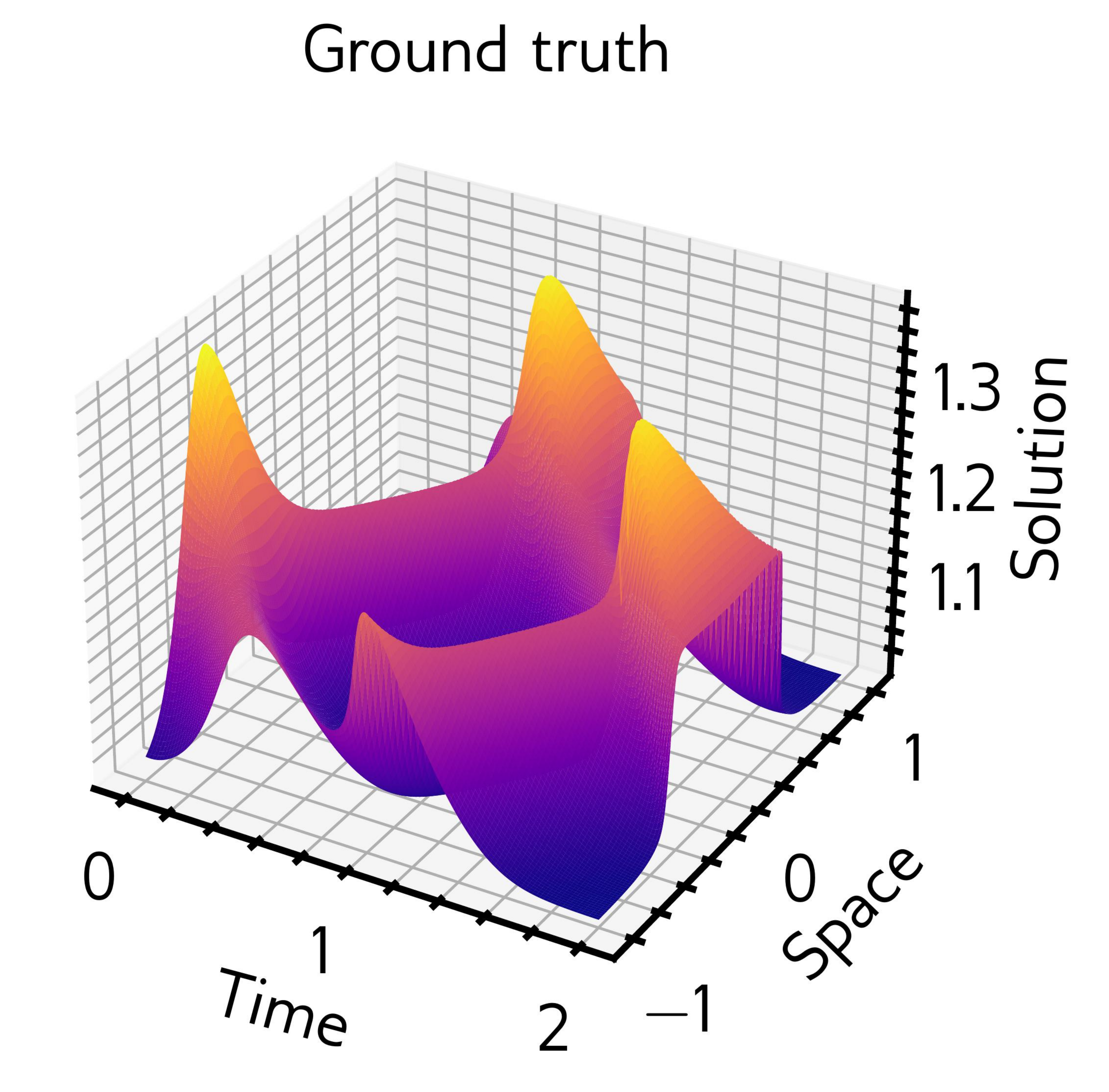}
	\end{subfigure}
	\begin{subfigure}[b]{0.32\textwidth}
		\centering
		\includegraphics[width=0.9\textwidth, trim=0 0 0 200, clip]{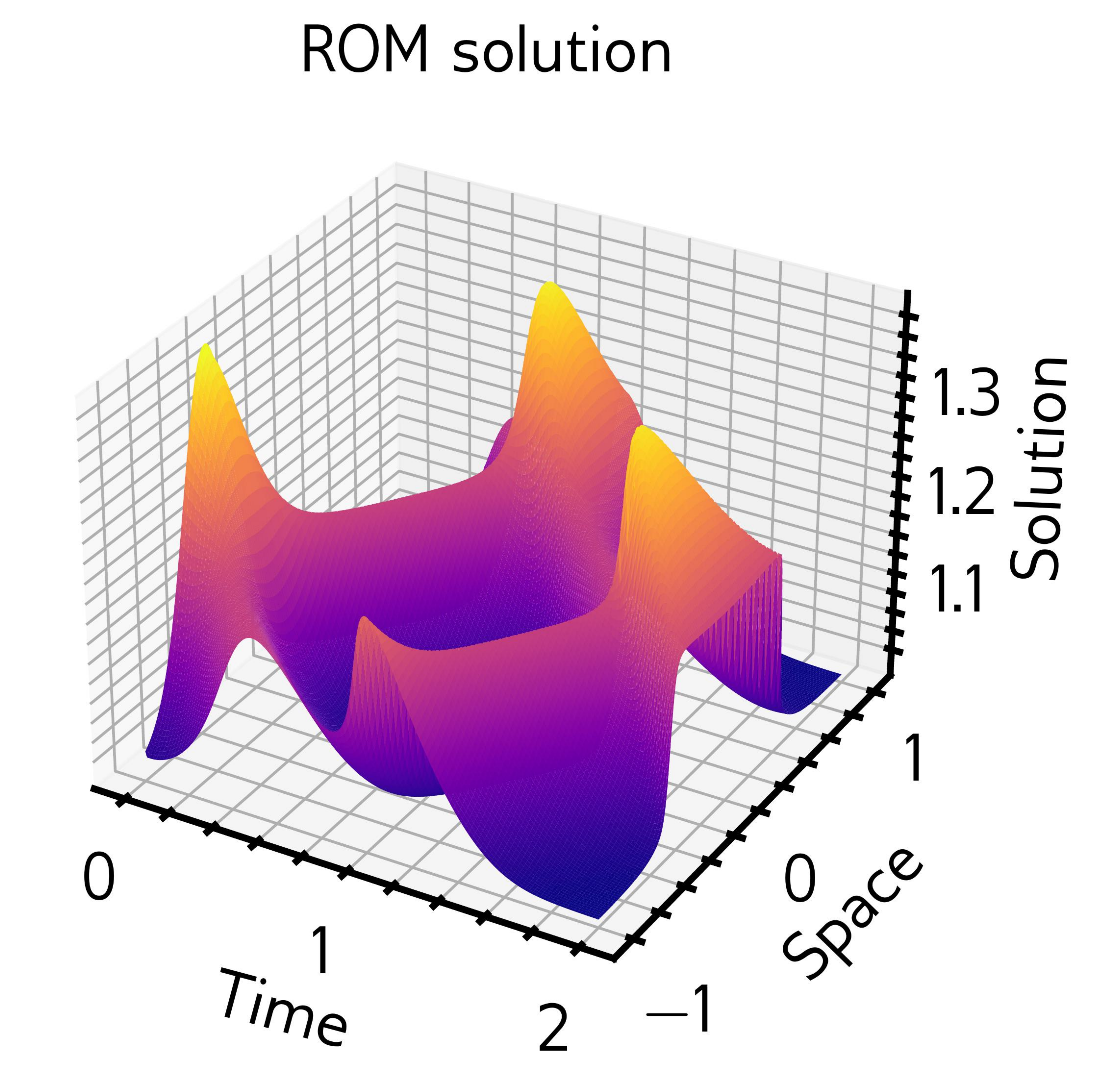}
	\end{subfigure}
	\begin{subfigure}[b]{0.32\textwidth}
		\centering
		\includegraphics[width=0.85\textwidth, trim=0 0 0 200, clip]{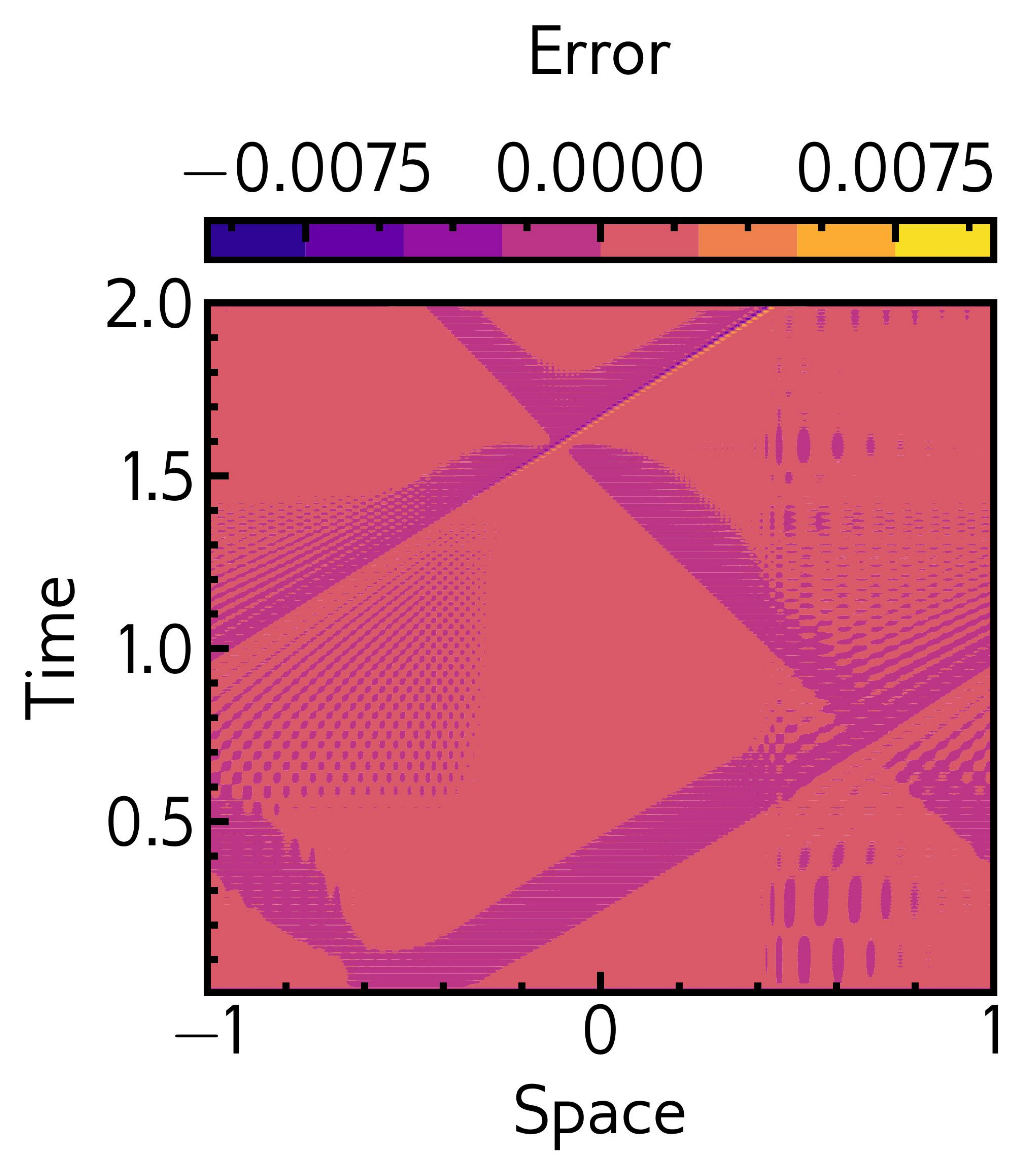}
	\end{subfigure}
	\begin{subfigure}[b]{0.32\textwidth}
		\centering
		\includegraphics[width=0.9\textwidth, trim=0 0 0 200, clip]{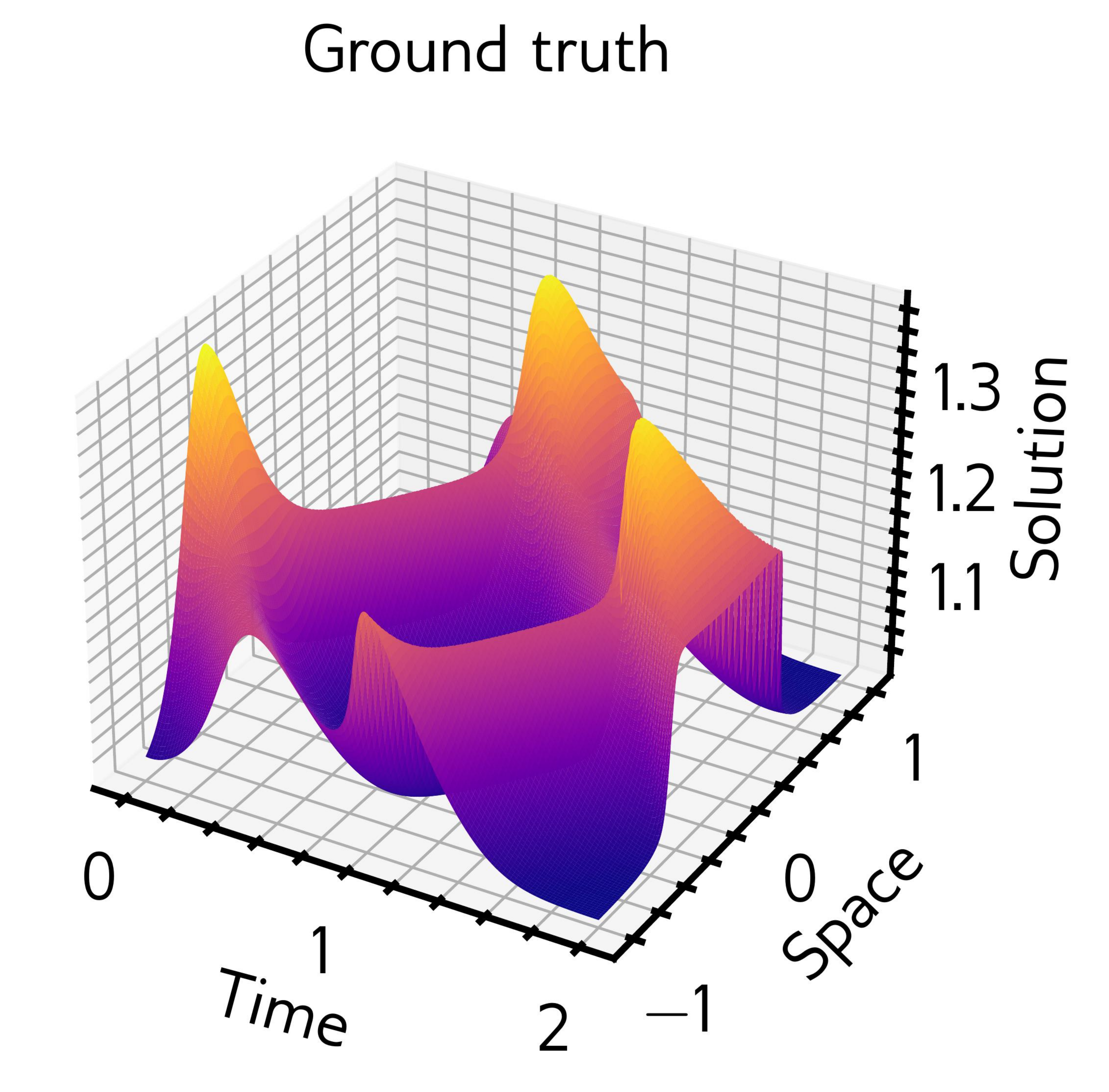}
	\end{subfigure}
	\begin{subfigure}[b]{0.32\textwidth}
		\centering
		\includegraphics[width=0.9\textwidth, trim=0 0 0 200, clip]{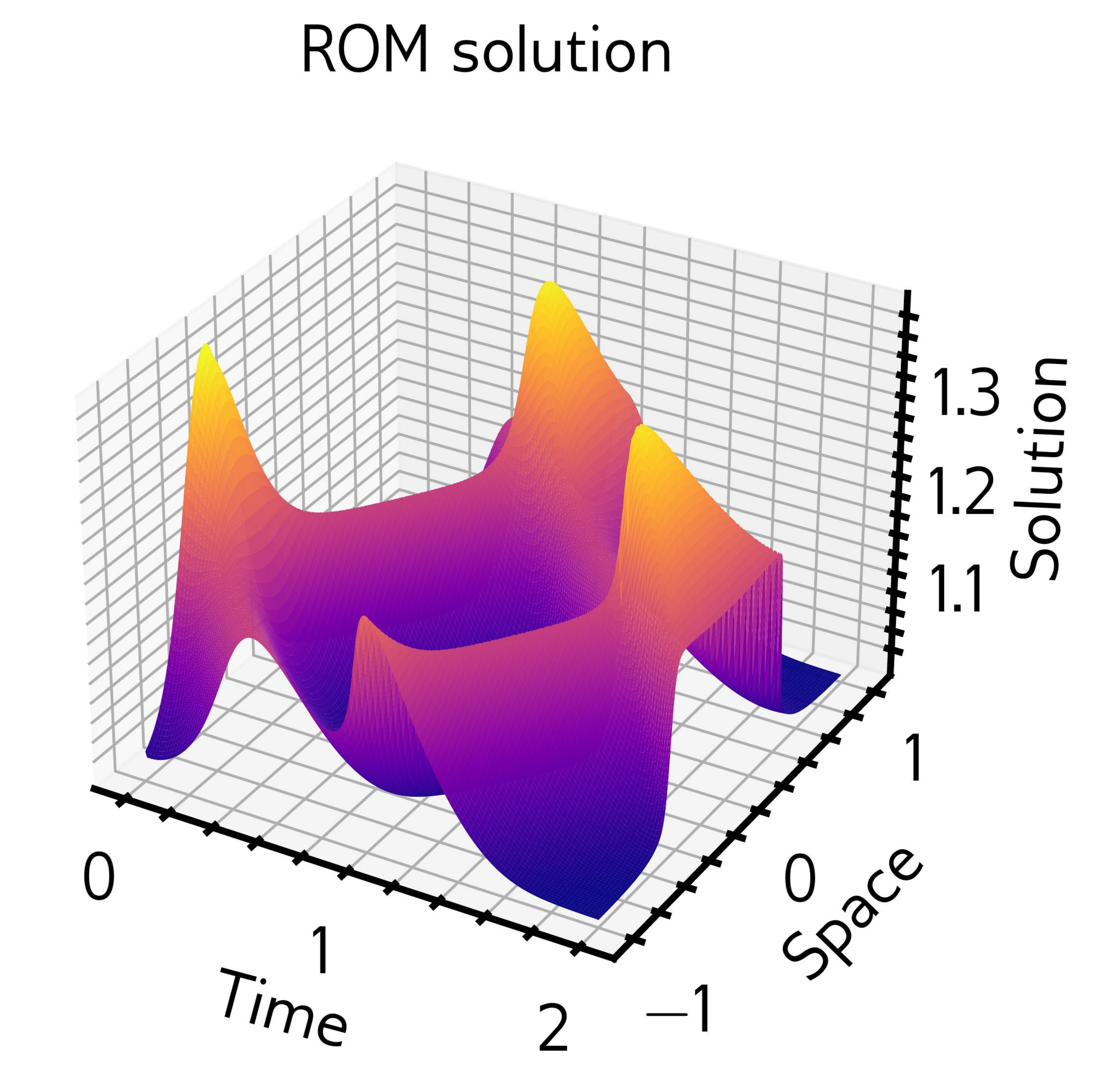}
	\end{subfigure}
	\begin{subfigure}[b]{0.32\textwidth}
		\centering
		\includegraphics[width=0.85\textwidth, trim=0 0 0 200, clip]{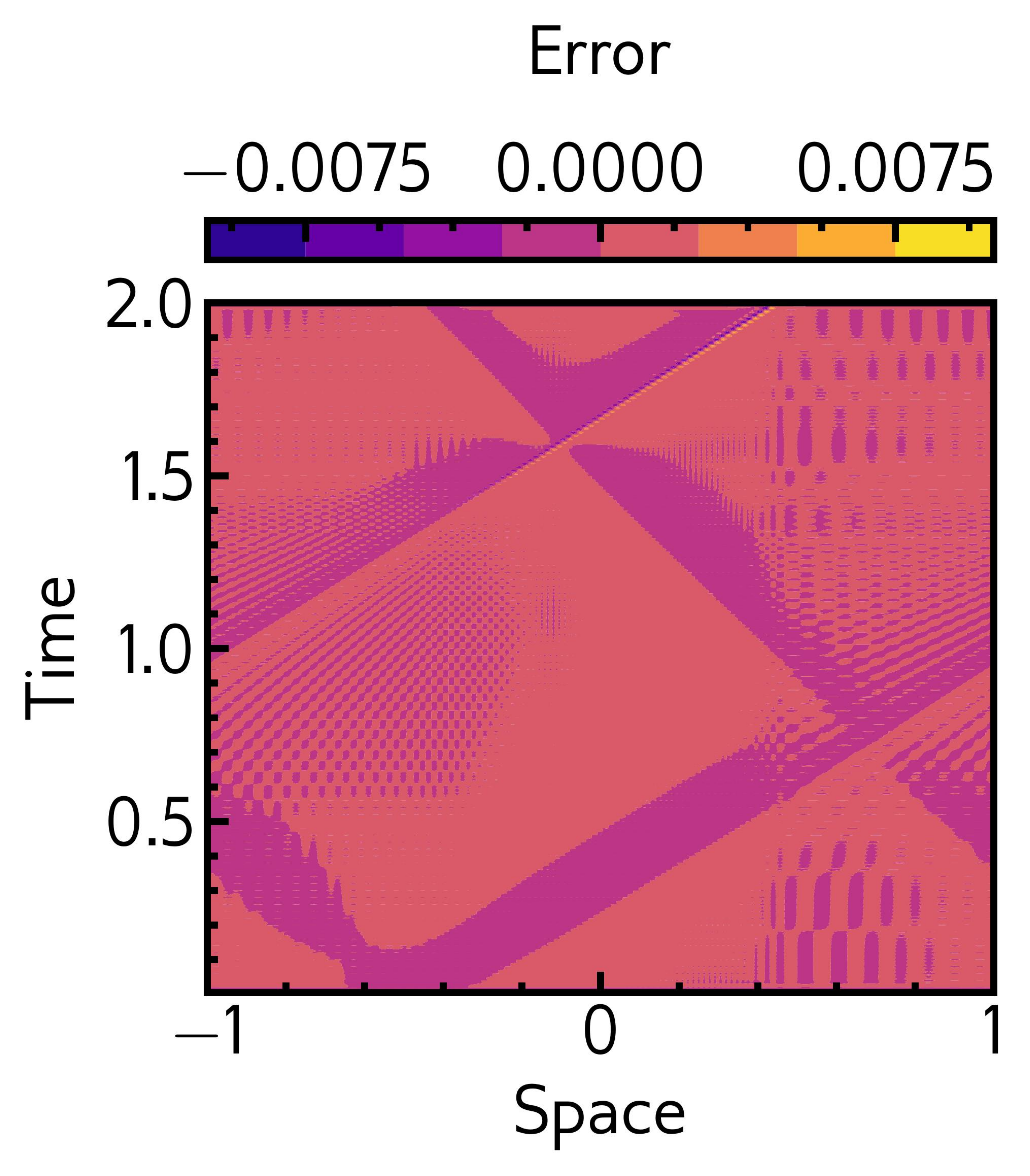}
	\end{subfigure}
	\caption{Shallow water equations' fluid height: The true solution (shown in first column), ActLearn-POD-KSNN solution (shown in second column), and the solution error (shown in third column). The error values correspond to the point-wise difference in the space-time domain between the ActLearn-POD-KSNN solution and the true solution. The viscocity $\nu$ going from top to bottom in the rows are in the following order: $\{5 \times 10^{-1}, 5 \times 10^{-2}, 5 \times 10^{-3}, 5 \times 10^{-4}, 5 \times 10^{-5}\}$. All these $\nu$ values and time instances are outside of the training set.}
	\label{fig:swe-h-rom-sol-surf}
\end{figure}

\begin{figure}[H]
	\centering
	\vspace{-0.4cm} 
	\begin{subfigure}[b]{0.32\textwidth}
		\centering
		\includegraphics[width=0.9\textwidth, trim=0 0 0 200, clip]{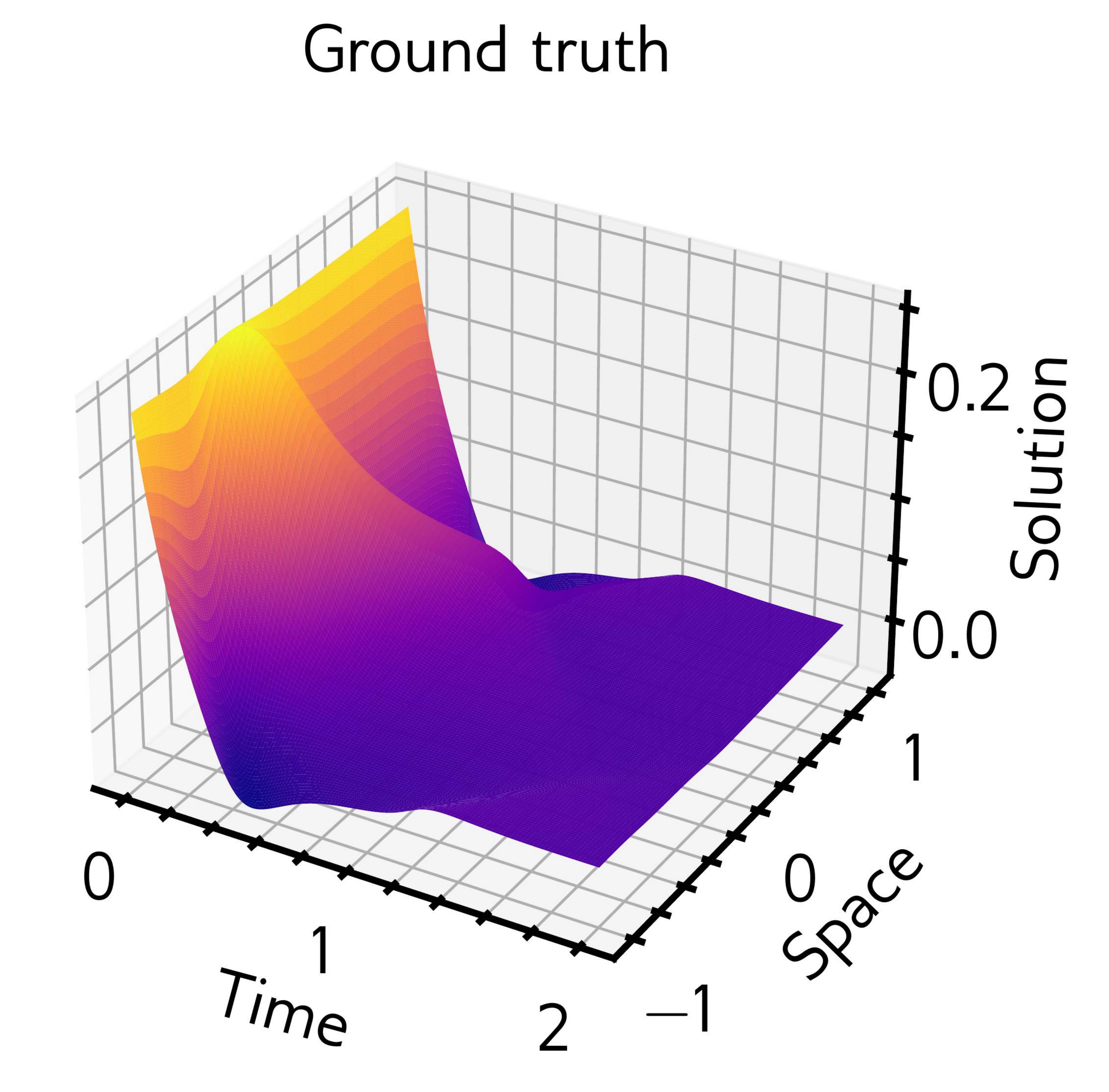}
	\end{subfigure}
	\begin{subfigure}[b]{0.32\textwidth}
		\centering
		\includegraphics[width=0.9\textwidth, trim=0 0 0 200, clip]{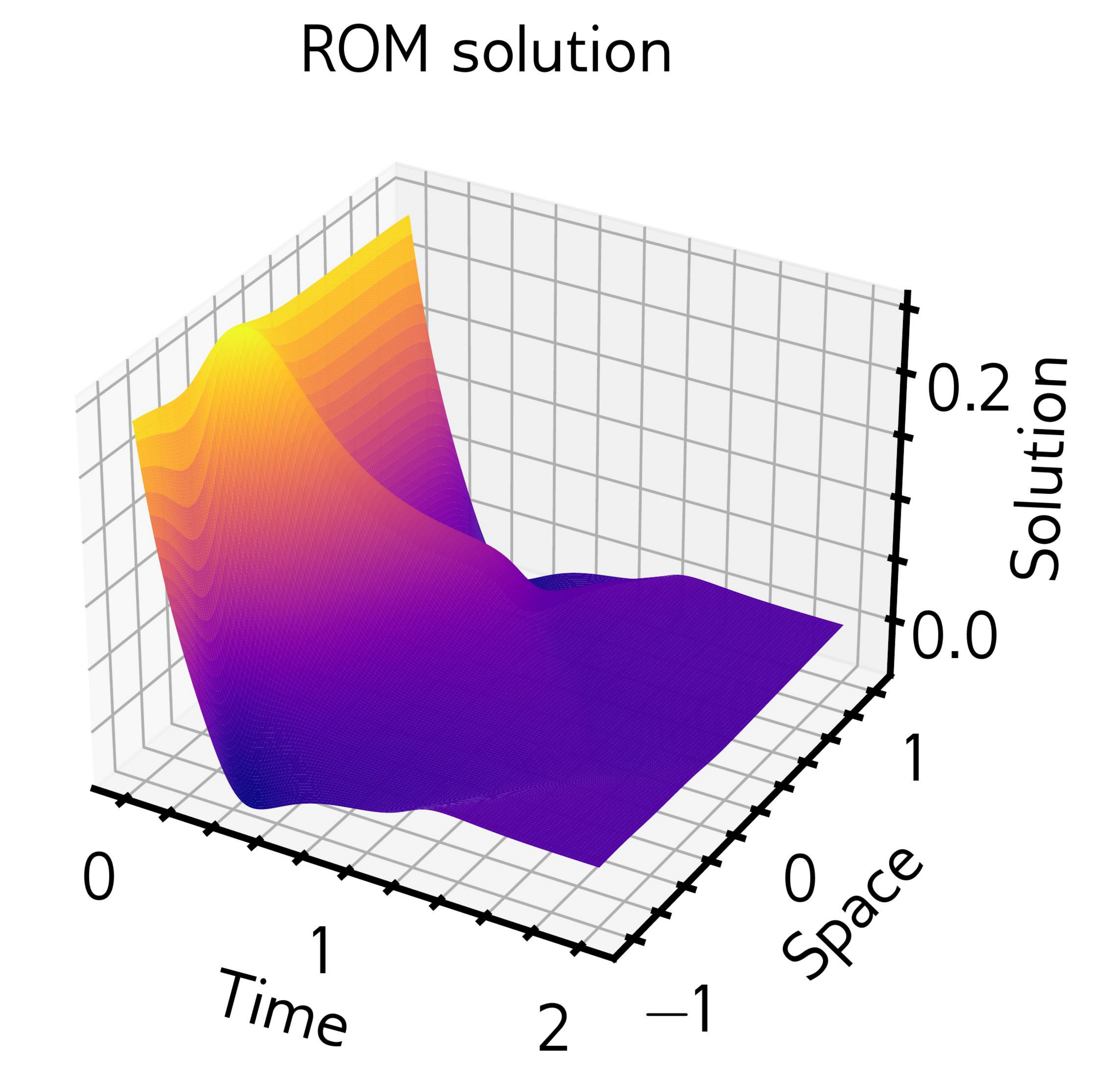}
	\end{subfigure}
	\begin{subfigure}[b]{0.32\textwidth}
		\centering
		\includegraphics[width=0.85\textwidth, trim=0 0 0 200, clip]{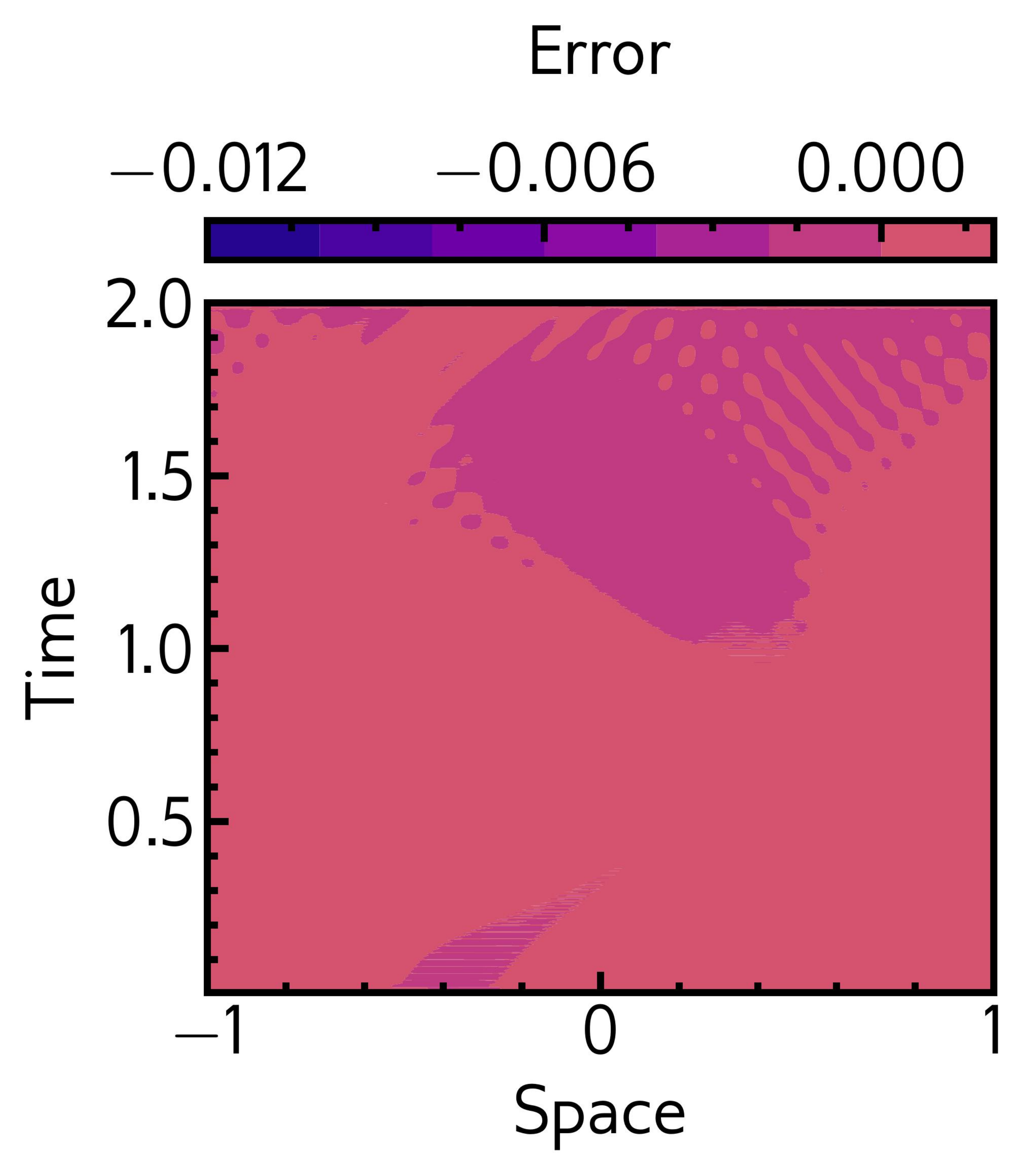}
	\end{subfigure}
	\begin{subfigure}[b]{0.32\textwidth}
		\centering
		\includegraphics[width=0.9\textwidth, trim=0 0 0 200, clip]{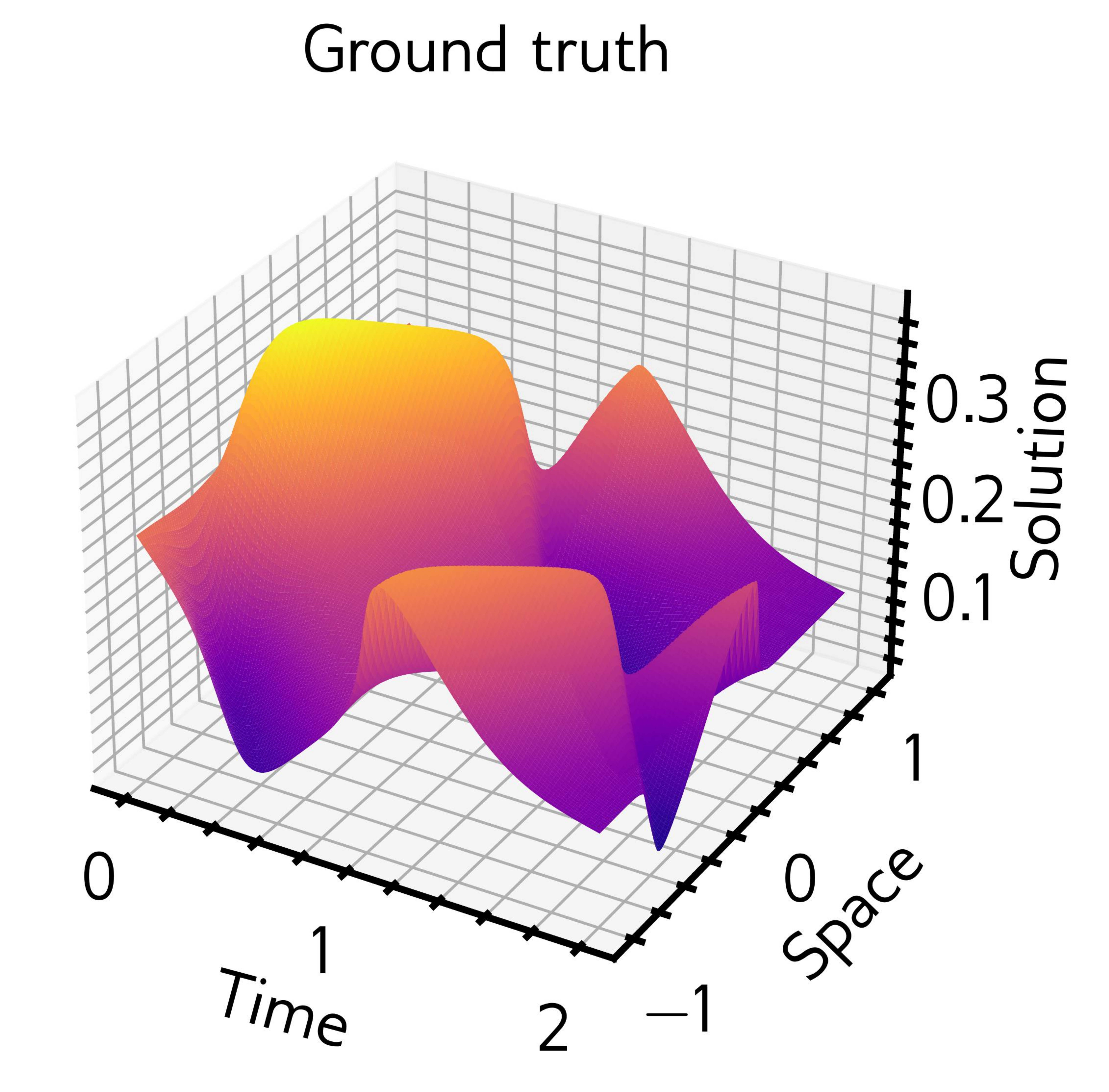}
	\end{subfigure}
	\begin{subfigure}[b]{0.32\textwidth}
		\centering
		\includegraphics[width=0.9\textwidth, trim=0 0 0 200, clip]{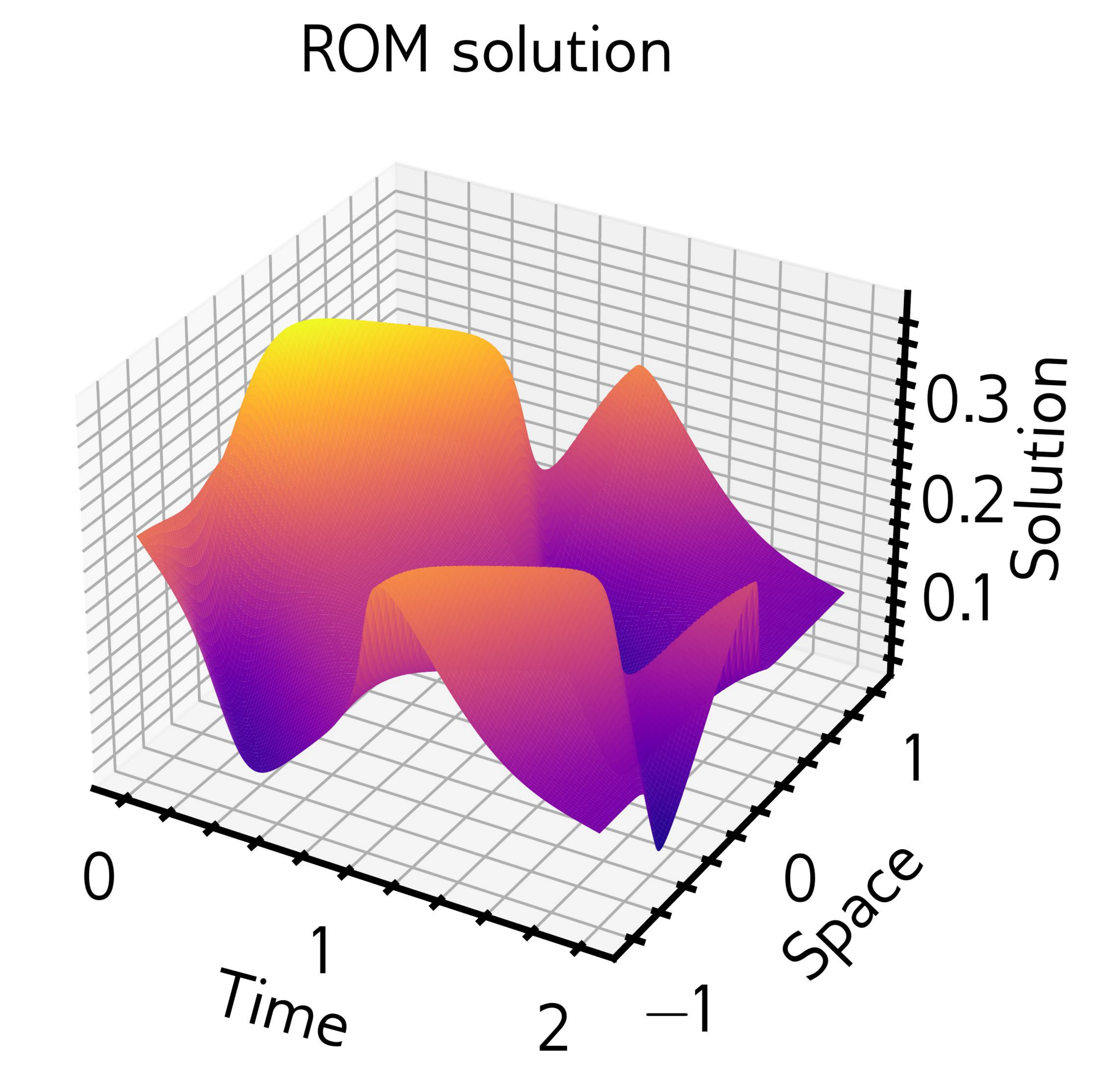}
	\end{subfigure}
	\begin{subfigure}[b]{0.32\textwidth}
		\centering
		\includegraphics[width=0.85\textwidth, trim=0 0 0 200, clip]{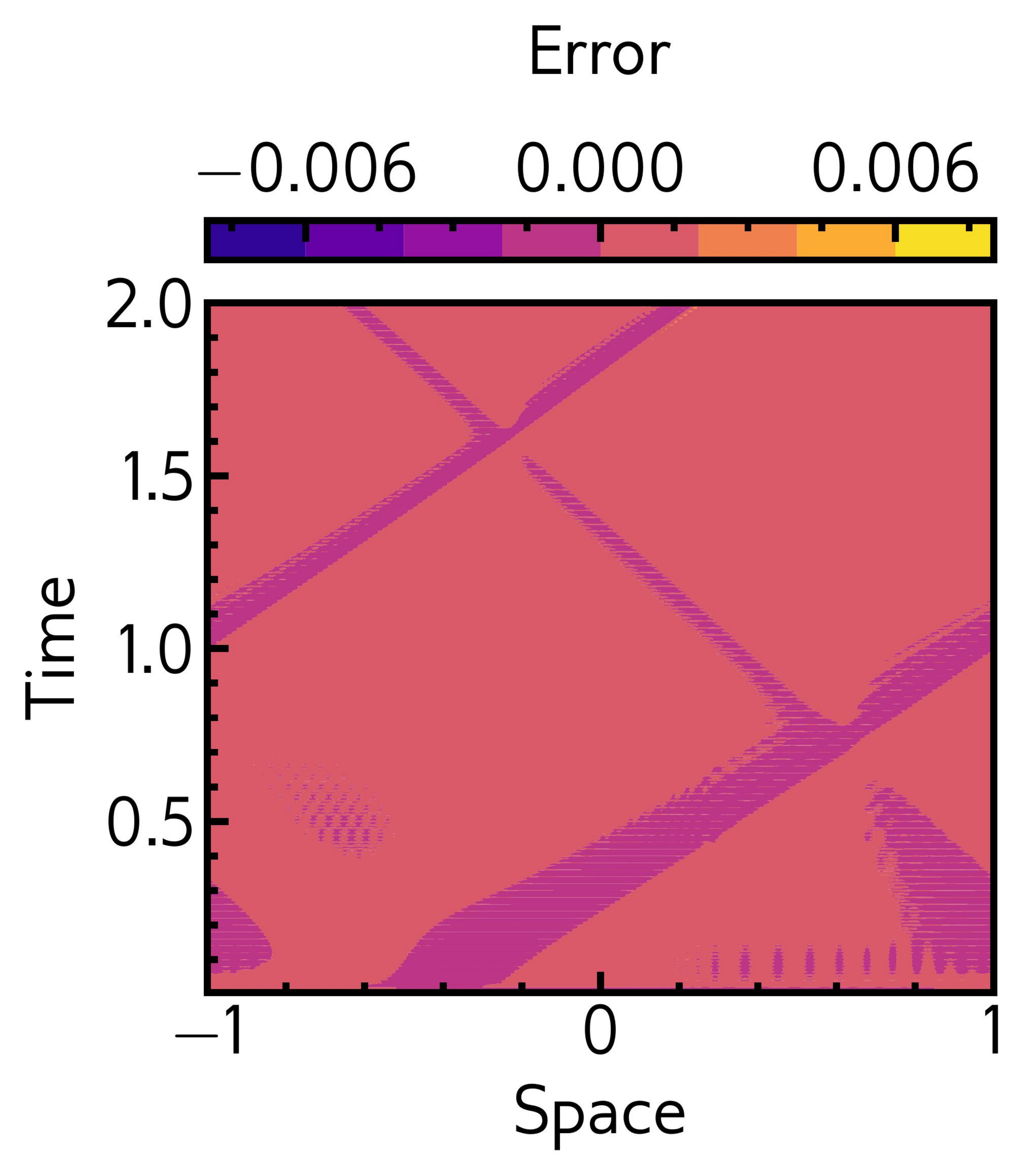}
	\end{subfigure}
	\begin{subfigure}[b]{0.32\textwidth}
		\centering
		\includegraphics[width=0.9\textwidth, trim=0 0 0 200, clip]{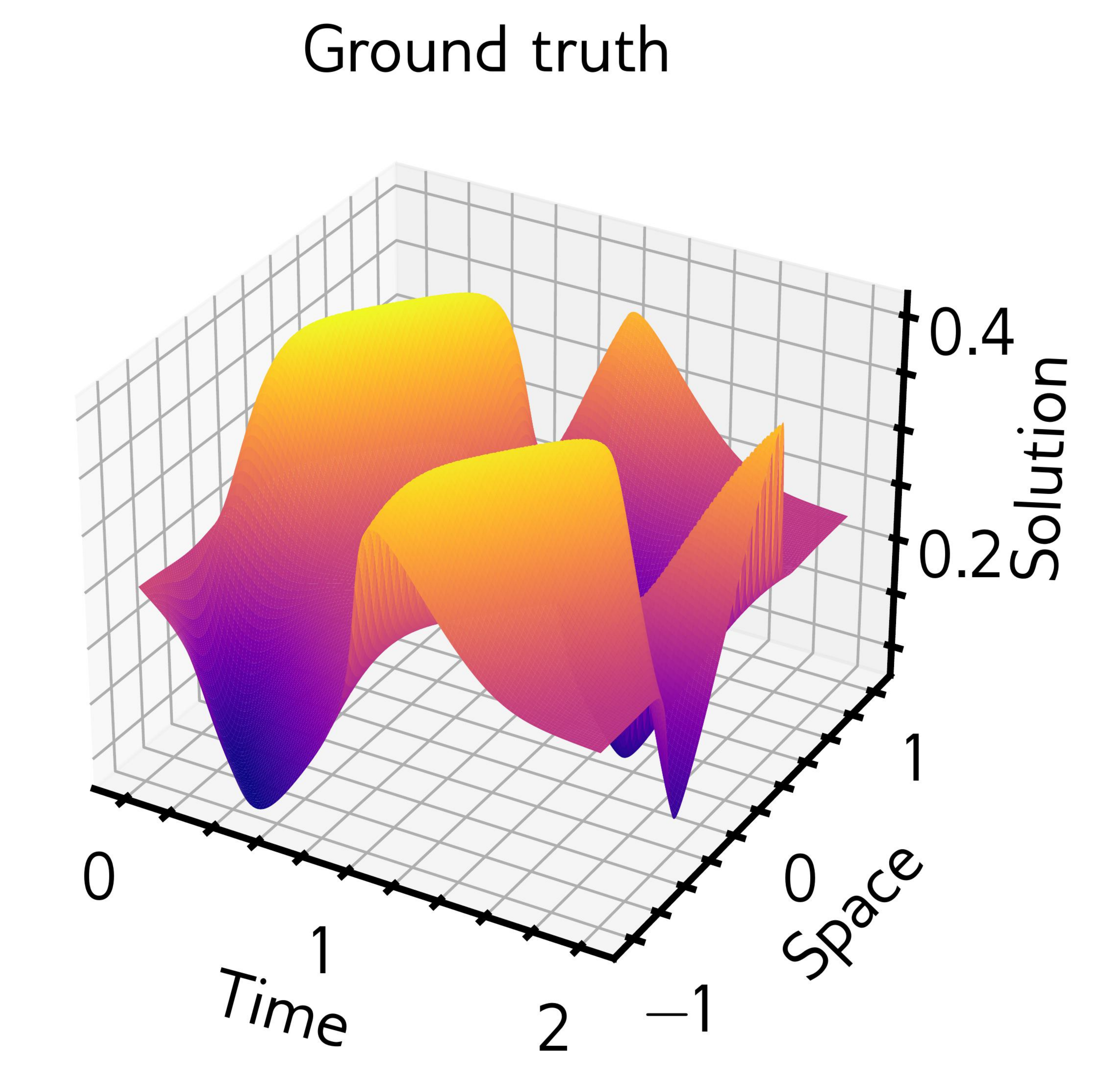}
	\end{subfigure}
	\begin{subfigure}[b]{0.32\textwidth}
		\centering
		\includegraphics[width=0.9\textwidth, trim=0 0 0 200, clip]{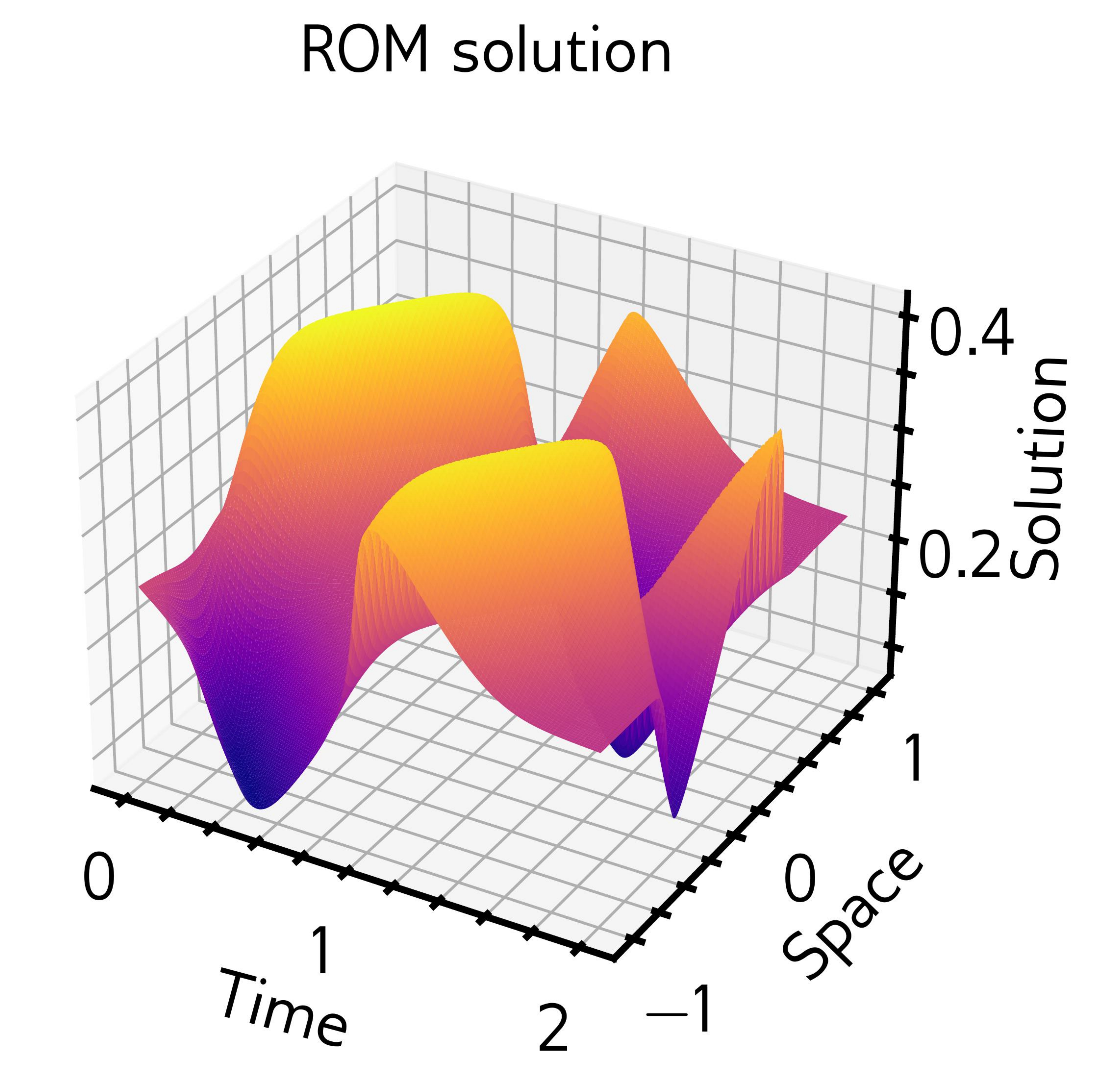}
	\end{subfigure}
	\begin{subfigure}[b]{0.32\textwidth}
		\centering
		\includegraphics[width=0.85\textwidth, trim=0 0 0 200, clip]{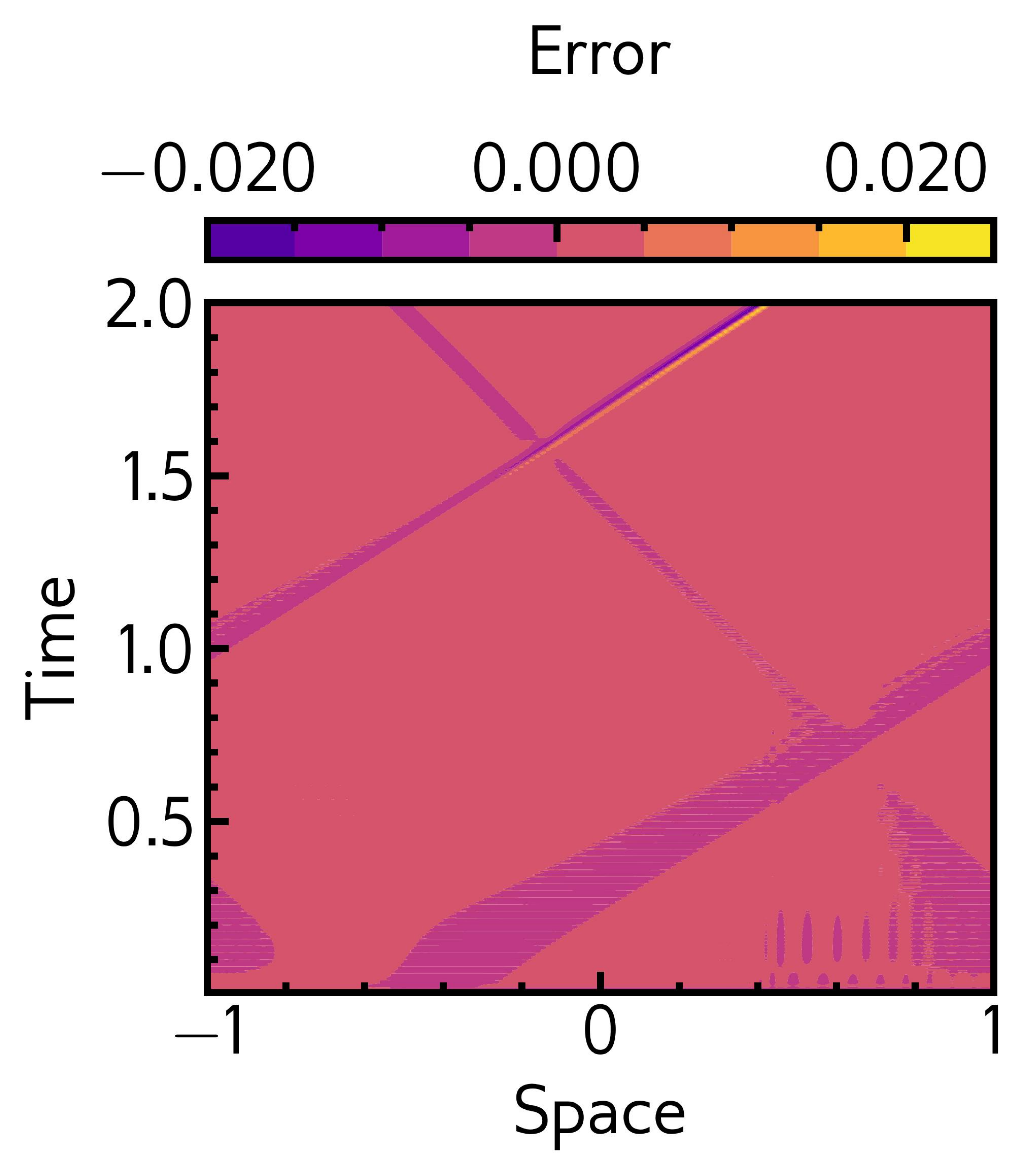}
	\end{subfigure}
	\begin{subfigure}[b]{0.32\textwidth}
		\centering
		\includegraphics[width=0.9\textwidth, trim=0 0 0 200, clip]{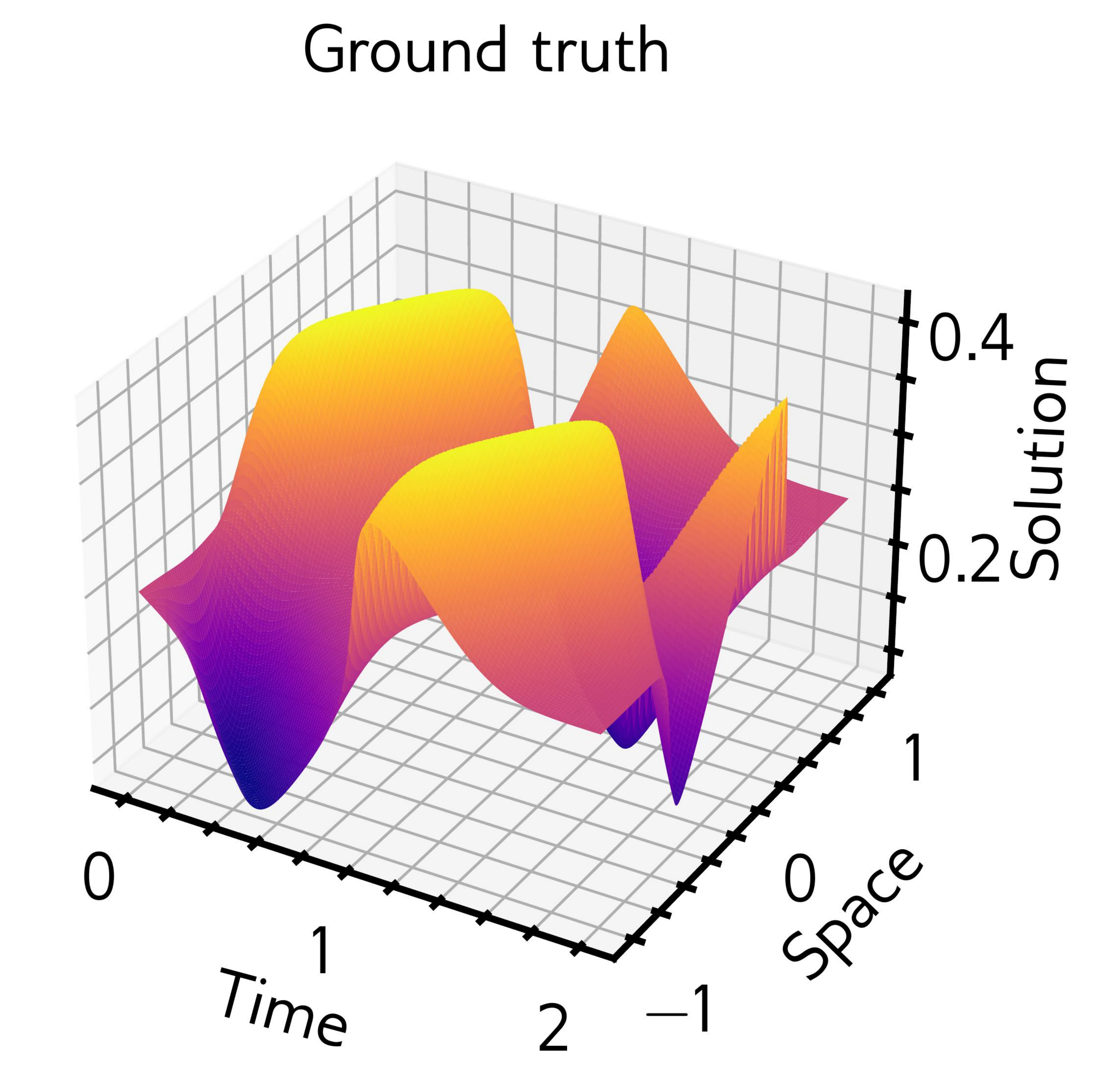}
	\end{subfigure}
	\begin{subfigure}[b]{0.32\textwidth}
		\centering
		\includegraphics[width=0.9\textwidth, trim=0 0 0 200, clip]{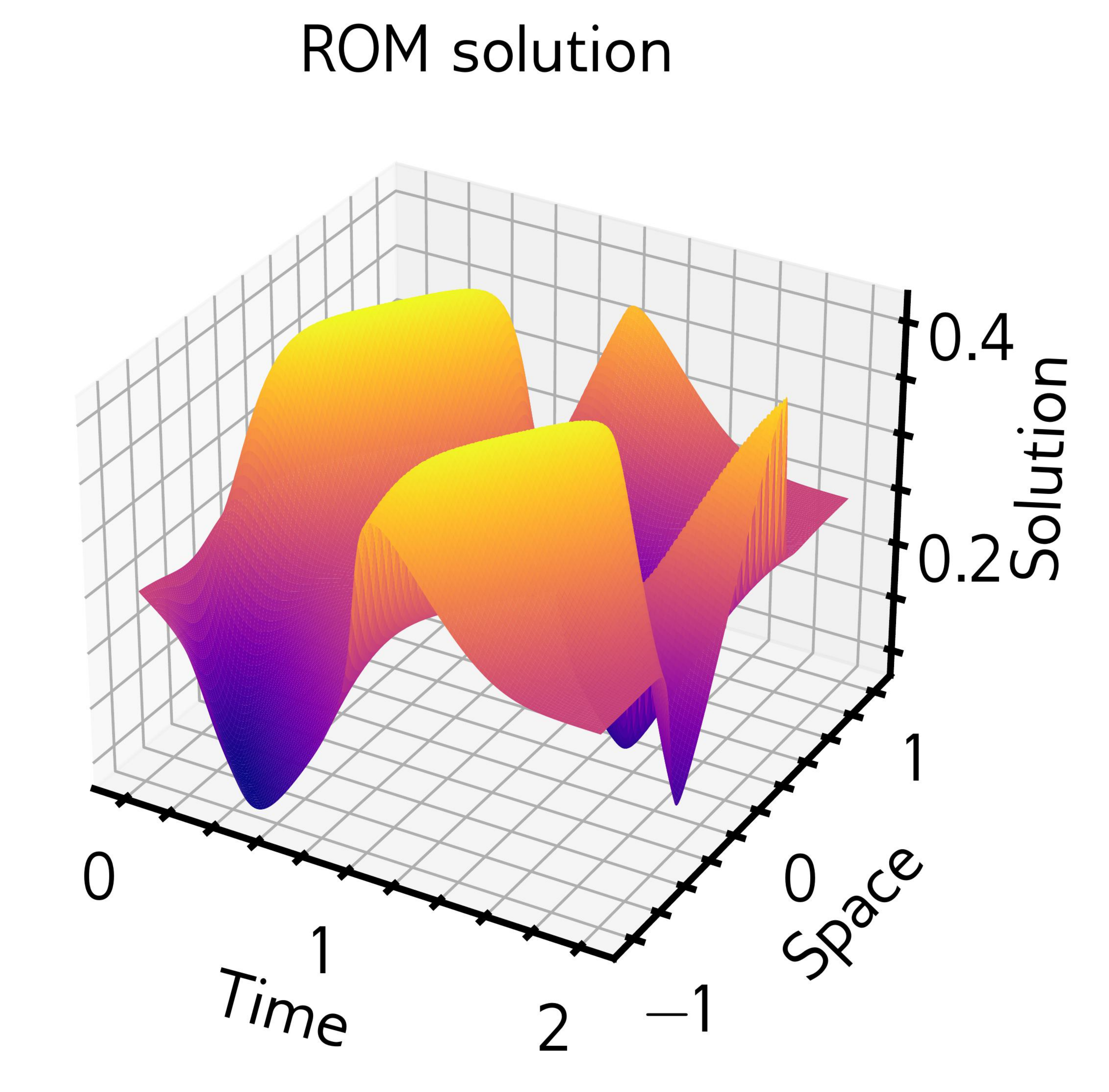}
	\end{subfigure}
	\begin{subfigure}[b]{0.32\textwidth}
		\centering
		\includegraphics[width=0.85\textwidth, trim=0 0 0 200, clip]{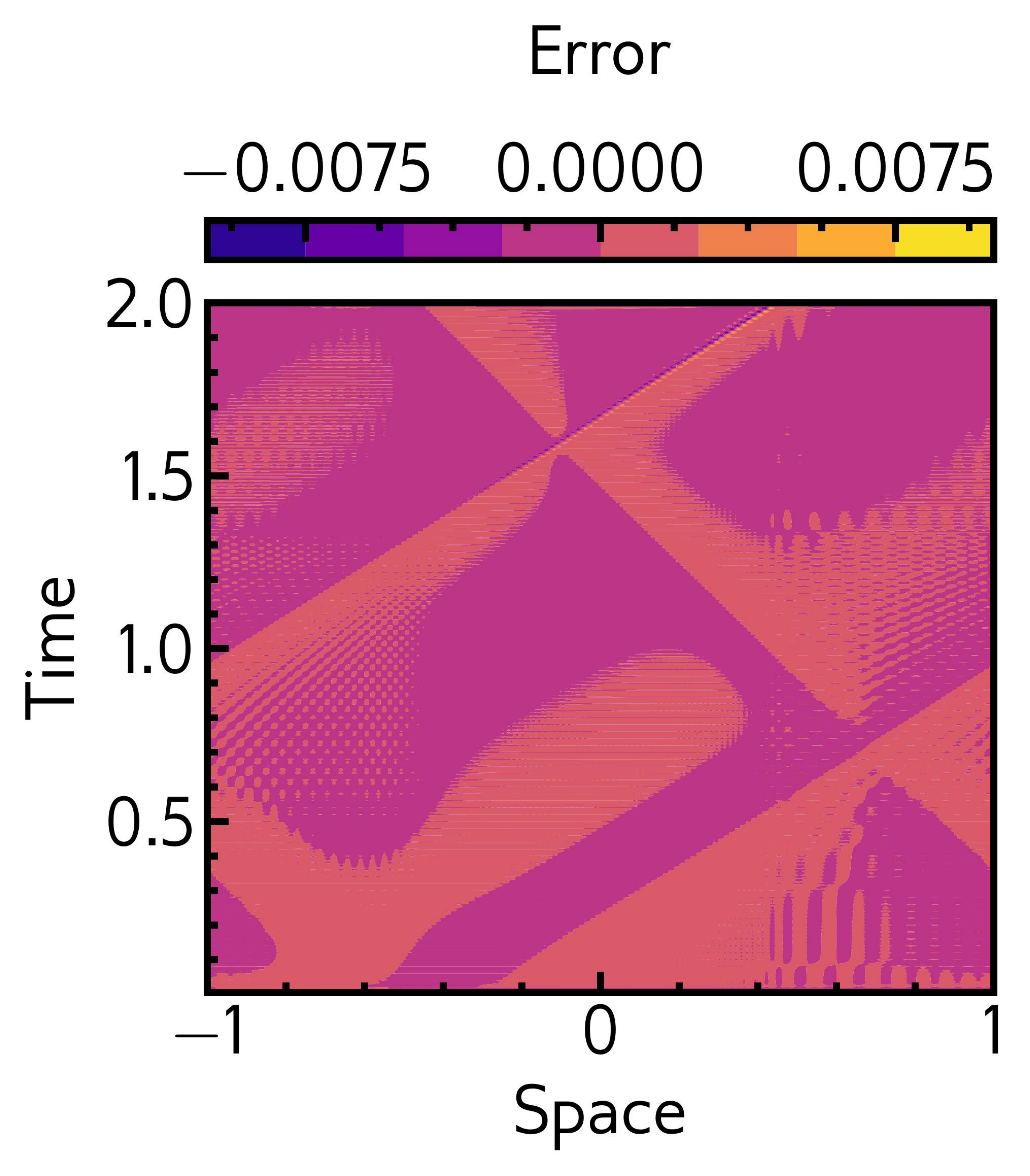}
	\end{subfigure}
	\begin{subfigure}[b]{0.32\textwidth}
		\centering
		\includegraphics[width=0.9\textwidth, trim=0 0 0 200, clip]{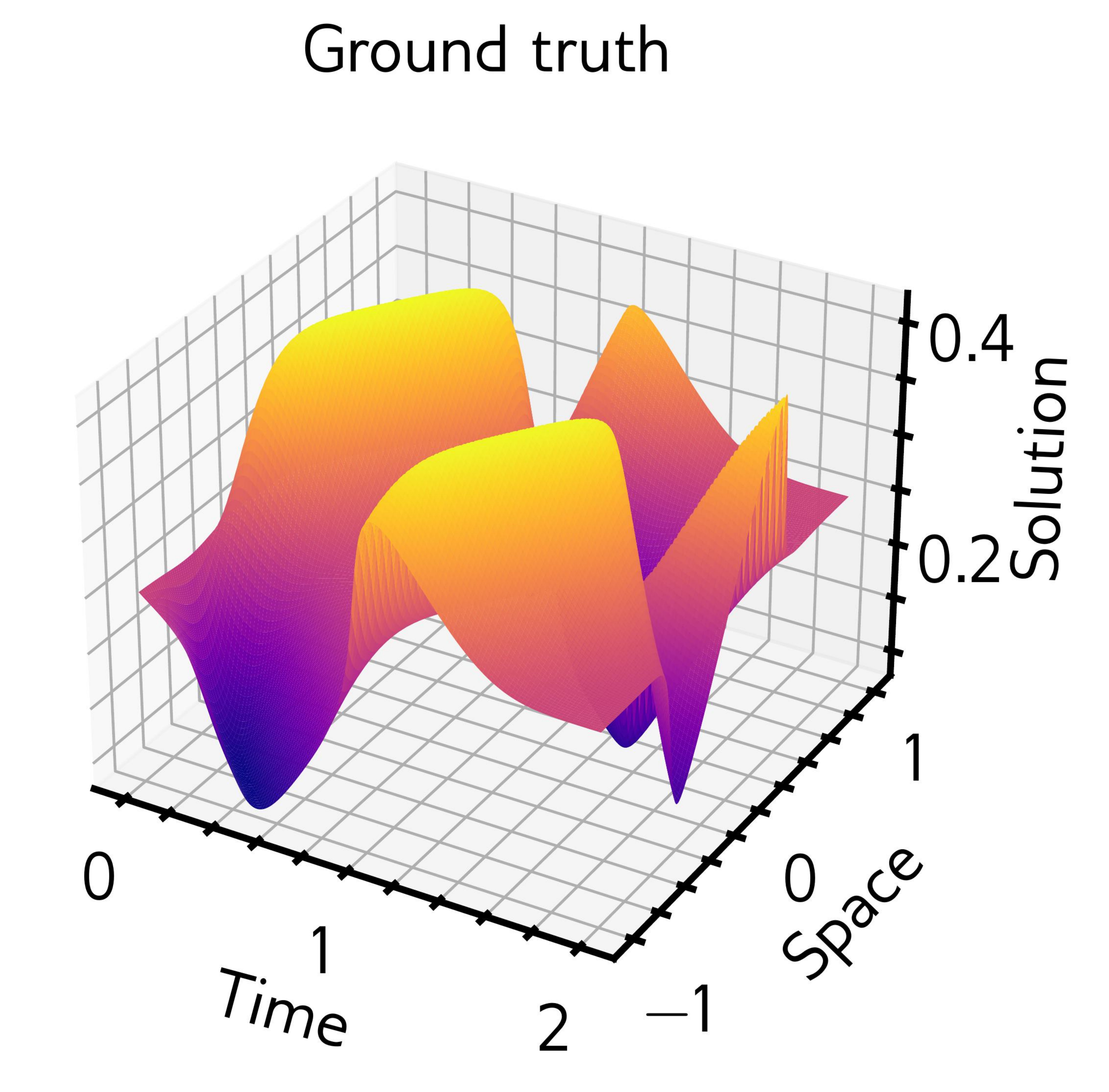}
	\end{subfigure}
	\begin{subfigure}[b]{0.32\textwidth}
		\centering
		\includegraphics[width=0.9\textwidth, trim=0 0 0 200, clip]{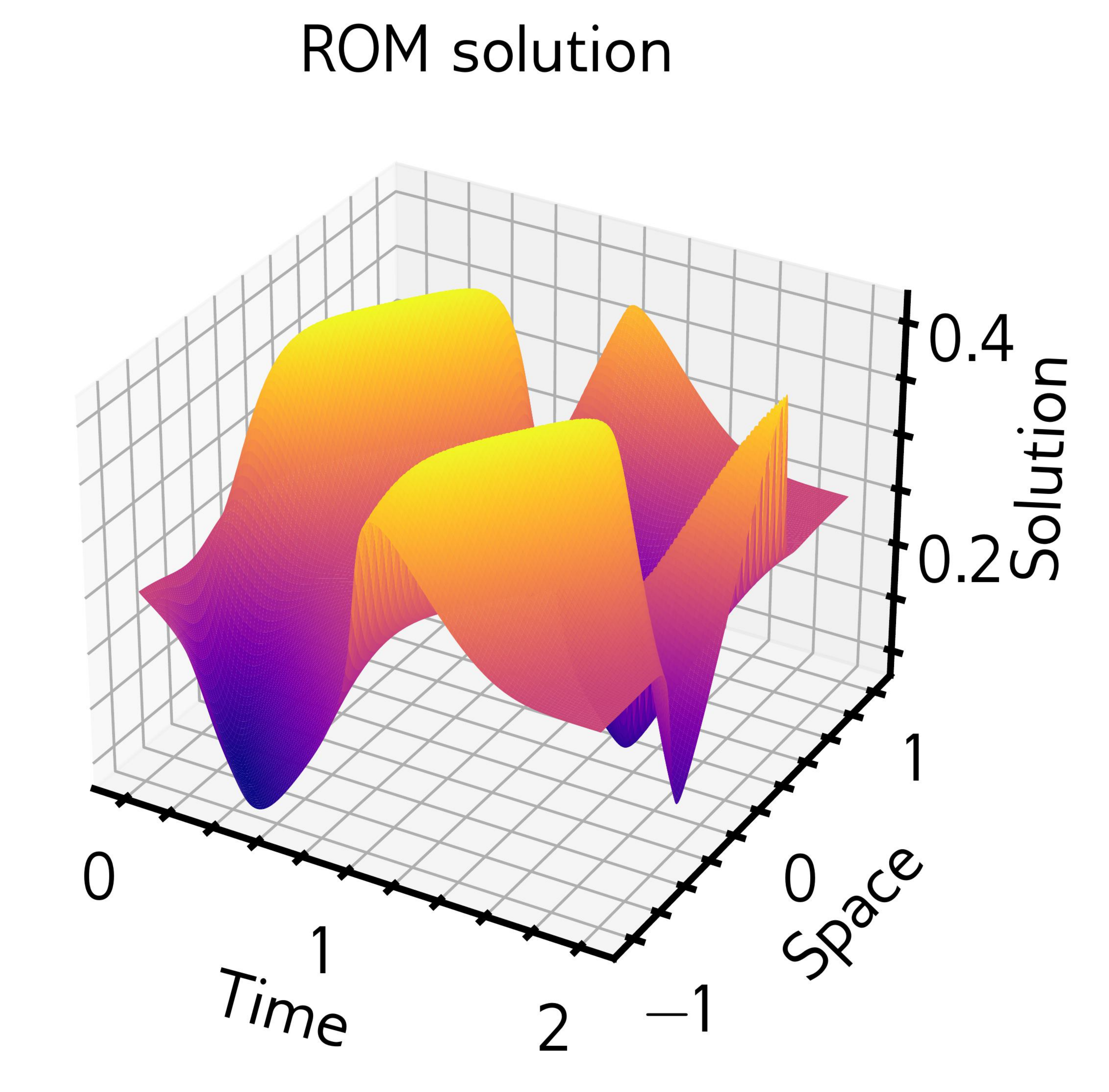}
	\end{subfigure}
	\begin{subfigure}[b]{0.32\textwidth}
		\centering
		\includegraphics[width=0.85\textwidth, trim=0 0 0 200, clip]{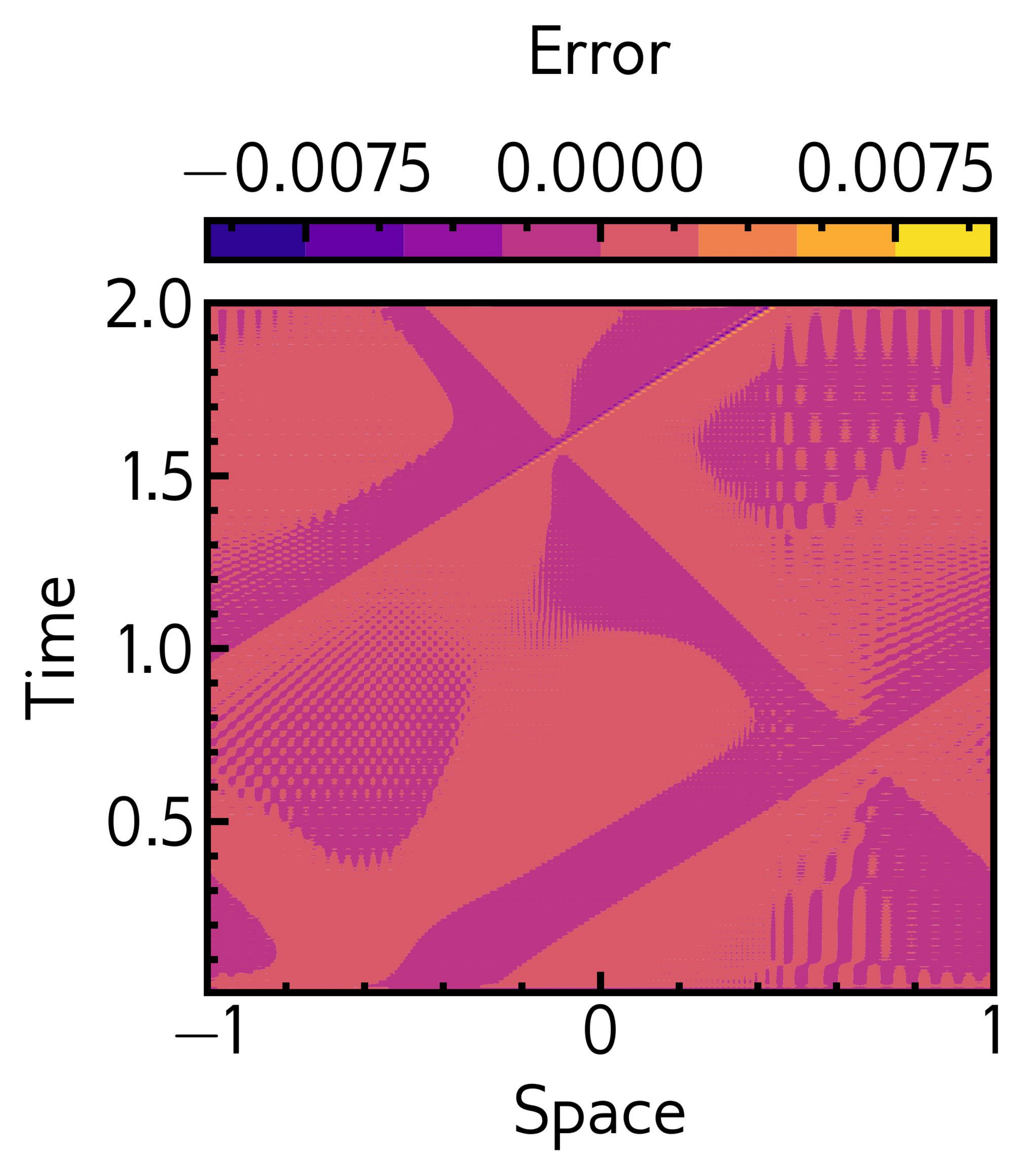}
	\end{subfigure}
	\caption{Shallow water equations' fluid velocity: The true solution (shown in first column), ActLearn-POD-KSNN solution (shown in second column), and the solution error (shown in third column). The error values correspond to the point-wise difference in the space-time domain between the ActLearn-POD-KSNN solution and the true solution. The viscosity $\nu$ going from top to bottom in the rows are in the following order: $\{5 \times 10^{-1}, 5 \times 10^{-2}, 5 \times 10^{-3}, 5 \times 10^{-4}, 5 \times 10^{-5}\}$. All these $\nu$ values and time instances are outside of the training set.}
	\label{fig:swe-u-rom-sol-surf}
\end{figure}

\begin{figure}[H]
	\centering
	\begin{subfigure}[b]{0.32\textwidth}
		\centering
		\includegraphics[width=1\textwidth, trim=0 0 0 0, clip]{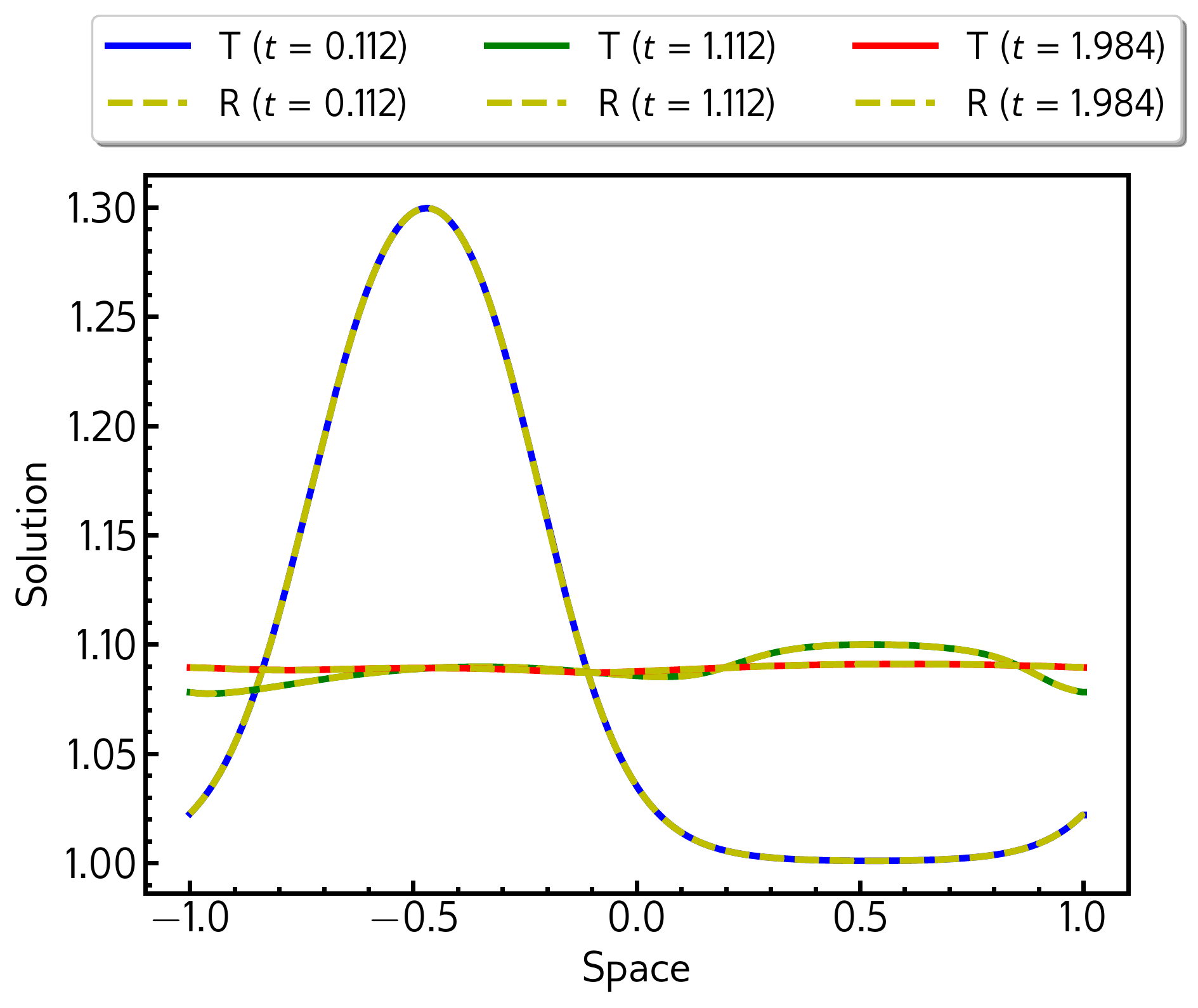}
		\caption{$\nu = 5 \times 10^{-1}$}
		\label{fig:swe-h-rom-sol-a}
	\end{subfigure}
	\begin{subfigure}[b]{0.32\textwidth}
		\centering
		\includegraphics[width=1\textwidth, trim=0 0 0 0, clip]{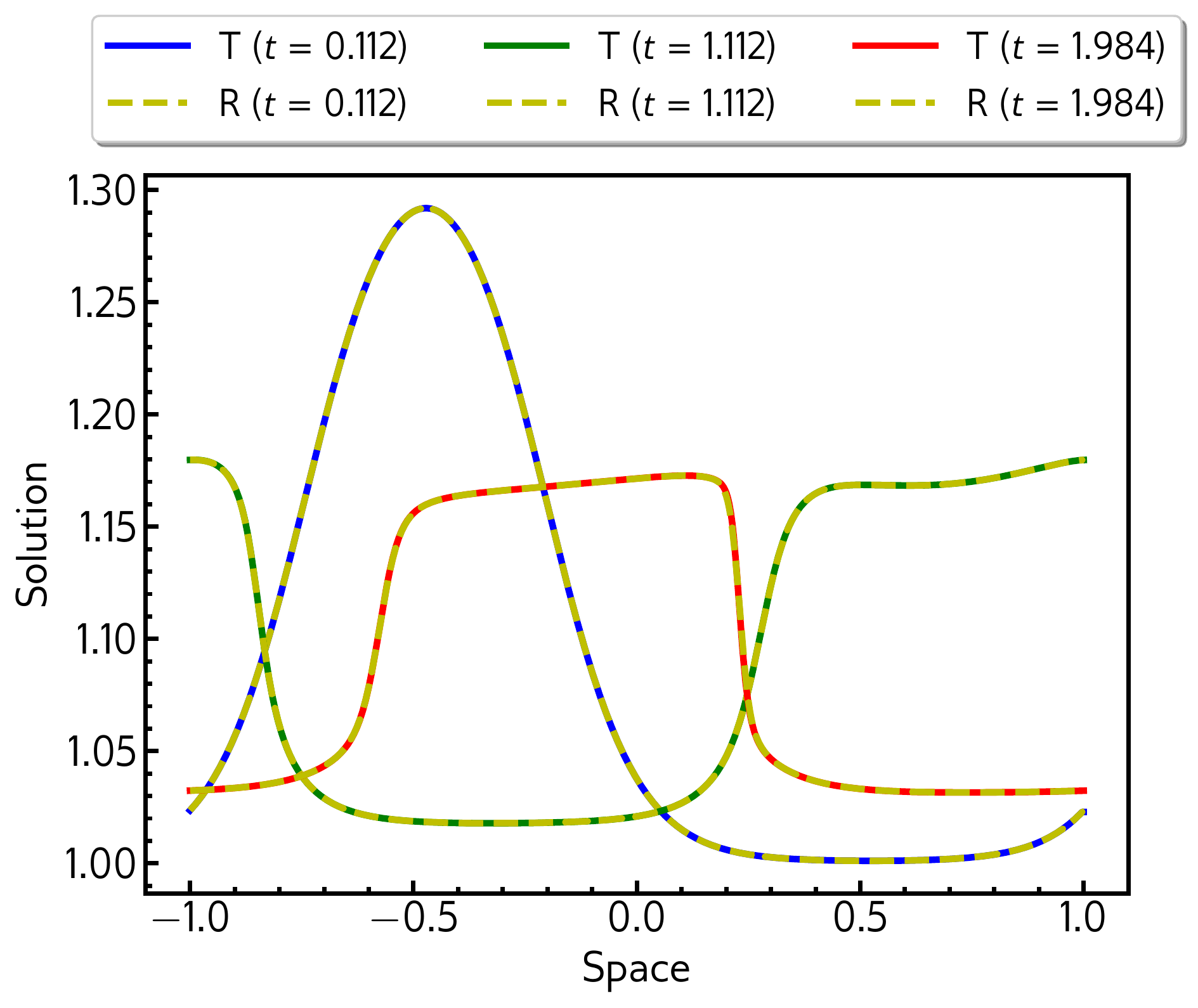}
		\caption{$\nu = 5 \times 10^{-2}$}
		\label{fig:swe-h-rom-sol-b}
	\end{subfigure}
	\begin{subfigure}[b]{0.32\textwidth}
		\centering
		\includegraphics[width=1\textwidth, trim=0 0 0 0, clip]{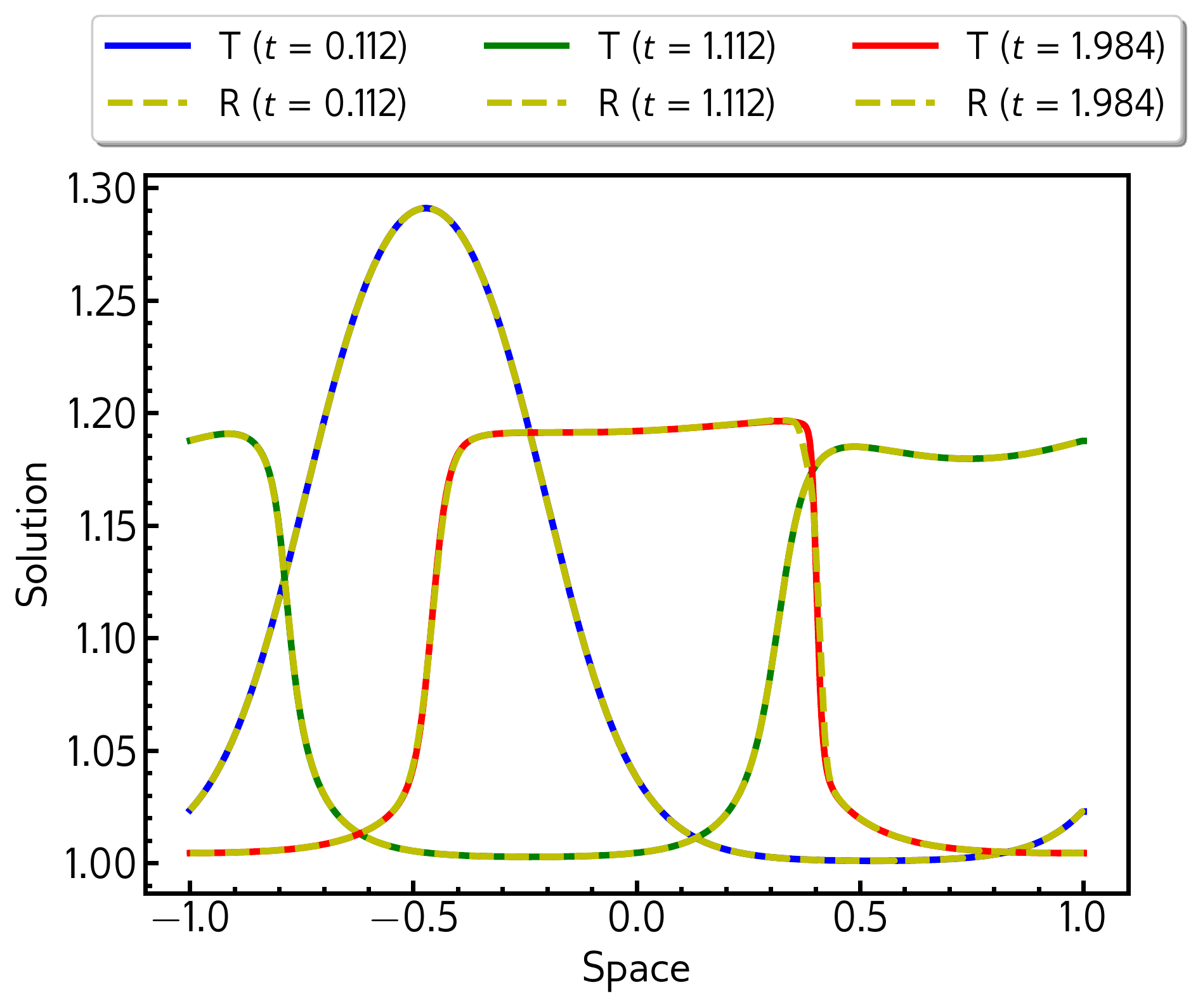}
		\caption{$\nu = 5 \times 10^{-3}$}
		\label{fig:swe-h-rom-sol-c}
	\end{subfigure}
	\begin{subfigure}[b]{0.32\textwidth}
		\centering
		\includegraphics[width=1\textwidth, trim=0 0 0 0, clip]{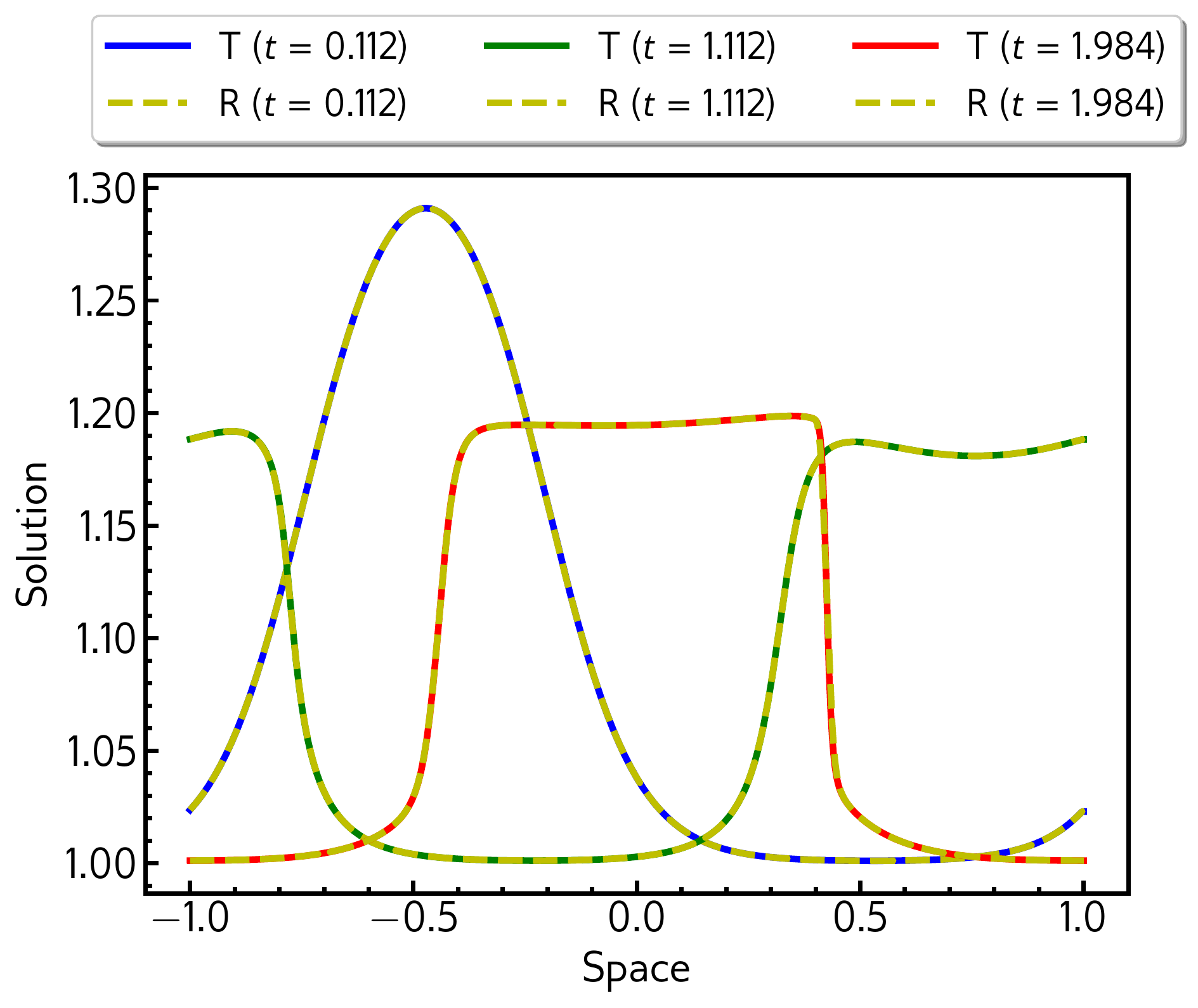}
		\caption{$\nu = 5 \times 10^{-4}$}
		\label{fig:swe-h-rom-sol-d}
	\end{subfigure}
	\begin{subfigure}[b]{0.32\textwidth}
		\centering
		\includegraphics[width=1\textwidth, trim=0 0 0 0, clip]{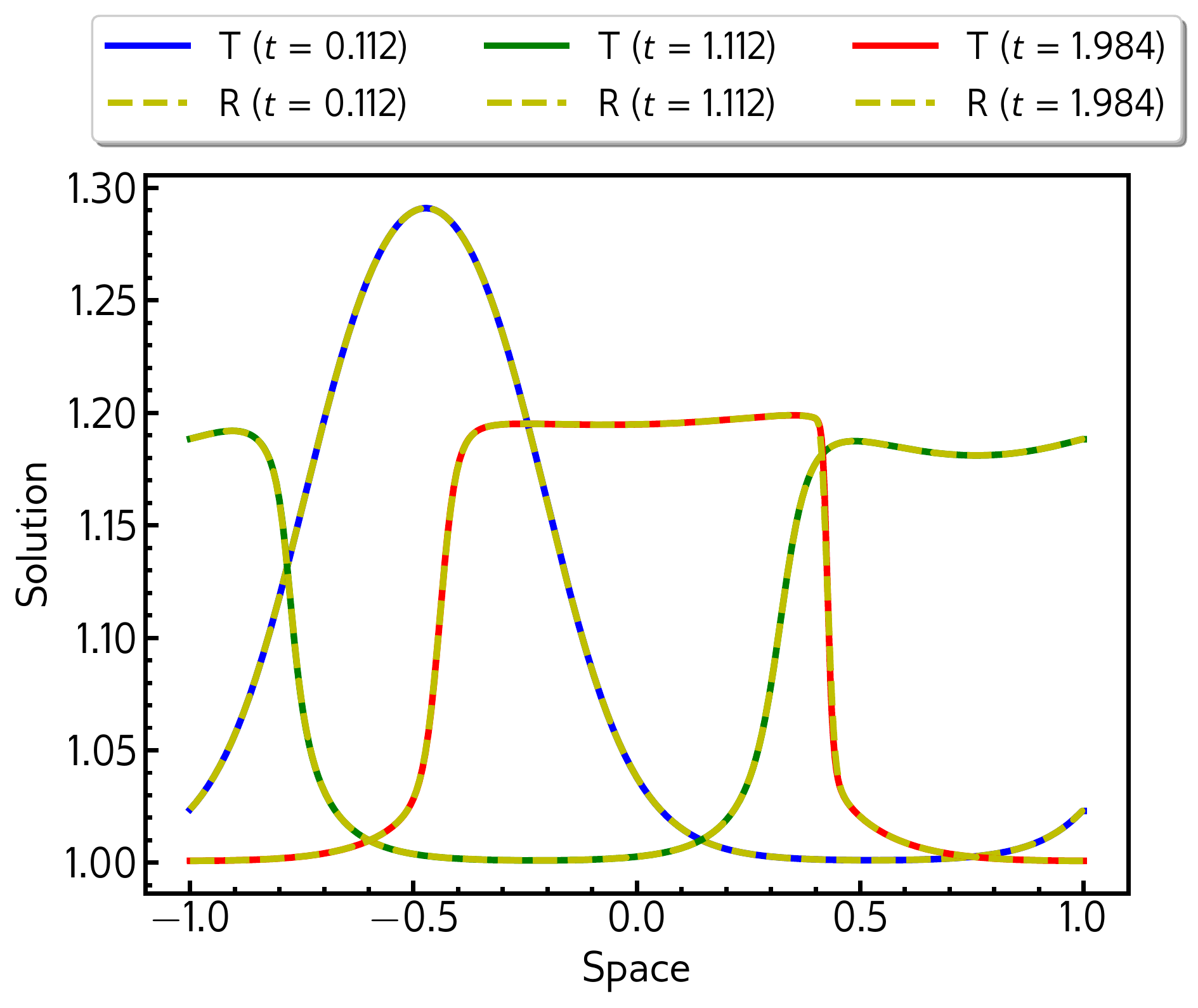}
		\caption{$\nu = 5 \times 10^{-5}$}
		\label{fig:swe-h-rom-sol-e}
	\end{subfigure}
	\caption{Comparison of the ActLearn-POD-KSNN solution (denoted by R) and true solution (denoted by T) for the fluid height in the shallow water equations. All the $\nu$ values and time instances $t$ are outside of the training set.}
	\label{fig:swe-h-rom-sol}
\end{figure}

\begin{figure}[H]
	\centering
	\begin{subfigure}[b]{0.32\textwidth}
		\centering
		\includegraphics[width=1\textwidth, trim=0 0 0 0, clip]{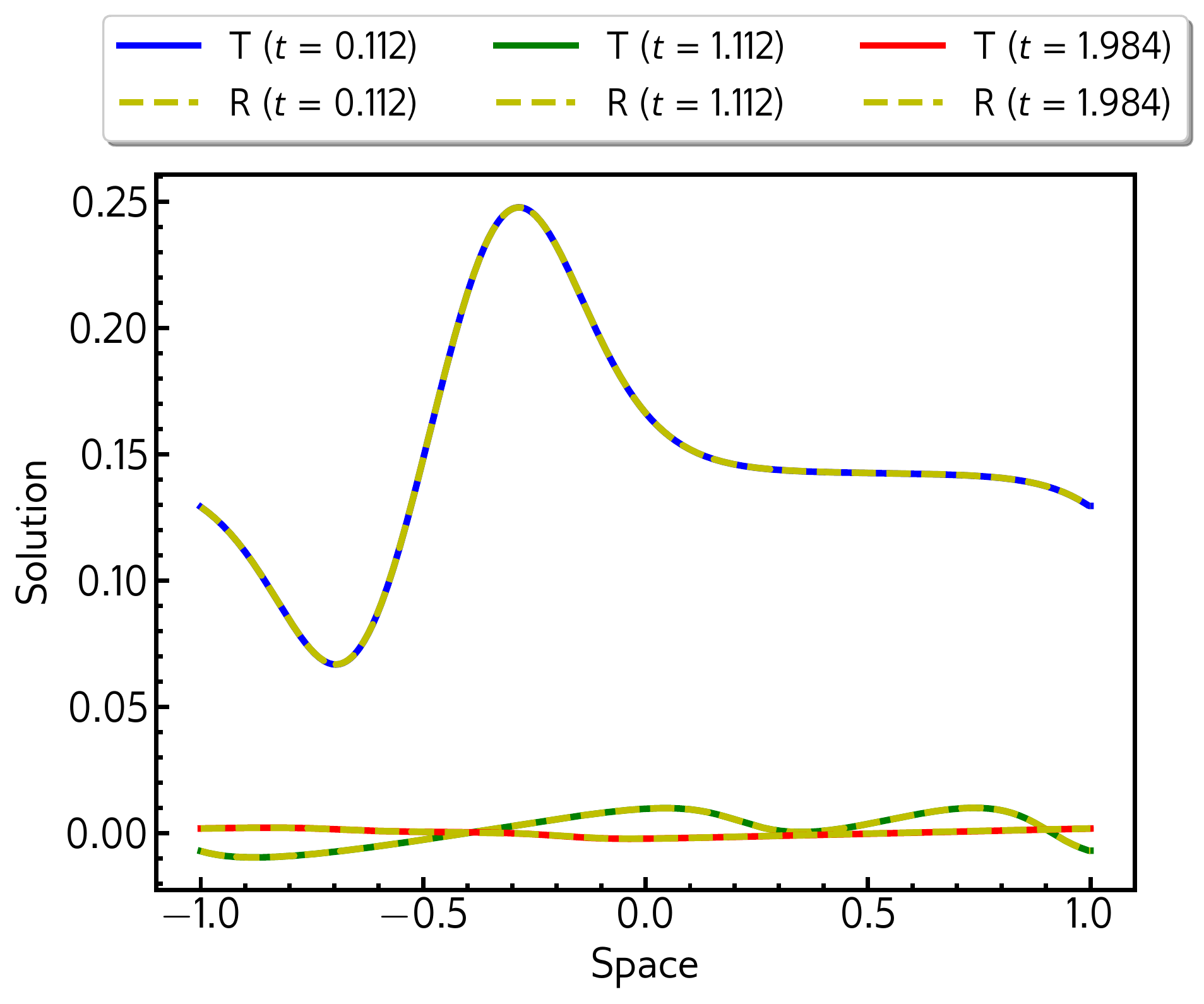}
		\caption{$\nu = 5 \times 10^{-1}$}
		\label{fig:swe-u-rom-sol-a}
	\end{subfigure}
	\begin{subfigure}[b]{0.32\textwidth}
		\centering
		\includegraphics[width=1\textwidth, trim=0 0 0 0, clip]{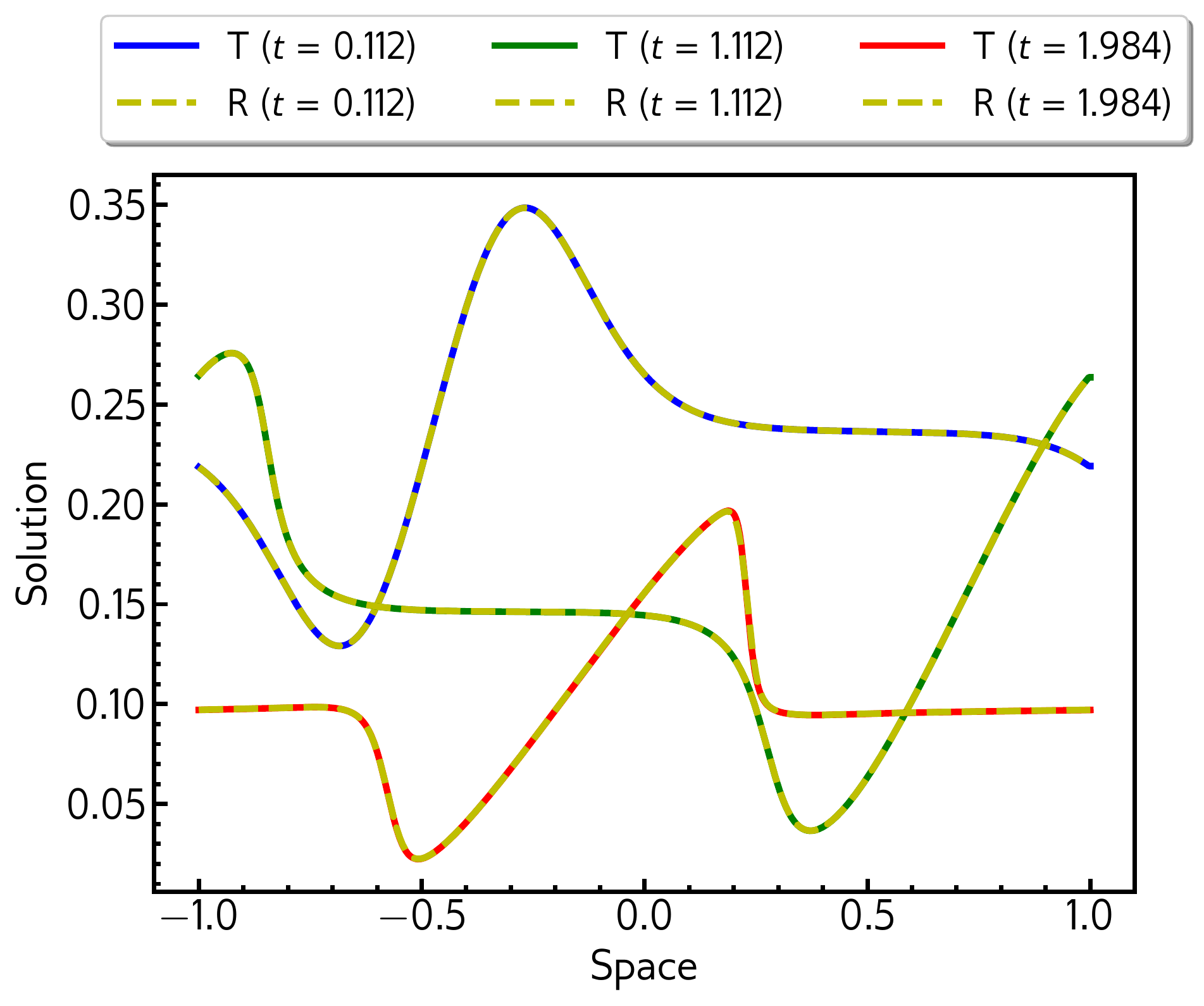}
		\caption{$\nu = 5 \times 10^{-2}$}
		\label{fig:swe-u-rom-sol-b}
	\end{subfigure}
	\begin{subfigure}[b]{0.32\textwidth}
		\centering
		\includegraphics[width=1\textwidth, trim=0 0 0 0, clip]{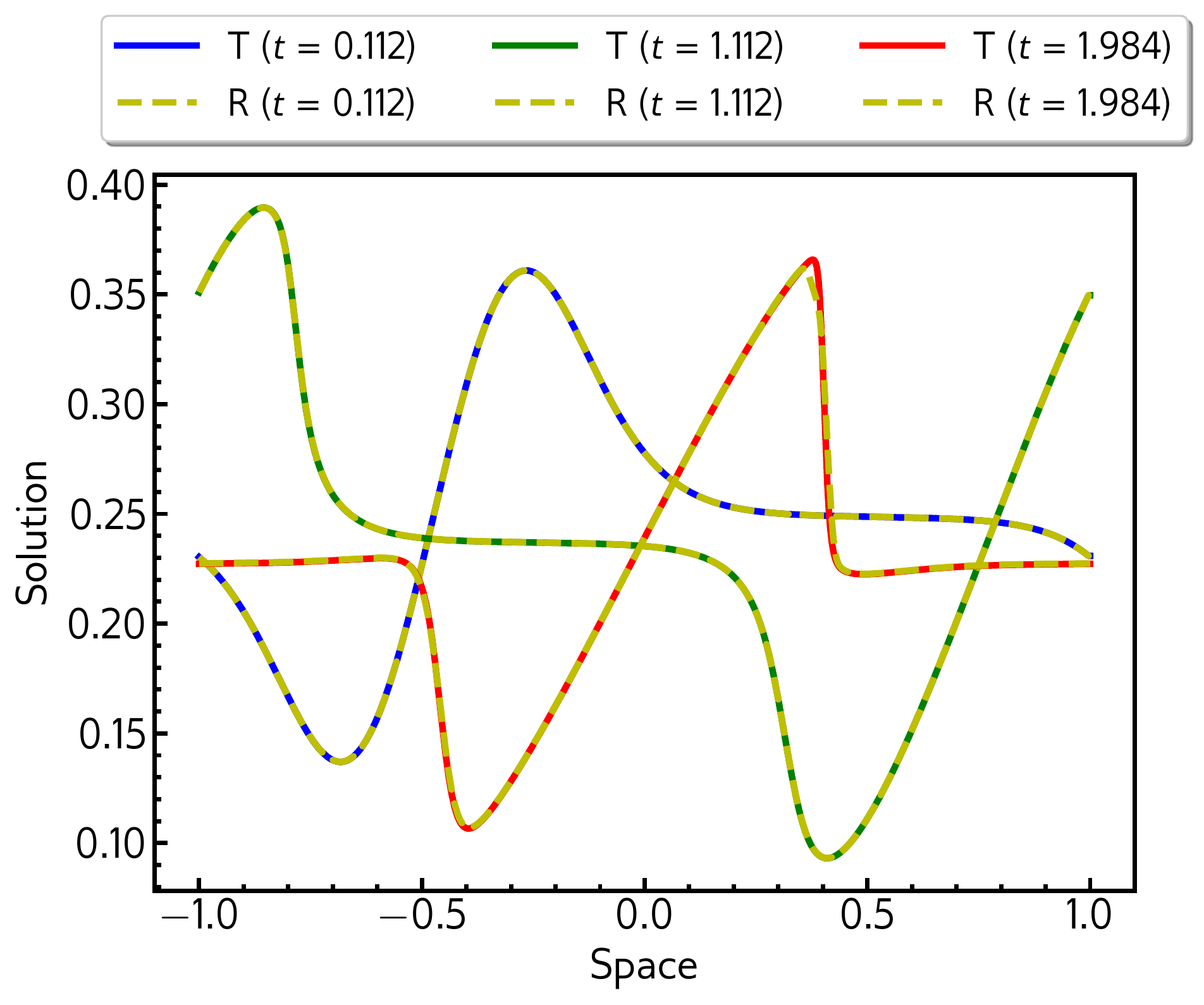}
		\caption{$\nu = 5 \times 10^{-3}$}
		\label{fig:swe-u-rom-sol-c}
	\end{subfigure}
	\begin{subfigure}[b]{0.32\textwidth}
		\centering
		\includegraphics[width=1\textwidth, trim=0 0 0 0, clip]{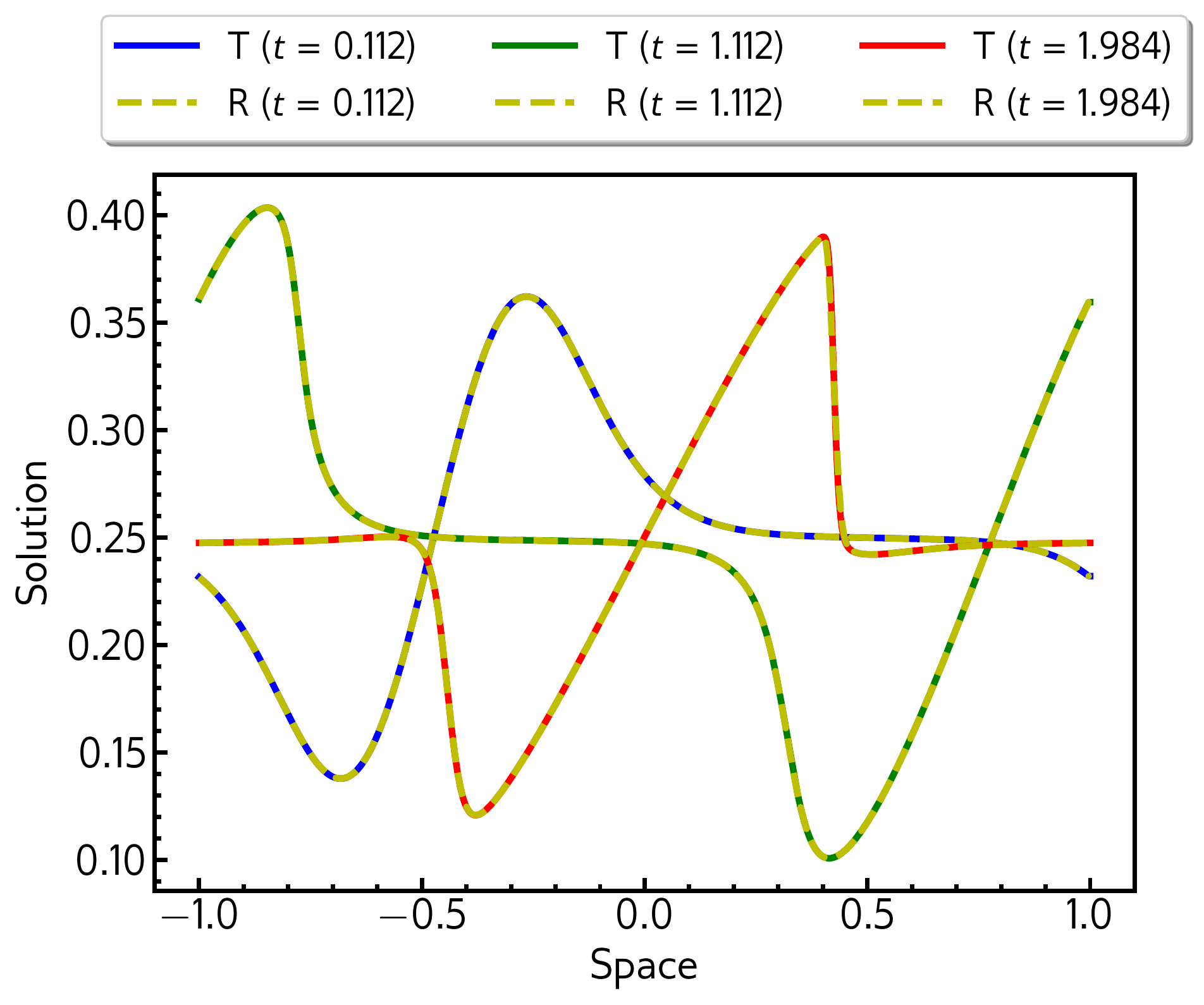}
		\caption{$\nu = 5 \times 10^{-4}$}
		\label{fig:swe-u-rom-sol-d}
	\end{subfigure}
	\begin{subfigure}[b]{0.32\textwidth}
		\centering
		\includegraphics[width=1\textwidth, trim=0 0 0 0, clip]{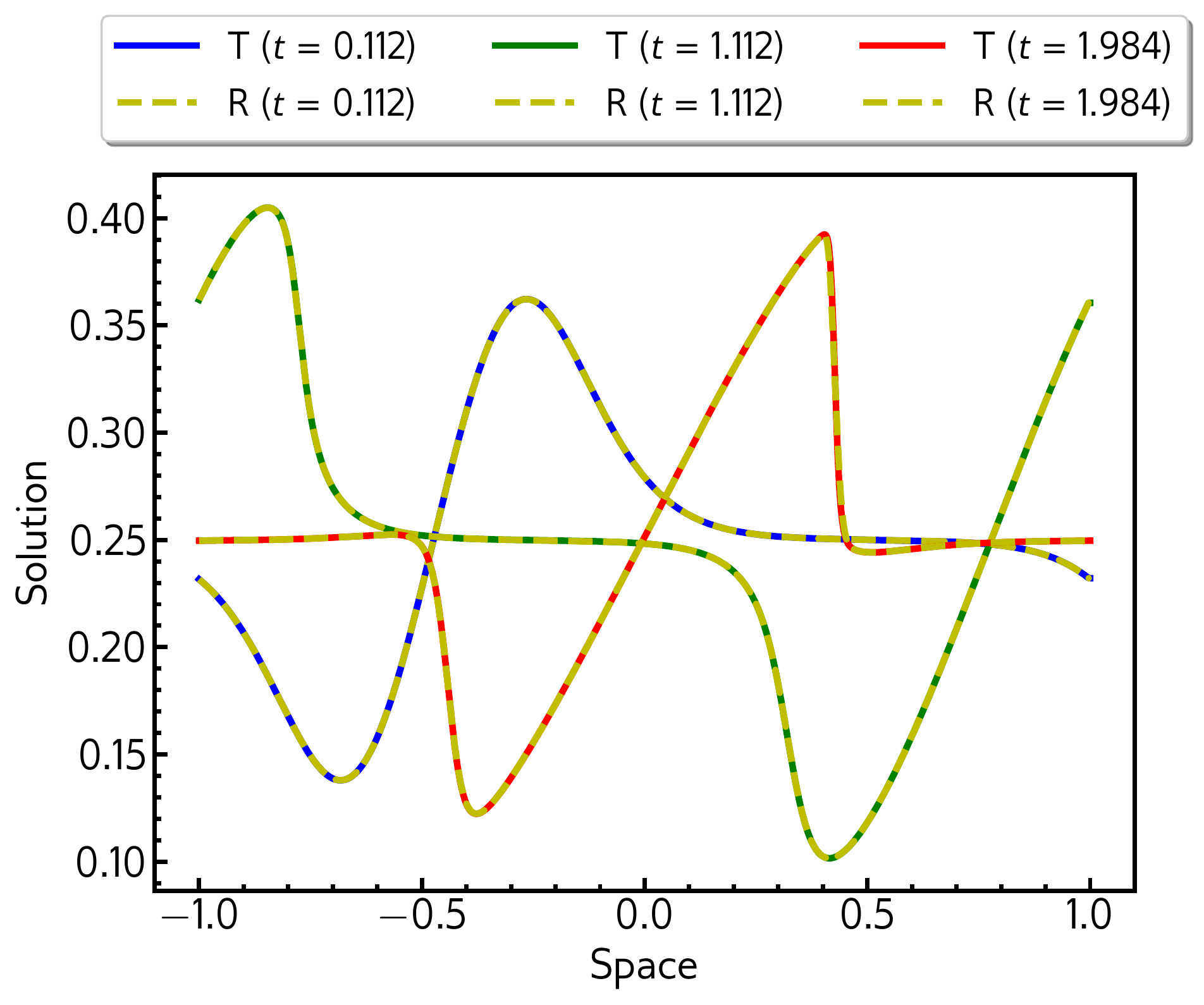}
		\caption{$\nu = 5 \times 10^{-5}$}
		\label{fig:swe-u-rom-sol-e}
	\end{subfigure}
	\caption{Comparison of the ActLearn-POD-KSNN solution (denoted by R) and true solution (denoted by T) for the fluid velocity in the shallow water equations. All the $\nu$ values and time instances $t$ are outside of the training set.}
	\label{fig:swe-u-rom-sol}
\end{figure}

\begin{figure}[H]
	\centering
	\begin{subfigure}[b]{0.47\textwidth}
		\centering
		\includegraphics[width=0.85\textwidth, trim=0 0 0 0, clip]{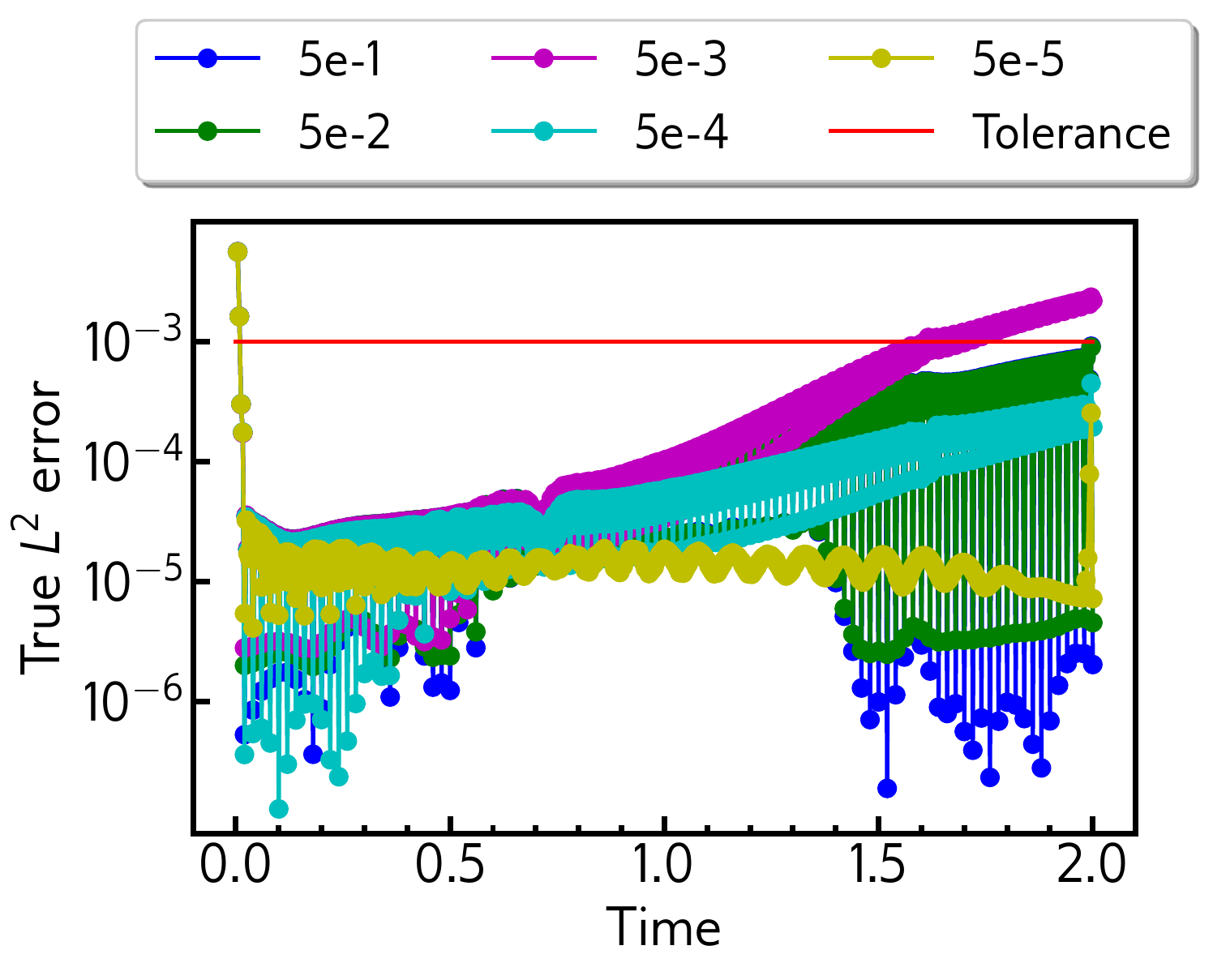}
		\caption{}
		\label{fig:swe-h-error-plot-a}
	\end{subfigure}
	\begin{subfigure}[b]{0.47\textwidth}
		\centering
		\includegraphics[width=0.85\textwidth, trim=0 0 0 0, clip]{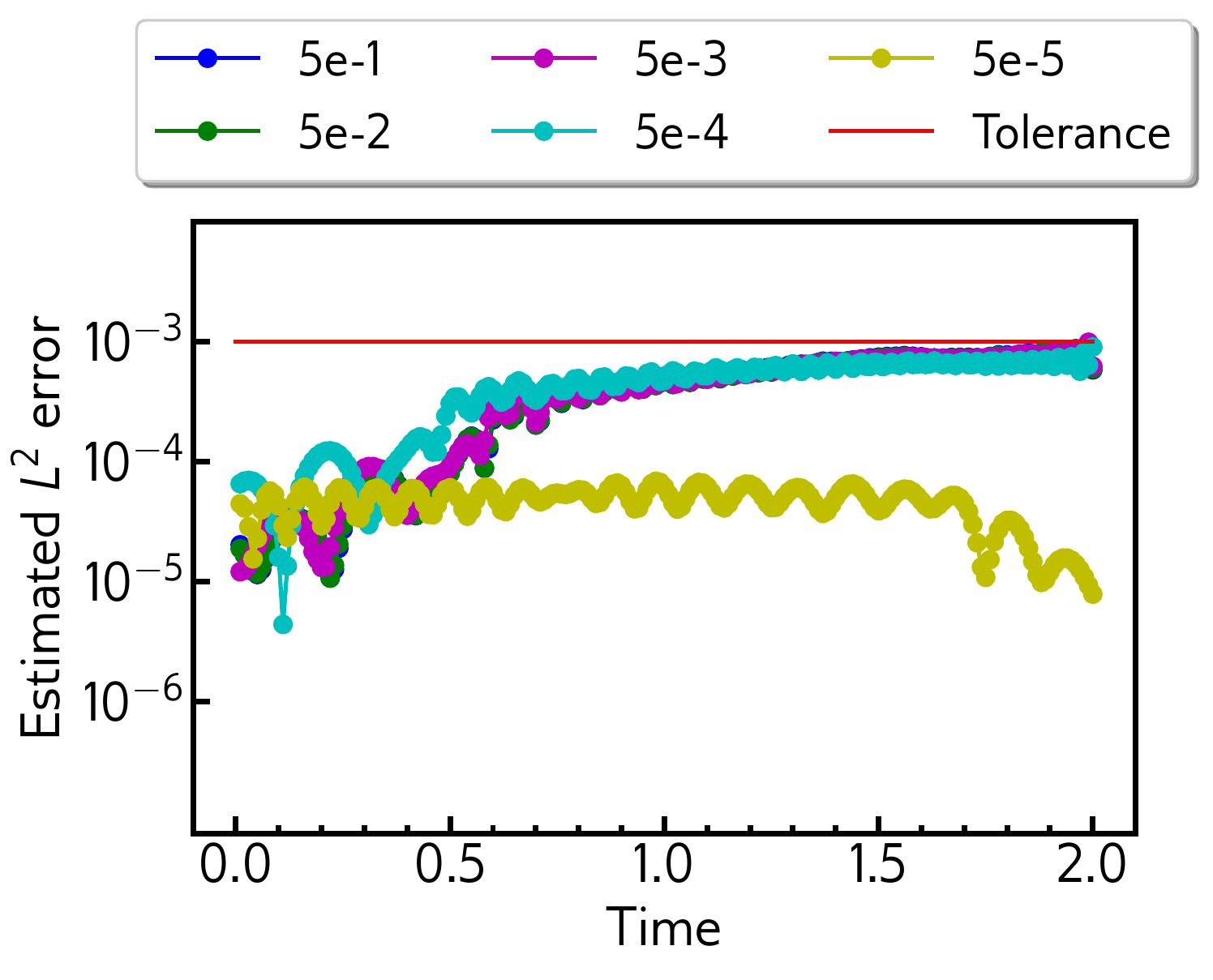}
		\caption{}
		\label{fig:swe-h-error-plot-b}
	\end{subfigure}
	\caption{Plot (a) shows the true error of the ActLearn-POD-KSNN solution for the fluid height of the shallow water equations on a new time grid corresponding to several out-of-training samples of $\nu$. The tolerance used for termination of the active learning procedure is also shown for comparison. Plot (b) shows the estimated error for the fluid height on the original time grid.}
	\label{fig:swe-h-error-plot}
\end{figure}

\begin{figure}[H]
	\centering
	\begin{subfigure}[b]{0.47\textwidth}
		\centering
		\includegraphics[width=0.85\textwidth, trim=0 0 0 0, clip]{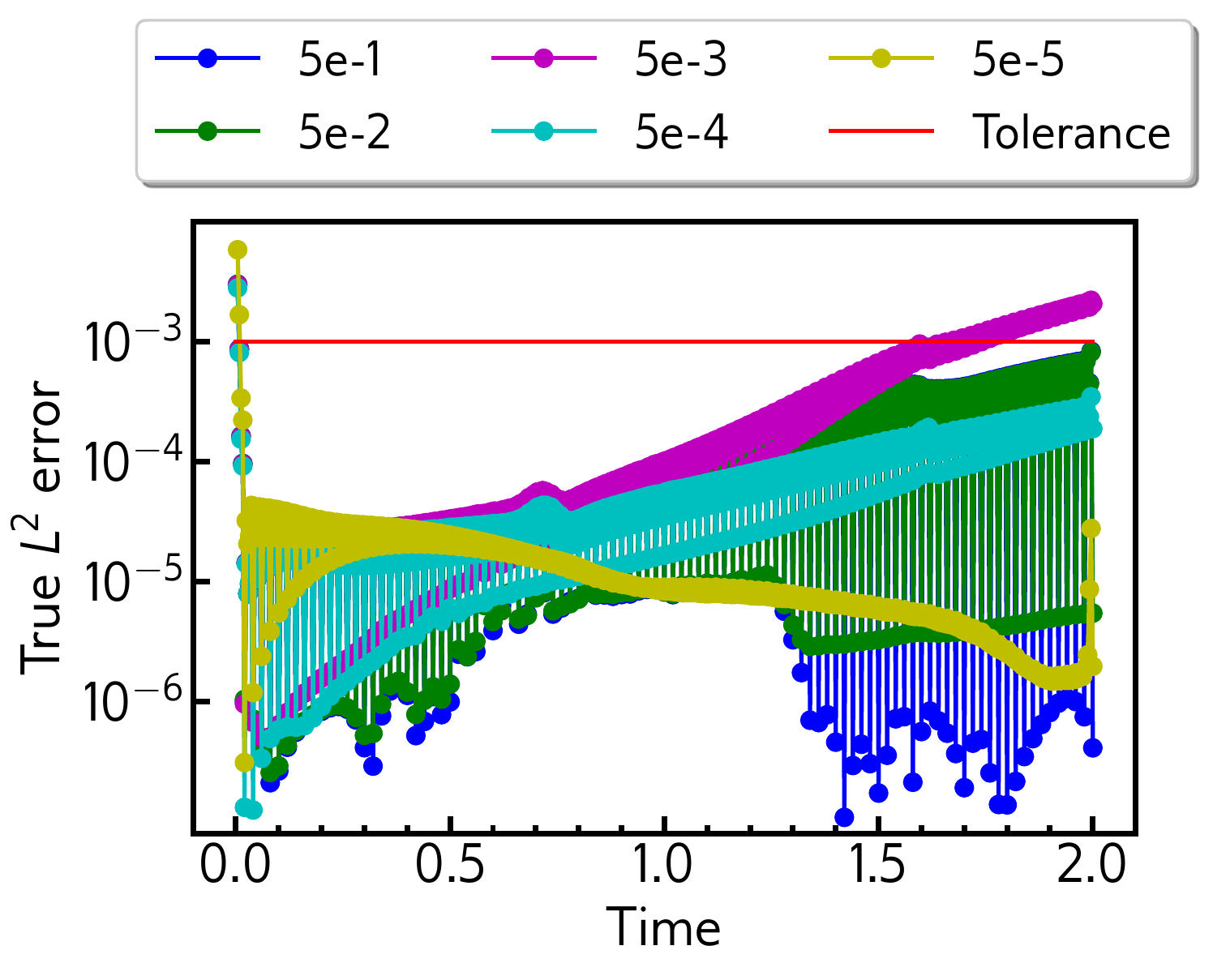}
		\caption{}
		\label{fig:swe-u-error-plot-a}
	\end{subfigure}
	\begin{subfigure}[b]{0.47\textwidth}
		\centering
		\includegraphics[width=0.85\textwidth, trim=0 0 0 0, clip]{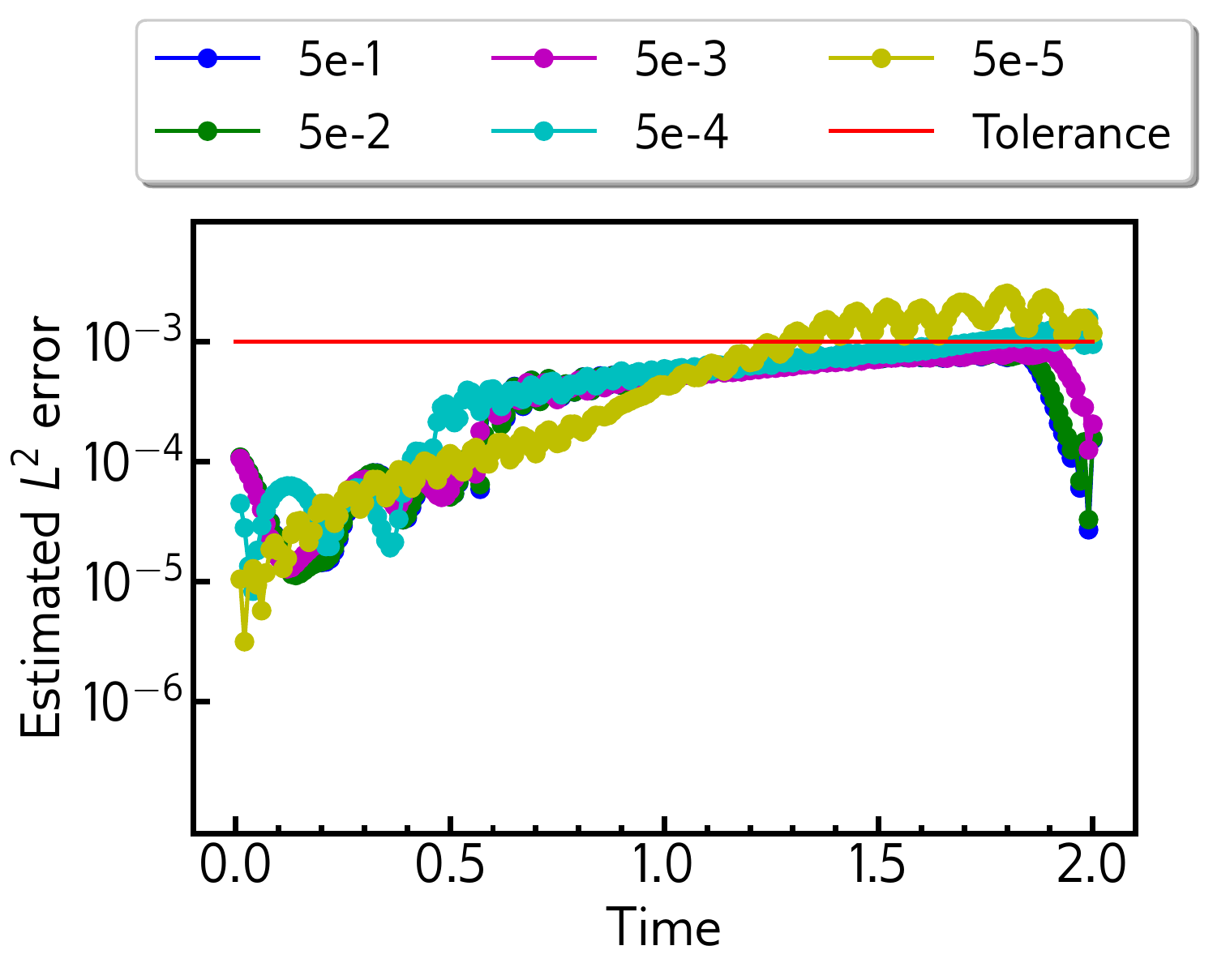}
		\caption{}
		\label{fig:swe-u-error-plot-b}
	\end{subfigure}
	\caption{Plot (a) shows the true error of the ActLearn-POD-KSNN solution for the fluid velocity of the shallow water equations on a new time grid corresponding to several out-of-training samples of $\nu$. The tolerance used for termination of the active learning procedure is also shown for comparison. Plot (b) shows the estimated error for the fluid velocity on the original time grid.}
	\label{fig:swe-u-error-plot}
\end{figure}

\begin{figure}[!t] %
	\centering
	\begin{subfigure}[b]{\textwidth}
		\centering
		\includegraphics[width=\textwidth, trim=0 0 0 0, clip]{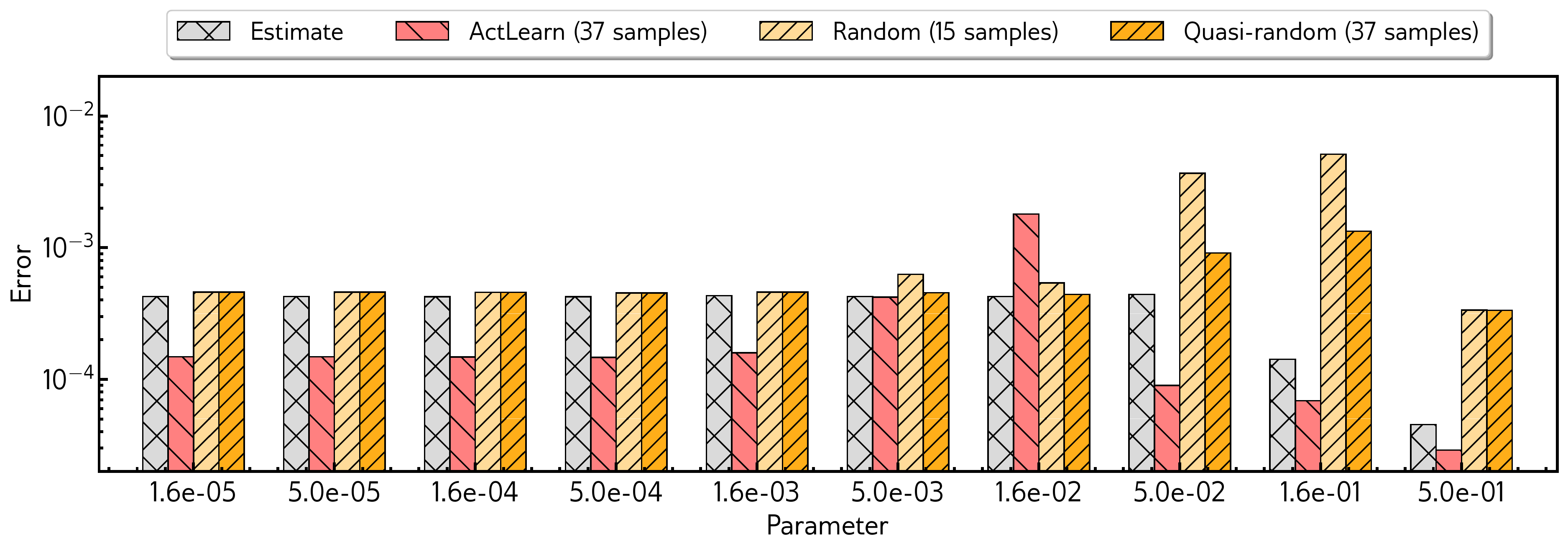}
		\caption{Error in the fluid height for several out-of-training parameter samples $\nu$.}
		\label{fig:swe-random-comparison-a}
	\end{subfigure}
	\begin{subfigure}[b]{\textwidth}
		\centering
		\includegraphics[width=\textwidth, trim=0 0 0 -20, clip]{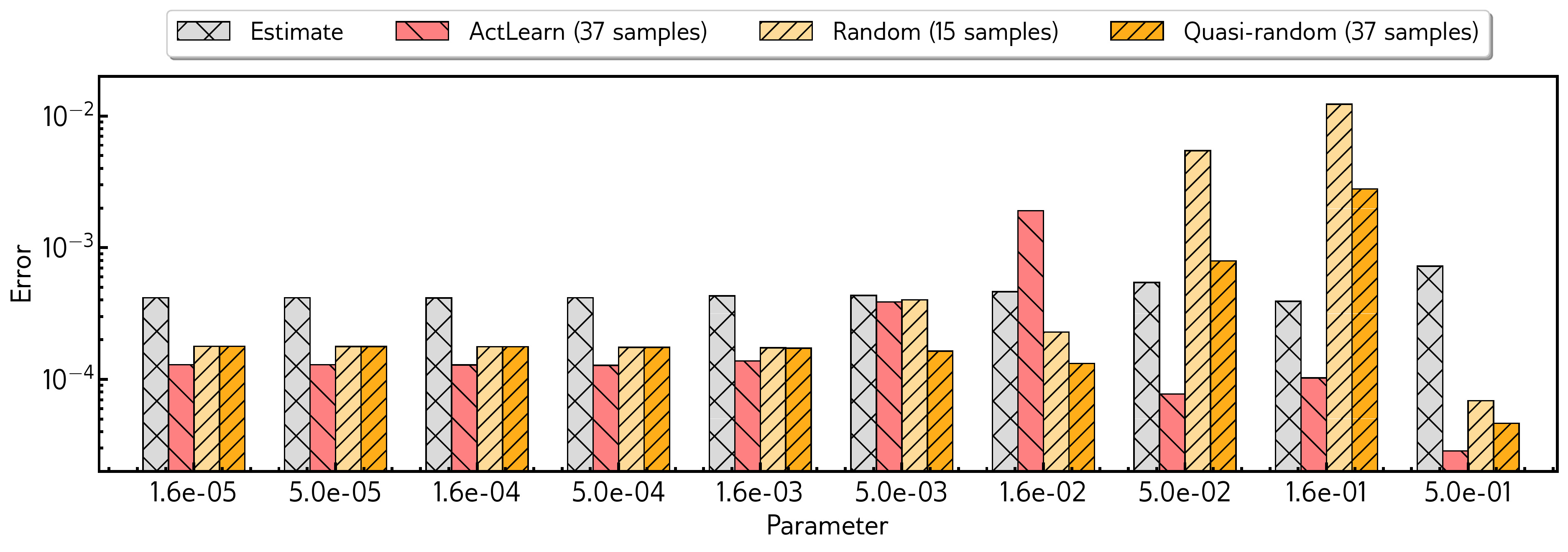}
		\caption{Error in the fluid velocity for several out-of-training parameter samples $\nu$.}
		\label{fig:swe-random-comparison-b}
	\end{subfigure}
	\caption{Shallow water equations: Error comparison between the ActLearn-POD-KSNN solution and the POD-KSNN solutions upon a random (15 samples, seed 10) and quasi-random (37 samples, seed 10) selection of parametric training data. The values labeled 'ActLearn', 'Random', and 'Quasi-random' are the time-averaged (over the test time grid) relative $l^2$ errors in the spatial domain. The values labeled 'Estimate' are the time-average (over the training time grid) of the error estimate values given by \cref{eqn:error-est-time-each-entry}.}
	\label{fig:swe-random-comparison}
\end{figure}


In \Cref{fig:swe-h-rom-sol-surf,fig:swe-u-rom-sol-surf}, the fluid height and velocity obtained from the ActLearn-POD-KSNN surrogate model are compared with the true height and velocity over the entire space-time domain. We can see that the ActLearn-POD-KSNN solutions are able to capture the multiple shock interactions over time, in both the fluid height and the fluid velocity. The viscosity values are taken outside the training set: $\{5 \times 10^{-1}, 5 \times 10^{-2}, 5 \times 10^{-3}, 5 \times 10^{-4}, 5 \times 10^{-5}\}$. The solution is computed on a new test time grid with $499$ instances starting from $0.004$ with a step size of $0.004$. We can see that the ActLearn-POD-KSNN solution agrees well with the ground truth. To further visualize and compare the surrogate solutions with the true solutions, we plot them in \Cref{fig:swe-h-rom-sol,fig:swe-u-rom-sol} at three representative time instances from the start, middle, and end of the time domain. The error contours in the third column of \Cref{fig:swe-h-rom-sol-surf,fig:swe-u-rom-sol-surf} show the point-wise difference in the space-time domain between the surrogate solution and the ground truth.

\begin{figure}[!t] %
	\centering
	\begin{subfigure}[b]{\textwidth}
		\centering
		\includegraphics[width=\textwidth, trim=0 0 0 0, clip]{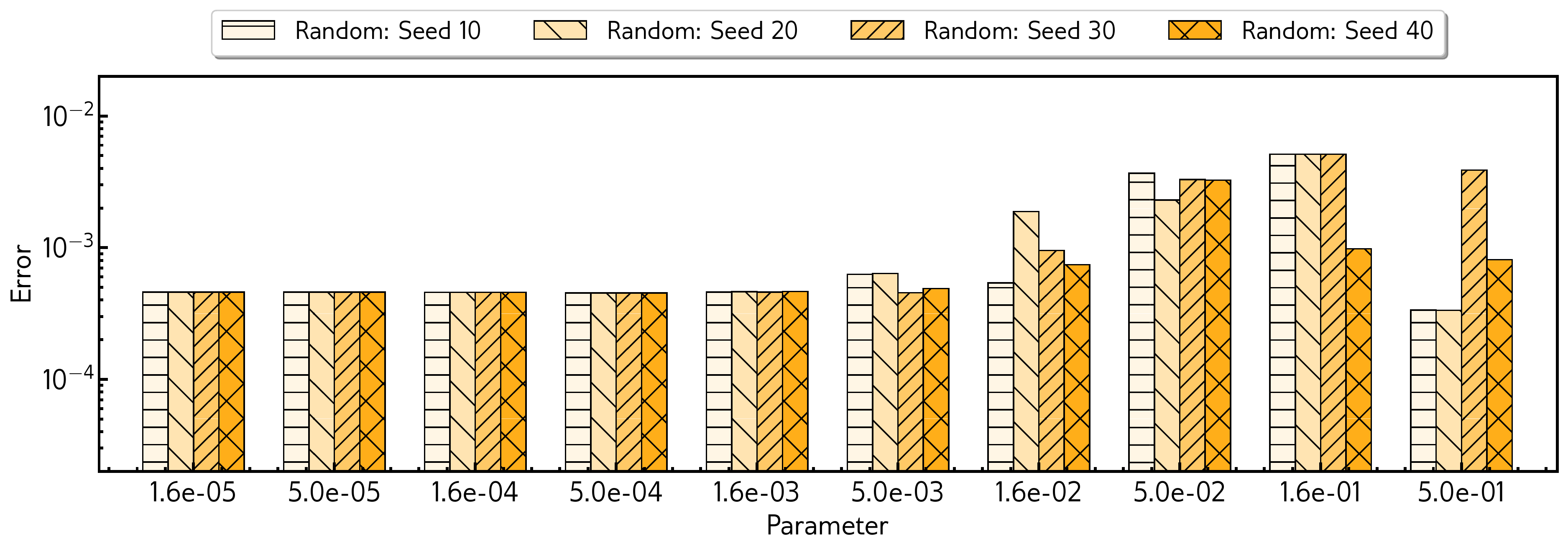}
		\caption{Error in the fluid height upon random selection of parametric training data.}
		\label{fig:swe-random-seed-comparison_h-a}
	\end{subfigure}
	\begin{subfigure}[b]{\textwidth}
		\centering
		\includegraphics[width=\textwidth, trim=0 0 0 -20, clip]{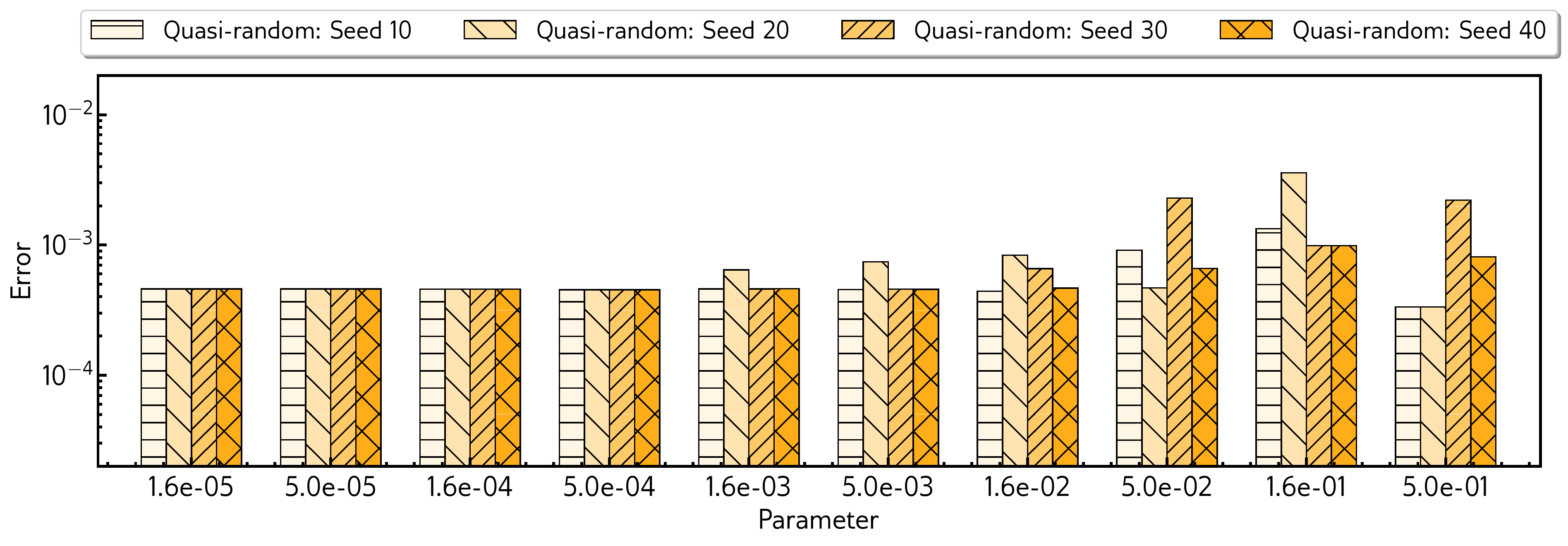}
		\caption{Error in the fluid height upon quasi-random selection of parametric training data.}
		\label{fig:swe-random-seed-comparison_h-b}
	\end{subfigure}
	\caption{Shallow water equations: Error comparison between the POD-KSNN fluid height solution upon a random (15 samples) and quasi-random (37 samples) selection of parametric training data with four different starting seeds: $\{10,20,30,40\}$. The error values are the time-averaged (over the test time grid) relative $l^2$ errors in the spatial domain. All the reported parameter samples $\nu$ are outside of the training set.}
	\label{fig:swe-random-seed-comparison_h}
\end{figure}

An estimation of the error in the fluid height and velocity at several out-of-training parameters is shown in \Cref{fig:swe-h-error-plot-b,fig:swe-u-error-plot-b}, respectively, for the training time grid. Similar to the Burgers' equation, the reported estimates are computed by training KSNNs and obtaining values for $\boldsymbol{\tilde{\varepsilon}}(\boldsymbol{\mu}^*)$ in \cref{eqn:error-est-time}. \Cref{fig:swe-h-error-plot-a,fig:swe-u-error-plot-a} show the true relative error values of the fluid height and velocity obtained from the ActLearn-POD-KSNN surrogate. Here, the time instances are different from the training time grid. For the most part, the true relative errors are bounded by the tolerance criterion $10^{-3}$ which is used for the active learning loop.

\Cref{fig:swe-random-comparison} provides a comparative study between the ActLearn-POD-KSNN solution error and the POD-KSNN solution errors that are obtained by randomly picking $15$ and $37$ parameter samples. The errors in fluid height and velocity are respectively reported in~\Cref{fig:swe-random-comparison-a,fig:swe-random-comparison-b}. Similar to the Burgers' equation example, we label the surrogate solution error obtained by training with $37$ random samples as quasi-random, because this choice is inspired from the active learning procedure. For the random selection from $100$ values of $\nu$ (which are the same as described before during the preparation of sets $P$ and $P^*$), we again fix the random seed in NumPy to $10$. All the surrogate error values and the estimates in~\Cref{fig:swe-random-comparison} are computed in the same way as those in~\Cref{fig:burgers-random-comparison}, i.e., the relative $l^2$ errors in the spatial domain are time-averaged over the test time grid ($t_k$ with $k=1,\ldots, 499$, $t_1=0.004$, a uniform step size of 0.004, and the total discrete time instances are $\tilde{N}_t = 499$), whereas, the reported error estimate values are time-averaged over the training time grid ($t_j$ with $j=1,\ldots, 200$, $t_1=0.0$, a uniform step size of 0.01, and the total discrete time instances are $N_t = 200$).

\begin{figure}[!t]
	\centering
	\begin{subfigure}[b]{\textwidth}
		\centering
		\includegraphics[width=\textwidth, trim=0 0 0 -20, clip]{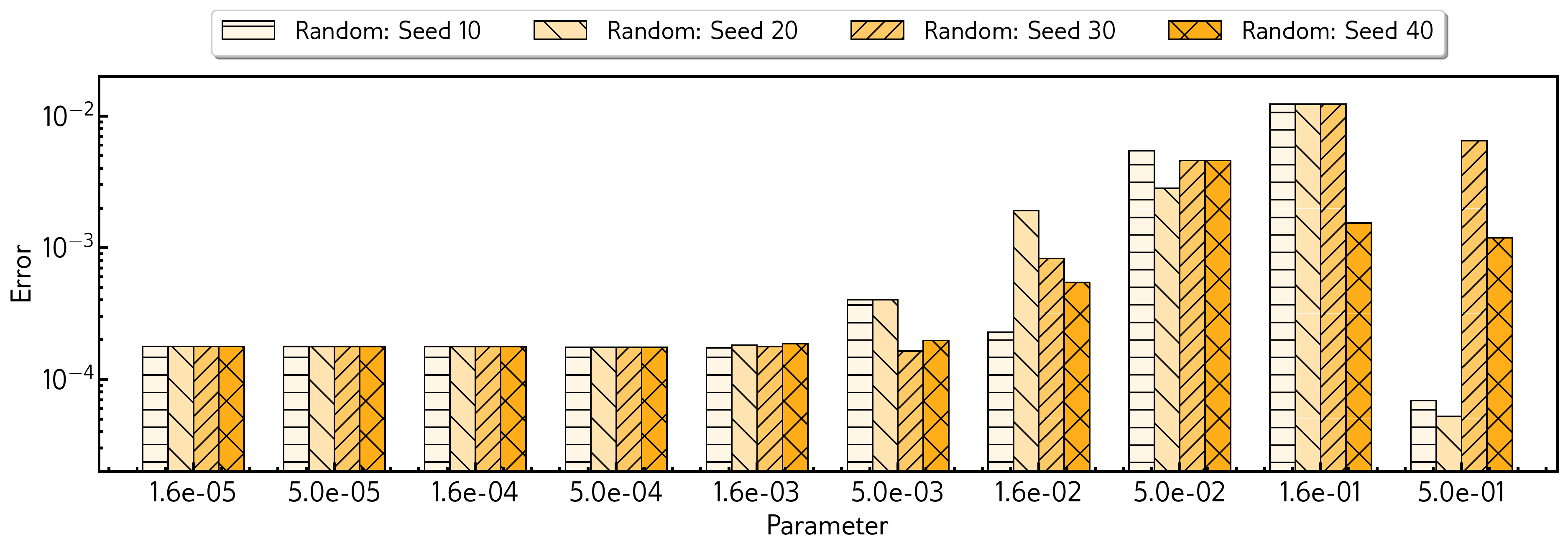}
		\caption{Error in the fluid velocity upon random selection of parametric training data.}
		\label{fig:swe-random-seed-comparison_u-a}
	\end{subfigure}
	\begin{subfigure}[b]{\textwidth}
		\centering
		\includegraphics[width=\textwidth, trim=0 0 0 -20, clip]{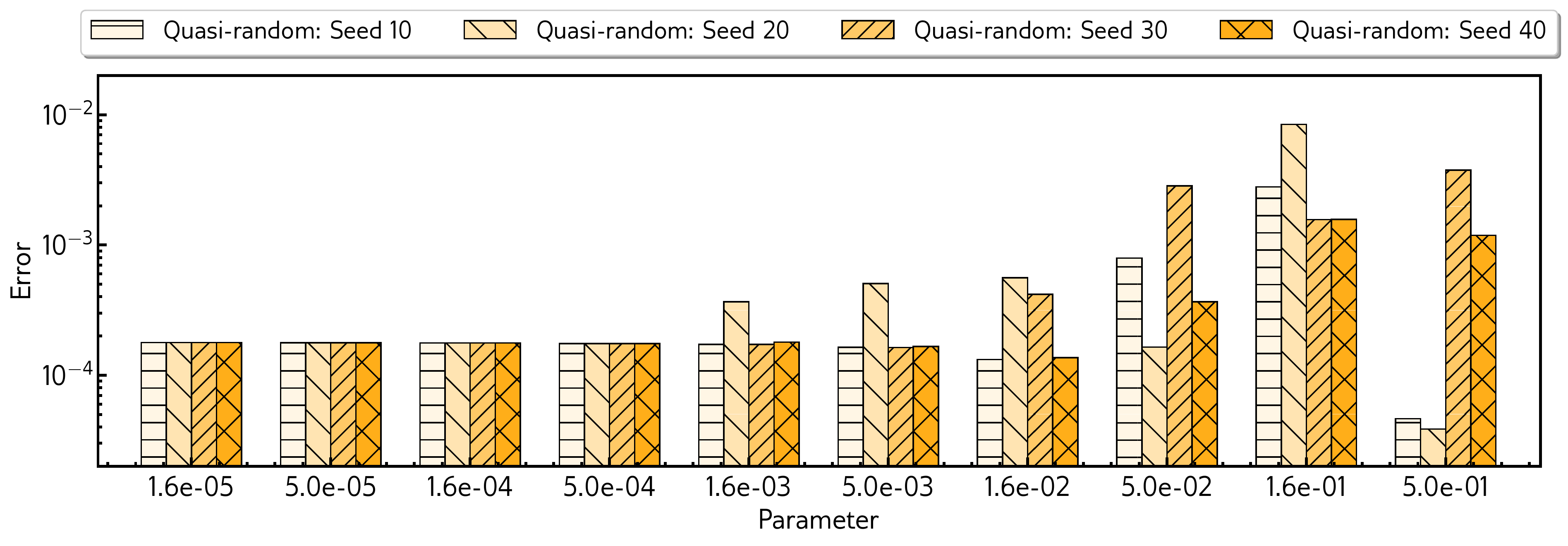}
		\caption{Error in the fluid velocity upon quasi-random selection of parametric training data.}
		\label{fig:swe-random-seed-comparison_u-b}
	\end{subfigure}
	\caption{Shallow water equations: Error comparison between the POD-KSNN fluid velocity solution upon a random (15 samples) and quasi-random (37 samples) selection of parametric training data with four different starting seeds: $\{10,20,30,40\}$. The error values are the time-averaged (over the test time grid) relative $l^2$ errors in the spatial domain. All the reported parameter samples $\nu$ are outside of the training set.}
	\label{fig:swe-random-seed-comparison_u}
\end{figure}

In~\Cref{fig:swe-random-seed-comparison_h,fig:swe-random-seed-comparison_u}, we compare the error in the fluid height and velocity approximated by the POD-KSNN surrogate for scenarios when $15$ and $37$ parameters are randomly picked with four different starting seed values: $\{10,20,30,40\}$. Similar to \Cref{fig:swe-random-comparison}, the reported error values are the time-average (over the test time grid) of the relative $l^2$ error in space. \Cref{fig:swe-random-seed-comparison_h-a,fig:swe-random-seed-comparison_u-a} show that with $15$ random samples, the error goes up to $10^{-2}$, and similar to our observation for Burgers' equation, there is also noticeable difference among the solution error values corresponding to certain testing viscosities for different starting seeds. The error values reduce by adding more samples as seen in \Cref{fig:swe-random-seed-comparison_h-b,fig:swe-random-seed-comparison_u-b}, but there is still a noticeable variation between the values for different starting seeds. So, the accuracy of the surrogate solution is dependent on how the random sampling is done, i.e., the choice of the starting seed. The active learning procedure resolves this situation by picking the parameter samples based on an optimality criterion.

From \Cref{fig:swe-random-comparison} we notice that the error of the ActLearn-POD-KSNN solution is generally the lowest and bounded by the tolerance of $10^{-3}$ specified for termination of the active learning procedure. For $\nu = 1.6 \times 10^{-2}$, the error in the fluid height is $1.79 \times 10^{-3}$ and the fluid velocity is $1.91 \times 10^{-3}$, which is slightly higher than the tolerance. However, these error values are still lower than the maximum error values we observe with random and quasi-random sampling in \Cref{fig:swe-random-seed-comparison_h,fig:swe-random-seed-comparison_u}, i.e., for $\nu = 1.6 \times 10^{-1}$ with a starting seed of $20$. The active learning procedure gives an idea about the most informative parameter samples useful for preparing the snapshot training data. This way the ActLearn-POD-KSNN surrogate solution provides a reasonable accuracy without oversampling the parameter space for preparation of the training snapshot data, thereby staying computationally efficient.

\begin{table}[!b]
	\begin{center}
		\begingroup
		\setlength{\tabcolsep}{10pt} 
		\renewcommand{\arraystretch}{1.5} 
		\begin{tabular}{c || c | c | c}
			\multirow{2}{*}{Number of grid nodes} & \multirow{2}{*}{FOM solver} & \multicolumn{2}{c}{ActLearn-POD-KSNN surrogate model} \\ \cline{3-4}
			& & Offline phase & Online phase \\
			\hline 
			$601$ & $249.56$ & $6.23 + (26 \times 249.56) = 6494.78$ & $0.04$ \\
			$1201$ & $894.62$ & $7.63 + (17 \times 894.62) = 15216.17$ & $0.08$ \\
			\hline
		\end{tabular}
		\endgroup
		\caption{Comparison between runtime (in seconds) of the full-order model (FOM) and the ActLearn-POD-KSNN surrogate model. The simulations are performed for the shallow water equations at two spatial grid sizes. All the reported timings are the average of three independent executions.} 
		\label{tab:swe-timings}
	\end{center}
\end{table}

We report runtime of the full-order shallow water equation solver and the ActLearn-POD-KSNN surrogate model in \Cref{tab:swe-timings}. The numerical tests are carried out on a laptop with Intel{\small\textsuperscript{\textregistered}} Core{\texttrademark} i5-1035G1 CPU @ 1.00GHz and 16 GB of RAM. All the reported timings are the average of three independent executions. The timings reported under offline phase and online phase of the surrogate model are the total execution times for \Cref{alg:summary-offline,alg:summary-online} respectively. The time for active sampling and other offline computations required for building the surrogate, excluding the full-order model query time, is only $6.23$ seconds for a grid size of $601$, and $7.63$ seconds for a grid size of $1201$. The total full-order model query time during the offline phase depends on the number of parameter samples picked by the active learning procedure until its termination, and on the time it takes for generating the full-order solution for one parameter sample. Once the surrogate is built, a fast approximation of the solution is possible, in just $0.04$ seconds for a grid size of $601$, and $0.08$ seconds for a grid size of $1201$. Compared to evaluating the full-order solver at a new parameter sample, we can query the surrogate and obtain the approximate solution with a speedup of greater than $\mathcal{O}(10^3)$, as evident from \Cref{tab:swe-timings}. Note that the surrogate model becomes efficient as soon as it is called more often for unseen parameter values than used in the offline time.


\section{Conclusions}%
\label{sec:conclude}

We have proposed an active learning framework for parametric non-linear dynamical systems that generates solution snapshots at new parameter locations by evaluating the high-fidelity model when necessary. This, in turn, improves the accuracy of the data-driven surrogate model. The central driving force of the active learning process is an non-intrusive error-estimation-based optimality criterion. It is designed from the parameter-specific relative POD approximation errors. Through active learning, we iteratively arrive at a good selection of solution snapshots which are then used to train the data-driven surrogate. In doing so, we relax the vast data requirement for training data-driven surrogate models to some extent, and also provide an estimation of the surrogate accuracy.

The numerical results show that the developed active learning framework iteratively detects locations in the parameter domain where the variation in solution features is high, and prefers new snapshot generation in those regions. For the Burgers' equation, the ActLearn-POD-KSNN surrogate model is able to successfully gauge the variation in its initial conditions and capture the transport of shock profile accurately in time, over the entire range of viscosity values. Moreover, for the shallow water equations, the surrogate model is able to efficiently predict, at new parameter locations, the interacting shock waves that morph into each other over time. The parameter-specific adaptive POD subspaces make our approach efficient, even for problems with mixed---convective and diffusive---phenomena, where each of them dominate in certain regions. Additionally, we observe that the true surrogate errors stay under or are very close to the tolerance level used to terminate the active learning procedure. This indicates reliability of the proposed error estimate that provides us a good measure to gauge the accuracy of the constructed ActLearn-POD-KSNN surrogate model.

The interpolation steps in the active learning loop as well as within the surrogate model's construction are carried out by automatically building, training, and evaluating several kernel-based shallow neural networks. Such a shallow architecture results in a fast offline training stage, as well as a fast online evaluation stage, further reducing the overall computational burden. The training strategy for our ActLearn-POD-KSNN surrogate model is problem independent, and automatically selects the parameter locations whose additional solution snapshots would most improve the non-linear reduced basis space. This minimizes the user interaction for data-driven surrogates built using machine-learning, and the fast online deployment phase brings us a step closer to real-time simulations for high-fidelity parametric physical systems.


\section*{Acknowledgments}%
\addcontentsline{toc}{section}{Acknowledgments}

Harshit Kapadia is supported by the International Max
Planck Research School for Advanced Methods in Process
and Systems Engineering (IMPRS-ProEng).


\addcontentsline{toc}{section}{References}
\bibliographystyle{plainurl} 
\bibliography{refs2} 
  
\end{document}